\documentclass[12pt]{article}

\makeatletter
\def\newleaf{\newpage
\newcount\tmp
\tmp=\c@page
\divide\tmp by 2
\multiply\tmp by 2
\ifnum\c@page=\tmp
~\newpage
\fi
}
\makeatother

\expandafter\ifx\csname proofnewpage\endcsname\relax

\fi

\def\color[#1]#2{}

\long\def\nop#1{}

\def\comment{\edef\cps{\the\parskip} \parskip=0.5cm \begingroup \tt}

\expandafter\ifx\csname shortcite\endcsname\relax

\fi

\newbox\current

\long\def\plframebox#1{
\setbox\current\vbox{#1}		

\vbox to \ht\current {\hrule\vss
\hbox to \wd\current {%
\vrule \hss\box\current\hss \vrule}
\vss\hrule }
}

\long\def\eatpar#1{%
\ifx#1\par                      
\let\nextmove=\eatpar           
\else
\let\nextmove=#1
\fi
\noexpand\nextmove
}

\def\modifymargins#1#2{
\newdimen\addtoh
\newdimen\addtow
\addtoh=#1
\addtow=#2

\advance\topmargin by -\addtoh
\multiply\addtoh by 2
\advance\textheight by \addtoh

\advance\oddsidemargin by -\addtow
\advance\evensidemargin by -\addtow
\multiply\addtow by 2
\advance\textwidth by \addtow
}

\begingroup
\catcode`\~=11
\gdef\centertilde#1{\lower #1pt\hbox{~}}
\endgroup

\newcount\currenttime
\newcount\hour
\newcount\minute

\def\printtime{%
\currenttime=\time
\hour=\currenttime
\divide\hour by 60
\minute=-\hour
\multiply\minute by 60
\advance\minute by \currenttime
\the\hour:\ifnum\minute<10 0\fi\the\minute
}

\begingroup
\makeatletter
\global\let\@@date=\@date
\gdef\@date{\@@date\ --- \printtime}
\endgroup

\def\oggi{\number\day\space 
\ifcase\month\or
Gennaio\or Febbraio\or Marzo\or Aprile\or Maggio\or Giugno\or
Luglio\or Agosto\or Settembre\or Ottobre\or Novembre\or Dicembre\fi
\space \number\year}

\newcounter{rmexample}

\def\proof{\noindent {\sl Proof.\ \ }}

\def\qed{\hfill{\boxit{}}
  \ifdim\lastskip<\medskipamount \removelastskip\penalty55\medskip\fi}
\def\qedn#1{\hfill{\boxit{}$_#1$}
  \ifdim\lastskip<\medskipamount \removelastskip\penalty55\medskip\fi}
\long\def\boxit#1{\vbox{\hrule\hbox{\vrule\kern3pt
                  \vbox{\kern3pt#1\kern3pt}\kern3pt\vrule}\hrule}}

\def\eg{e.g.}

\def\mod{M\!od}

\def\true{{\sf true}}
\def\false{{\sf false}}

\def\p{{\rm P}}
\def\np{{\rm NP}}
\def\conp{{\rm coNP}}

\def\S#1{\mbox{$\Sigma^p_{#1}$}}
\def\P#1{\mbox{$\Pi^p_{#1}$}}

\def\profont{\sf}

\def\x3c{{\profont x3c}}

\def\possnewtheorem#1#2{
\expandafter\ifx\csname #1\endcsname\relax
\newtheorem{#1}{#2}
\fi
}

\def\possnewtheoremthree#1[#2]#3{
\expandafter\ifx\csname #1\endcsname\relax
\newtheorem{#1}[#2]{#3}
\fi
}

\possnewtheorem{theorem}{Theorem}
\possnewtheorem{corollary}{Corollary}
\possnewtheorem{lemma}{Lemma}
\possnewtheoremthree{proposition}[theorem]{Proposition}
\possnewtheorem{definition}{Definition}
\possnewtheorem{question}{Question}
\possnewtheorem{example}{Example}
\possnewtheorem{nontheorem}{Counterexample}
\possnewtheorem{property}{Property}
\possnewtheorem{assumption}{Assumption}
\possnewtheorem{conjecture}{Conjecture}
\possnewtheorem{notation}{Notation}
\newenvironment{theorem*}[1]{{\noindent \bf Theorem~#1}\begin{it}}{\end{it}\

}

\modifymargins{20pt}{20pt}
\possnewtheorem{algorithm}{Algorithm}

\iffalse
\def\frac#1#2{#1/#2}
\makeatletter
\def\@ttyfig#1#2{\def\specialtext{\special{#2}}\specialtext\egroup}
\def
\input{\bgroup\obeylines\obeyspaces\let\-\relax\@ttyfig.tex.latex.tex}
VC
\nop

\let\ttytex\ttyfig
\makeatother
\else
\def\ttytex#1{#1\nop}				
\fi

\title{Belief revision by examples}
\author{Paolo Liberatore\\
DIIAG - Sapienza University of Rome\\
Via Ariosto 25, 00185 Rome, Italy\\
Tel: +39 347 6906915\\
Email: \tt paolo@liberatore.org}

\begin{document}

\iffalse
\long\def\hidden#1{\vskip 0.5cm #1 \vskip 0.5cm}
{\bf remove hidden}
\else
\long\def\hidden#1{}
\fi

\maketitle

\begin{abstract}

A common assumption in belief revision is that the reliability of the
information sources is either given, derived from temporal information, or the
same for all. {\bf This article does not describe a new semantics for
integration} but the problem of obtaining the reliability of the sources given
the result of a previous merging. As an example, the relative reliability of
two sensors can be assessed given some certain observation, and allows for
subsequent mergings of data coming from them.

\end{abstract}

 %


\section{Introduction}

When integrating information coming from different sources, a distinction is
made between revision~\cite{gard-88,darw-pear-97,jin-thie-07,pepp-08,delg-12}
(new information more reliable than old) and
merging~\cite{libe-scha-98-b,chop-etal-06,koni-pere-11} (same reliability).
More generally, priorities or weights are assigned to the sources to indicate
their reliability~\cite{nebe-92,nebe-98,rott-93,delg-dubo-lang-06}. {\em
Measures} and {\em aggregation functions} allow for fine-grained policies of
integration~\cite{koni-etal-04,evar-etal-10,koni-pere-11}. Families of
operators are then defined, all depending in a way or another from the relative
reliability of the sources. The two basic cases of non-iterated revision and
merging result from giving priority to the new information or the same to all
pieces of information to be incorporated, respectively. The strenght of
information sources has been studied in the field of cognitive psychology,
where it was determined to depend on the order in which the information is
given~\cite{wang-etal-00}, on the size of the group generating
it~\cite{mann-09} and other social factors~\cite{see-etal-11}.

The first time merging is done, the relative reliability of the pieces of
information to be integrated cannot come other than from sources external to
the merging process. However, subsequent mergings may then take advantage from
the previous results.

\begin{example}

The two long-range sensors of an unmanned vehicle detect an object. One of the
two identifies it as a wall, the other one as a fence. As the vehicle
approaches, the object enters the range of the vision system, which definitely
concludes it to be a fence. The vehicle turns, and after some distance is
traveled the two long-range sensors disagree again. How the previous conflict
was resolved suggests that the second sensor is more precise than the first.

\end{example}

A similar scenario is that of database fixing after integration: some databases
are merged with equal reliability (in lack of information indicating one to be
more reliable than the other), inconsistencies in the result detected and
corrected by the operators or programmers. If the fixed database is the same of
what would result from merging the original ones with some assumption about the
relative reliability of the sources, that assumption can be considered correct,
and one that should have been used in the first place. This way, integration
and correction provide an ordering of the sources to be used when integrating
other data coming from them.

The problem considered in this article is to estimate the reliability of
formulae $K_1,\ldots,K_m$ so that their integration produces a given other
formula $R$. Contrary to most work in belief revision, no new semantics for
merging are introduced, and this is because the point is not on how to obtain
$R$ from $K_1,\ldots,K_m$, but how to reckon the reliability of
$K_1,\ldots,K_m$ from these formulae and $R$. This formula $R$ is given, not
the outcome of the process: it is the data from the vision system in the first
example and the corrected database in the second. As an example:

\begin{itemize}

\item two sources provide $a$ and $\neg a \wedge b$; lacking information about
their reliability, the result is the disjunction $a \vee (\neg a \wedge b) =
\true$;

\item the actual state of the world is detected to be $\neg a \wedge b$;

\item this formula $\neg a \wedge b$ is the result of merging $a$ and $\neg a
\wedge b$ when the source of the second is assumed more reliable;

\item other two formulae $a \wedge c$ and $b \wedge \neg c$ arrive from the
same sources; given that the second is more reliable, merging produces $b
\wedge \neg c$.

\end{itemize}

The procedure looks straightforward because it involves only two very simple
formulae under a trivial semantics of merging by taking either one of them or
their disjunction, depending on their relative reliability. If none of these
possible outcomes coincide with the given formula then one may (more detailes
are in Section~\ref{unobtainable}):

\begin{enumerate}

\item assume that $R$ is not equal to the expected result of merging but a
``more precise'' formula, or that it represents incomplete information;

\item take into account that some sources produce reliable information on some
aspects of the domain and unreliable in others, so they may be split for
example on the variables;

\item check whether the result can be obtained using a different method of
integration.

\end{enumerate}

The present articles analyze the problem for two existing merging semantics:
minimal sum of
distances~\cite{koni-pere-11,koni-lang-marq-02,koni-lang-marq-04} and
prioritized base merging~\cite{nebe-92,nebe-98,rott-93}, also called discrimin
merging~\cite{delg-dubo-lang-06}. However, any other of the several existing
merging semantics can be used~\cite{koni-pere-11,delg-dubo-lang-06}.

For merging based on sums of
distances~\cite{koni-pere-11,koni-lang-marq-02,koni-lang-marq-04}, a necessary
and sufficient condition for $R$ to be the result of merging $K_1$ and $K_2$
with some weights is given. This result allows to easily derive upper bounds on
the complexity of obtainability, which is in \P{i+1} whenever checking distance
is in \P{i} or in \S{i}. This implies that the problem is in \conp\ for the
drastic distance and in \P{2} for the Hamming distance. Hardness for these
classes is proved. A tractable case for the Hamming distance is determined.
Using the same necessary and sufficient condition, a local search algorithm for
determining the weights is shown

The properties proved for prioritized base
merging~\cite{nebe-92,nebe-98,rott-93} are: some formulae $R$ cannot be
obtained from $K_1,\ldots,K_m$ even if $R$ is the disjunction of some of the
maximally consistent subsets of them; such a condition is only possible with
$m>4$; some other formulae $R$ can be obtained only using $n$ priorities
levels, with an arbitrary $n$ (that requires $n+4$ formulae); if the maximally
consistent subsets form a Berge-acyclic graph, every disjunction of some of
them is obtainable; an algorithm for producing the priority ordering in this
case is given.

If all maximally consistent subsets have size two or less the problem becomes a
problem on graphs, where weights are to be assigned to nodes in such a way some
edges are selected and some other are excluded. In this case, a simple
necessary and sufficient condition for obtainability exists: non-obtainability
is the same as the presence of {\em alternating} cycles of edges.

Surprisingly, complexity turns out not to be higher than that of computing the
result of
merging~\cite{eite-gott-91-b,eite-gott-96,libe-97,libe-97-c,nebe-98,libe-scha-99}
at least in some cases. For example, given a consistent $R$ and
$K_1,\ldots,K_m$ with constant $m$ or with maximally consistent subsets of size
two or less, checking whether $R$ is obtainable is only \conp-complete, thus
solvable within a reasonable size of formulae by modern SAT-solvers.

The article is organized as follows: a section introduces the basic settings,
the following the definitions and results using the sums of distances and
prioritized base merging, respectively, including an algorithm each. Then, the
question on what to do if a given formula is not obtainable is considered. A
final section draws some conclusions.

 %

\section{Preliminaries}

The knowledge bases to be merged are denoted by $K_1,\ldots,K_m$ throughout
this article. They are assumed to be consistent propositional formulae. The
same for the expected result $R$, unless explicitly indicated otherwise.

Two merging semantics are considered in this article, the first based on the
weighted sum of distances, the second on a priority ordering. Formula $R$ is
{\em obtainable} from $K_1,\ldots,K_m$ if it is the result of merging these
formula with some weights or priorities. Using the first semantics, this
amounts to checking the existence of weights such that $R$ is the result of
merging $K_1,\ldots,K_m$ with these weights. For the second semantics, the
definition is the same with a priority ordering instead of the weights.

Obtainability means that $R$ is the result of merging $K_1,\ldots,K_m$ with
some relative reliability among these knowledge bases. Determining this
reliability ordering is the aim of two algorithms, one for each of the
considered merging semantics. What to do if $R$ is not obtainable is considered
in Section~\ref{unobtainable}.

 %


\section{Weighted sum}

\hidden{

[this section: weighted sum of distances, drastic or hamming]
\newline
[this file: only things that are common to weighted sum mergings]
\newline
[completo fino a local search algorithm incluso]
\newline
[ricordarsi sempre che p() puo' sempre essere negativo]

possibili risultati che ha senso inserire:

\begin{itemize}

\item pesi superiori a n+1 sono inutili in ogni caso (o forse no), ma esistono
casi ottenibili solo con pesi n+1 (ma questo nel caso hamming, con drastic in
effetti basta 1,2)

\item se la distanza massima di un modello e' n allora con (n+1,1) si ottiene
la prima kb

\item se un modello ha vettore di distanze superiori entrambe a un altro,
allora non e' mai preso per nessuna coppia di pesi

\end{itemize}

notare da qualche parte che kb di letterali non sono un caso irrilevante (vedi
esempio dei sensori)

}

Model-based merging
operators~\cite{koni-pere-11,koni-lang-marq-02,koni-lang-marq-04} work from a
measure of the distance between models, selecting only the ones that are at
minimal total distance from the knowledge bases. Different semantics result
from different distances measures and different methods for combining them. Two
measures of interest are~\cite{koni-lang-marq-02,reve-97,lin-mend-99}:

\begin{description}

\item[Drastic distance:] $d(I,I)=0$, $d(I,J)=1$ if $J \not= I$;

\item[Hamming distance:] $d(I,J)$ is the number of variables evaluated
differently by $I$ and $J$.

\end{description}

Distance measures extend to knowledge bases: $d(I,K)$ is the minimal value of
$d(I,J)$ for $J \models K$. The drastic distance from a model to a knowledge
base is therefore $0$ if the model satisfies the base and $1$ otherwise. The
Hamming distance is the minimal number of variables that are assigned different
values by the model and by a model of the knowledge base.

Distances can be further extended from one to more knowledge bases in various
ways. One is to define $d(I,K_1,\ldots,K_m)$ to be the sum of the distances
$d(I,K_i)$; other methods exists~\cite{koni-pere-11}. If the sources of the
knowledge base differ in reliability, a weighted sum can be used in place of
the sum~\cite{koni-lang-marq-02,koni-lang-marq-04}. Let $\{w_1,\ldots,w_m\}$ be
the weights, which are assumed positive integers (null, negative or real values
can also be of interest, but are not considered in this article). The weighted
distance from $I$ to $\{K_1,\ldots,K_m\}$ is:

\[
d(I,K_1,\ldots,K_m) = \sum_{1 \leq i \leq m} w_i \times d(I,K_i)
\]

Alternatively, the {\em distance vector} of $I$ is the array
$(d(I,K_1),\ldots,d(I,K_m))$ and the weighted distance is obtained by
multiplying it with the weight vector $(w_1,\ldots,w_m)$. Either way, merging
selects the models of minimal weighted distance from the knowledge
bases~\cite{reve-97,lin-mend-99,koni-lang-marq-02,koni-lang-marq-04}.

The problem of obtainability is that of finding positive integers
$w_1,\ldots,w_m$ such that the result of merging $K_1,\ldots,K_m$ is a given
formula $R$. As usual, the complexity analysis is done on the decision version
of this problem, that of checking the existence of such weights. The algorithm
in Section~\ref{search} searches for actual values. Some considerations on what
to do if they do not exist are in Section~\ref{unobtainable}.

The following restriction is considered in this section: two knowledge bases
only. In other words, $m=2$, and the knowledge bases are $K_1$ and $K_2$. This
restriction simplifies the definition to:

\[
d(I,K_1,K_2) = w_1 \times d(I,K_1) + w_2 \times d(I,K_2)
\]

For every model $I$, its distance vector from $\{K_1,K_2\}$ is
$(d(I,K_1),d(I,K_2))$.

Obtainability amounts to checking the existence of weights that produce the
given result $R$. However, weights $(1,2)$ produce the same results of $(2,4)$,
since the weighted distance of the first pair is double that of the second for
every model; therefore, minimal models are the same. As a result, instead of a
pair of weights $w_1$ and $w_2$ suffices to search for the value of their ratio
$\frac{w_1}{w_2}$. This is a simpler problem because such a value can be
obtained by simple algebraic manipulation from two models of $R$ in most cases.
Otherwise, some constraints on its value derives from models of $\neg R$.

The following expression is useful for relating models, as it often coincides
with $\frac{w_1}{w_2}$ if $I$ and $J$ both satisfy $R$ and gives a bound to
this fraction if $I$ does and $J$ does not.

\ttytex{
\[
p(I,J;K_1,K_2) =
\frac{d(J,K_2)-d(I,K_2)}{d(I,K_1)-d(J,K_1)}
\]
}{

                 d(J,K_2)-d(I,K_2)
p(I,J;K_1,K_2) = -----------------
                 d(I,K_1)-d(J,K_1)

}

Since $K_1$ and $K_2$ are fixed in this section, $p(I,J;K_1,K_2)$ can be
shortened to $p(I,J)$. The knowledge bases $K_1$ and $K_2$ are clear from the
context.

\begin{property}
\label{equal-distance}

Two models $I$ and $J$ have the same distance from $\{K_1,K_2\}$ weighted by
$w_1$ and $w_2$ if and only if either $d(I,K_1)=d(J,K_1)$ and
$d(I,K_2)=d(J,K_2)$ or $d(I,K_1) \not= d(J,K_1)$ and $\frac{w_1}{w_2} =
p(I,J;K_1,K_2)$.

\end{property}

\proof The distance from $I$ and $J$ to $K_1$ and $K_2$ weighted by $w_1$ and
$w_2$ is:

\begin{eqnarray*}
d(I,K_1,K_2) &=& w_1 \times d(I,K_1) + w_2 \times d(I,K_1)	\\
d(J,K_1,K_2) &=& w_1 \times d(J,K_1) + w_2 \times d(J,K_1)
\end{eqnarray*}

If these amounts coincide, then:

\begin{eqnarray*}
w_1 \times d(I,K_1) + w_2 \times d(I,K_1)
&=&
w_1 \times d(J,K_1) + w_2 \times d(J,K_1)
								\\
w_1 \times (d(I,K_1) - d(J,K_1))
\nonumber
&=&
w_2 \times (d(J,K_2) - d(I,K_2))
\end{eqnarray*}

This equation is true if $d(I,K_1)=d(J,K_1)$ and $d(I,K_2)=d(J,K_2)$.
Otherwise, both sides can be divided by $d(I,K_1)=d(J,K_1)$ and by $w_2$, which
by assumption is larger than zero, obtaining:

\ttytex{
\[
\frac{w_1}{w_2}
=
\frac{d(J,K_2) - d(I,K_2)}{d(I,K_1) - d(J,K_1)}
\]
}{
w1   d(J,K_2) - d(I,K_2)
-- = -------------------
w2   d(I,K_1) - d(J,K_1)
}

The right-hand side of this equation is $p(I,J;K_1,K_2)$.~\qed

This property expresses a condition for $I$ and $J$ to have the same weighted
distance from the knowledge bases. If $R$ is the result of merging with weights
$w_1$ and $w_2$, it holds for every two models $I$ and $J$ of it. In
particular, $I$, $J$ and $L$ satisfy the result of merging only if $p(I,J)$ and
$p(I,L)$ both coincide with $\frac{w_1}{w_2}$, which implies $p(I,J)=p(I,L)$.
In other words, $p(I,J)$ gives the value of $\frac{w_1}{w_2}$, any other
$p(I,L)$ has to agree on it.

\begin{property}
\label{greater-distance}

Model $I$ is closer than model $M$ to $\{K_1,K_2\}$ with weights $w_1$ and
$w_2$ if and only if:

\begin{itemize}

\item $d(I,K_1)=d(M,K_1)$ and $d(I,K_2)<d(M,K_2)$; or

\item $d(I,K_1)-d(M,K_1)>0$ and $\frac{w_1}{w_2} < p(I,M;K_1,K_2)$; or

\item $d(I,K_1)-d(M,K_1)<0$ and $\frac{w_1}{w_2} > p(I,M;K_1,K_2)$.

\end{itemize}

\end{property}

\proof The distance is $w_1 \times d(I,K_1) + w_2 \times d(I,K_1)$ for $I$ and
$w_1 \times d(M,K_1) + w_2 \times d(M,K_1)$ for $M$. Therefore, $I$ is closer
than $M$ to $\{K_1,K_2\}$ if:

\begin{eqnarray*}
w_1 \times d(I,K_1) + w_2 \times d(I,K_1)
&<&
w_1 \times d(M,K_1) + w_2 \times d(M,K_1)
								\\
\nonumber
w_1 \times (d(I,K_1) - d(M,K_1))
&<&
w_2 \times (d(M,K_2) - d(I,K_2))
\end{eqnarray*}

By assumption, $w_2$ is strictly positive. Therefore, both sides of this
inequation can be divided by it. Instead, $d(I,K_1) - d(M,K_1)$ may be
positive, negative or zero. In latter case, $d(I,K_1)=d(M,K_1)$, which implies
that $I$ is closer than $M$ to the bases if and only if $d(I,K_2)<d(M,K_2)$,
regardless of the weights.

If $d(I,K_1) - d(M,K_1)$ is positive, both sides of the inequation can be
divided by it:

\ttytex{
\[
\frac{w_1}{w_2}
<
\frac{d(M,K_2) - d(I,K_2)}{d(I,K_1) - d(M,K_1)}
\mbox{ if } d(I,K_1) - d(M,K_1) > 0
\]
}{
w1   d(M,K_2) - d(I,K_2)
-- < -------------------        if d(I,K_1) - d(M,K_1) > 0
w2   d(I,K_1) - d(M,K_1)
}

The inequation is $\frac{w_1}{w_2}<p(I,M;K_1,K_2)$. In the other case, dividing
both sides by the negative number $d(I,K_1) - d(M,K_1)$ changes $<$ into $>$:

\ttytex{
\[
\frac{w_1}{w_2}
>
\frac{d(M,K_2) - d(I,K_2)}{d(I,K_1) - d(M,K_1)}
\mbox{ if } d(I,K_1) - d(M,K_1) < 0
\]
}{
w1   d(M,K_2) - d(I,K_2)
-- > -------------------        if d(I,K_1) - d(M,K_1) < 0
w2   d(I,K_1) - d(M,K_1)
}

The inequation is $\frac{w_1}{w_2}>p(I,M;K_1,K_2)$.~\qed

These properties show that most pairs of models constraint the value of
$\frac{w_1}{w_2}$. In particular, two models of $R$ are enough to uniquely fix
it, unless they are at the same distance from $K_1$. Models that do not satisfy
$R$ only generate inequations. If there are at least two models of $R$ at
different distances from $K_1$ this is not a problem, as these determine
$\frac{w_1}{w_2}$ and what is left to do is check the inequations.

Otherwise, more complex constraints among models not satisfying $R$ may result.
As an example, if all models of $R$ are at distance $(4,4)$ and two models not
of $R$ at distance $(1,8)$ and $(8,1)$, then $R$ is obtainable with
$w_1=w_2=1$. Instead, two other models not in $R$ at distance $(1,5)$ and
$(5,1)$ make $R$ unobtainable.

If $I,J,L \models R$, then both $p(I,J)$ and $p(I,L)$ coincide with
$\frac{w_1}{w_2}$, and therefore coincide with each other: $p(I,J)=p(I,L)$. For
the same reason, if $I,J \models R$ and $L \not\models R$, then $p(I,J)<p(I,L)$
or $p(I,J)>p(I,L)$, depending on the sign of $d(I,K_1)-d(L,K_1)$.

These constraints are enough is $R$ has at least two models with differing
distance from $K_1$. Otherwise, $R$ does not set a value for $\frac{w_1}{w_2}$,
which can therefore be varied to exclude models not satisfying $R$. In
particular, two inequations of opposite comparison can be combined: if $I
\models R$, $M,N \not\models R$, $d(I,K_1)-d(N,K_1)>0$ and
$d(I,K_1)-d(M,K_1)<0$, then $\frac{w_1}{w_2}<p(I,N)$ and
$\frac{w_1}{w_2}>p(I,M)$, leading to $p(I,M)<p(I,N)$.

\begin{lemma}
\label{weighted-obtainable}

A satisfiable formula $R$ is obtainable from $\{K_1,K_2\}$ if and only if for
all $I,J,L \models R$ and $M,N \not\models R$, the following conditions hold:

\begin{enumerate}

\item if $d(I,K_1) \geq d(J,K_1)$ then $d(I,K_2) \leq d(J,K_2)$

\item if $d(I,K_1) \geq d(M,K_1)$ then $d(I,K_2)<d(M,K_2)$

\item $p(I,J) = p(I,L)$\hfill\break
if $d(I,K_1)-d(J,K_1) \not= 0$ and $d(I,K_1)-d(L,K_1) \not= 0$

\item $p(I,J) < p(I,M)$\hfill\break
if $d(I,K_1)-d(J,K_1) \not= 0$ and $d(I,K_1)-d(M,K_1)>0$

\item $p(I,J) > p(I,M)$\hfill\break
if $d(I,K_1)-d(J,K_1) \not= 0$ and $d(I,K_1)-d(M,K_1)<0$

\item $p(I,N) < p(I,M)$ \hfill
\break
if $d(I,K_1)-d(M,K_1)>0$ and $d(I,K_1)-d(N,K_1)<0$

\end{enumerate}

\

\end{lemma}

\proof Assuming the conditions true, we derive values of $w_1$ and $w_2$ that
make the result of merging being exactly $R$. Two cases are possible: in the
first, all models of $R$ have the same distance to $K_1$ and the same distance
to $K_2$; in the second, at least two models of $R$ have different distances.

If all models of $R$ are at the same distance from $K_1$ and from $K_2$, then
every pair of weights makes them having the same weighted distance. Therefore,
the problem is only with models not in $R$, which must be at a greater
distance. Let $I$, $M$ and $N$ be:

\begin{itemize}

\item $I$ is a model of $R$;

\item $M$ is one of the models not satisfying $R$ with a minimal value of
$p(I,M)$ among the ones with $d(I,K_1)-d(M,K_1)>0$, if any;

\item $N$ is one of the models not satisfying $R$ with a maximal value of
$p(I,N)$ among the ones with $d(I,K_1)-d(N,K_1)<0$, if any.

\end{itemize}

By the sixth condition of the lemma, in these conditions $p(I,N)<p(I,M)$. If
$\frac{w_1}{w_2}$ is between $p(I,N)$ and $p(I,M)$, then $\frac{w_1}{w_2}$ is
smaller than $p(I,M')$ for every $M' \not\models R$ with
$d(I,K_1)-d(M',K_1)>0$, thanks to the minimality of $M$. By
Property~\ref{greater-distance}, this implies that $M'$ is further from the
bases than $I$. The same applies to models $N'$ with $d(I,K_1)-d(N',K_1)<0$,
thanks to the maximality of $N$. For the models $L$ such that
$d(I,K_1)-d(L,K_1)=0$, the second condition of the lemma implies that
$d(I,K_2)<d(L,K_2)$, proving that they are further from the bases than $I$
regardless of the weights.

If no such $M$ or no such $N$ exist, the corresponding constraint is void. This
can be formalized by replacing $p(I,N)$ with $0$ and $p(I,M)$ with $n$.

A value between $p(I,M)$ and $p(I,N)$ is their average. However, this may be
negative, and negative weights are not allowed. In this case, a different
method can be employed.

If $d(I,N)$ is negative, $\frac{w_1}{w_2}$ is determined as follows. Since
$d(I,K_1)>d(M,K_1)$, by the second condition of the lemma $d(I,K_2)<d(M,K_2)$,
which ensures that $p(I,M)$ is strictly positive. By definition of this
expression, its minimal positive value is $\frac{1}{n}$, obtained by taking the
minimal value of the numerator ($1$ or $-1$) and the maximal value of the
denominator ($n$ or $-n$). Since $d(I,N)$ is negative, a value between it and
$\frac{1}{n}$ is $\frac{1}{n+1}$.

If $d(I,N)$ is positive, this value may not work, but the average between it
and $d(I,M)$ is positive, and can therefore be used. Let $p(I,M)=\frac{a}{b}$
and $p(I,N)=\frac{c}{d}$.

\begin{eqnarray*}
\lefteqn{\frac{p(I,N) + p(I,M)}{2}}		\\
&=& \frac{\frac{a}{b}+\frac{c}{d}}{2}		\\
&=& \frac{a}{2b}+\frac{c}{2d}			\\
&=& \frac{ad}{2bd}+\frac{cb}{2bd}		\\
&=& \frac{ad+cb}{2bd}
\end{eqnarray*}

Since this is the average between two positive values, it is positive. The
numerator and the denominator may both be negative, but their absolute values
produce the same fraction. Since this is $\frac{w_1}{w_2}$, the weights can be
taken to be:

\begin{eqnarray*}
w_1 &=&
	|
	(d(N,K_2)-d(I,K_2)) \times (d(I,K_1)-d(M,K_1))
	+
\\
&&
	(d(M,K_2)-d(I,K_2)) \times (d(I,K_1)-d(N,K_1))
	|
\\
w_2 &=& |2 \times (d(I,K_1)-d(M,K_1)) \times (d(I,K_1)-d(N,K_1))|
\end{eqnarray*}

Using such weights, every model not satisfying $R$ is further from the bases
than all models satisfying $R$, which proves that if all models of $R$ have the
same distances from $K_1$ and $K_2$, then $R$ is obtainable if the conditions
in the statement of the lemma are true.

\

If there exists $I$ and $J$ such that $d(I,K_1) \not= d(I,K_1)$, then
$\frac{w_1}{w_2}$ is uniquely determined by Property~\ref{equal-distance} to be
$p(I,J)$:

\ttytex{
\[
\frac{w_1}{w_2}
=
\frac{d(J,K_2) - d(I,K_2)}{d(I,K_1) - d(J,K_1)}
\]
}{
w1   d(J,K_2) - d(I,K_2)
-- = -------------------
w2   d(I,K_1) - d(J,K_1)
}

Two values producing this fraction are:

\begin{eqnarray*}
w_1 &=& |d(J,K_2) - d(I,K_2)| \\
w_2 &=& |d(I,K_1) - d(J,K_1)|
\end{eqnarray*}

By the first assumption of the lemma, if $d(I,K_1) - d(J,K_1)$ is negative then
$d(I,K_2) - d(J,K_2)$ is positive, and vice versa. As a result,
$\frac{w_1}{w_2}$ is $\frac{d(J,K_2) - d(I,K_2)}{d(I,K_1) - d(J,K_1)}$ despite
the absolute values.

Let $L$ be another model of $R$. If $d(I,K_1)=d(L,K_1)$, by the first condition
of the lemma $d(I,K_2)=d(L,K_2)$, which implies that $I$ and $L$ are at the
same weighted distance from the bases regardless of the weights. Otherwise,
$d(I,K_1) \not= d(I,K_2)$, and Property~\ref{equal-distance} applies: if
$\frac{w_1}{w_2}=p(I,L)$ then $I$ and $L$ are at the same distance from the
bases. But $\frac{w_1}{w_2}$ has been proved to be equal to $p(I,J)$, and by
the second assumption of the lemma $p(I,J)=p(I,L)$.

Let $M \not\models R$. By the assumptions of the lemma, $p(I,J) < p(I,M)$ if
$d(I,K_1)-d(M,K_1)>0$ and $p(I,J) > p(I,M)$ if $d(I,K_1)-d(M,K_1)<0$. By
Property~\ref{greater-distance}, the distance from $M$ to $\{K_1,K_2\}$ is
greater than that of $I$. That concludes the proof that if the conditions of
the lemma are true then $R$ is obtainable.

\

If some of the conditions of the lemma are falsified, then $R$ is not
obtainable from $\{K_1,K_2\}$ with any weights. This is proved for each
condition at time.

The first condition is false if $d(I,K_1) \geq d(J,K_1)$ but
$d(I,K_2)>d(J,K_2)$. In such conditions the weighted distance of $I$ is less
than that of $J$ regardless of the weights, implying that $J$ is not in the
result of the merging in spite of $J \models R$.

The second condition is false if $d(I,K_1) \geq d(M,K_1)$ and $d(I,K_2) \geq
d(M,K_2)$, which imply that the weighted distance of $I$ is greater than or
equal to that of $M$ regardless of the weights, implying that either $M$ is in
the result of merging or $I$ is not, while $I \models R$ and $M \not\models R$.

The third condition is false if $p(I,J) \not= p(I,L)$ for some $I,J,L \models
R$ with $d(I,K_1) \not= d(J,K_1)$ and $d(I,K_1) \not= d(L,K_1)$. By
Property~\ref{equal-distance}, $I$ and $J$ are at the same distance only if
$\frac{w_1}{w_2}$ is $p(I,J)$; $I$ and $L$ are at the same distance only if it
is $p(I,L)$. These are different, showing that no pair of weights makes $I$,
$J$ and $L$ to be at the same weighted distance from the bases.

The fourth condition is false if $d(I,K_1) \not= d(J,K_1)$, $d(I,K_1)>d(M,K_1)$
and $p(I,J) \geq p(I,M)$. The first implies $\frac{w_1}{w_2} = p(I,J)$ by
Property~\ref{equal-distance} and $I,J \models R$, the second that
$\frac{w_1}{w_2} < p(I,M)$ by Property~\ref{greater-distance} and $I \models R$
and $M \not\models R$. Therefore, $p(I,J)<p(I,M)$, contradicting $p(I,J) \geq
p(I,M)$. 

The fifth condition is similar, with $d(I,K_1)<d(M,K_1)$ implying
$\frac{w_1}{w_2} > p(I,M)$, which together with $\frac{w_1}{w_2}=p(I,J)$
contradicts $p(I,J) \leq p(I,M)$.

The sixth condition is false if $d(I,K_1)-d(N,K_1)>0$, $d(I,K_1)-d(M,K_1)<0$
and $p(I,M) \geq p(I,N)$. Since $I \models R$ and $M,N \not\models R$,
Property~\ref{greater-distance} applies: $p(I,M)<\frac{w_1}{w_2}<p(I,N)$,
contradicting $p(I,M) \geq p(I,N)$.~\qed

Lemma~\ref{weighted-obtainable} expresses obtainability in terms of a
universally quantified condition containing $d(I,K_i)$. If determining such a
value is polynomial, the problem is in \conp. Two cases where this happens are:

\begin{itemize}

\item $d$ is the drastic distance;

\item $d$ is the Hamming distance and both $K_1$ and $K_2$ are conjunctions of
literals.

\end{itemize}

If $d(I,K_i)$ is not polynomial to be determined, complexity increases. For the
Hamming distance $d(I,K_i)$ is the minimal number of literals that differ from
$I$ and a model of $K_i$. Obtainability amounts to:

\[
\forall I,J,\ldots ~
\forall d_I^1,d_I^2,d_J^1,\ldots ~
\left(
\begin{array}{l}
(\exists I' \models K_1 ~.~ d(I,I') \leq d_I^1) \wedge \\
(\forall I'' \models K_1 ~.~ d(I,I'') \geq d_I^1)
\end{array}
\right)
\wedge \cdots
\rightarrow
\mbox{ (conditions in Lemma~\ref{weighted-obtainable}) }
\]

Since the quantifiers $\exists I'$ and $\forall I''$ are inside the premise of
an implication, they are negated. However, they are still two independent
quantifiers. Therefore, this is a $\forall\exists QBF$, which proves that
obtainability is in \P{2}. The same happens if checking $d(I,K_i) \leq x$ is in
\np\  or in \conp. More generally, the complexity of obtainability is one level
over the complexity of calculating the distance between a model and a knowledge
base.

\begin{theorem}
\label{weighted-membership}

If determining $d(I,K) \leq x$ is in the complexity class \P{i} or \S{i}, then
obtainability of a satisfiable formula from two formulae with a weighted sum of
distances is in \P{i+1}.

\end{theorem}

\proof By Lemma~\ref{weighted-obtainable}, obtainability can be expressed as
formula with some universal quantifiers in the front $\forall I,J,L,M,N$ and a
formula $F$ containing $d(I,K_1$), $d(I,K_2)$, $d(J,K_1)$, etc. Equivalently:

\begin{eqnarray*}
\lefteqn{
\forall I,J,L,M,N ~
\forall d_I^1,d_I^2,d_J^1,d_J^2,\ldots 
} \\
&& (d(I,K_1) \leq d_I^1) \wedge \\
&& \neg (d(I,K_1) \leq d_I^1-1) \wedge \\
&& (d(J,K_1) \leq d_J^1) \wedge \\
&& \neg (d(J,K_1) \leq d_J^1-1) \wedge \\
&& \vdots \\
&& \rightarrow
F[d(I,K_1)/d_I^1,d(I,K_2)/d_I^2,d(J,K_1)/d_J^1,d(J,K_2)/d_J^2,\ldots]
\end{eqnarray*}

If $d$ can be calculated in polynomial time, the whole problem is in \conp.
Otherwise, subformulae $d(I,K_1) \leq d_I^1$ occur in the premise of an
implication, so they are in fact negated. However, if each is in \P{i} or in
\S{i}, they can be expressed as an alternation of $i$ quantifiers. The whole
problem, with the universal quantifier in the front, is therefore in
\P{i+1}.~\qed

This theorem implies the three ad-hoc complexity results obtained above: that
obtainability is in \conp\  for the drastic distance and for the Hamming
distance when the knowledge bases are conjunctions of literals, and is in \P{2}
in the general case for the Hamming distance. A general hardness result can be
given from some assumptions about the distance function.

A pseudodistance is a function such that $d(I,J)=d(J,I)$, $d(I,I)=0$ and
$d(I,J)>0$ for every $J \not=I$. Its extension to a distance from a knowledge
base obeys: $d(I,K)=0$ if $I \models K$ and $d(I,K)>0$ otherwise. If $K_1$ and
$K_2$ have some common models, these have weighted distance $0$ regardless of
the weights. Since merging selects minimal models, in this case the result
comprises exactly the common models. In particular, if $K_1$ and $K_2$
coincide, merge produces a formula equivalent to them. This holds for every
pseudodistance, and can be used to prove that obtainability is \conp-hard for
every pseudodistance.

\begin{theorem}

Obtainability of a consistent formula from two knowledge bases is \conp-hard
for every pseudodistance.

\end{theorem}

\proof The claim is proved by reduction from propositional unsatisfiability.
Let $F$ be a propositional formula. The corresponding obtainability problem is
defined by $K_1=K_2=y$ and $R=y \vee F$, where $y$ is a variable not in $F$.
Since $K_1$ and $K_2$ coincide, the result of merging is $y$. If $F$ is
satisfied by a model $I$ then $R$ has a model $I \cup \{\neg y\}$ that does not
satisfy $y$. Vice versa, if $F$ is unsatisfiable then $R$ coincides with
$y$.~\qed

Since obtainability for drastic distance and Hamming distance from conjunctions
of literals is in \conp, and these are pseudodistances, obtainability using
them is \conp\ complete.

 %

\subsection{Weighted sum of Hamming distance}

The problem of obtainability with the Hamming distance is \P{2}-hard. This is
proved by reduction from the problem of establishing the validity of a formula
$\forall X \exists Y . F$. The translation is based on two main ideas:

\begin{enumerate}

\item separate models having different evaluations of $X$ by a large distance;

\item for each evaluation of $X$, $K_1$ and $R$ contain the subformula $Y^\neg
\wedge Y'^\neg$ that sets all variables in $Y$ and a copy of it $Y'$ to false;
$K_2$ instead contains $F \wedge (Y \not\equiv Y')$.

\end{enumerate}

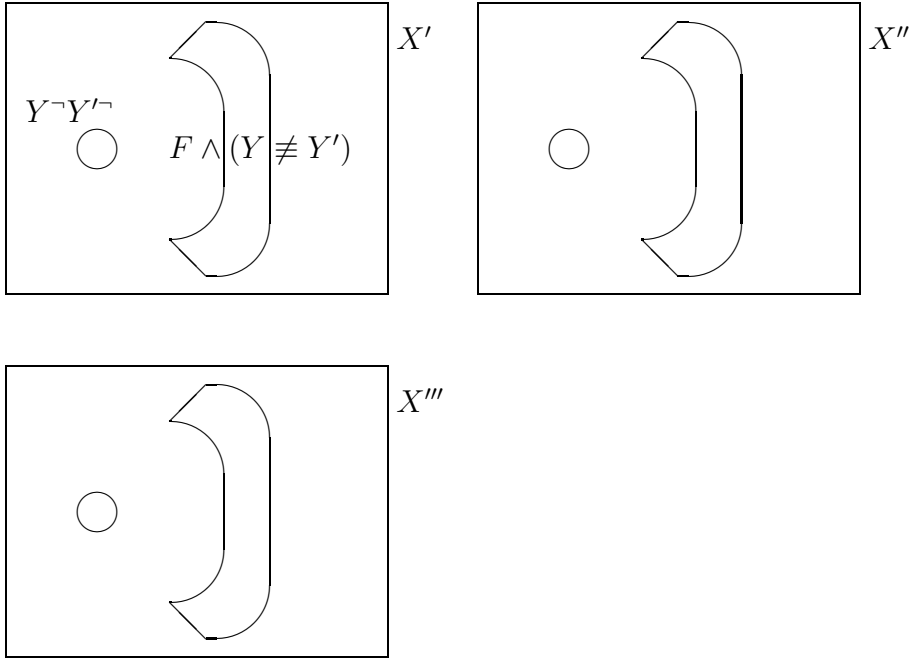
\setlength{\unitlength}{5000sp}%
\begingroup\makeatletter\ifx\SetFigFont\undefined%
\gdef\SetFigFont#1#2#3#4#5{%
  \reset@font\fontsize{#1}{#2pt}%
  \fontfamily{#3}\fontseries{#4}\fontshape{#5}%
  \selectfont}%
\fi\endgroup%
\begin{picture}(4302,3264)(259,-2593)
\thinlines
{\color[rgb]{0,0,0}\put(1081,-61){\oval(540,900)[br]}
\put(1081,-61){\oval(540,900)[tr]}
}%
{\color[rgb]{0,0,0}\put(1261,-61){\oval(630,1260)[br]}
\put(1261,-61){\oval(630,1260)[tr]}
}%
{\color[rgb]{0,0,0}\put(3421,-61){\oval(540,900)[br]}
\put(3421,-61){\oval(540,900)[tr]}
}%
{\color[rgb]{0,0,0}\put(3601,-61){\oval(630,1260)[br]}
\put(3601,-61){\oval(630,1260)[tr]}
}%
{\color[rgb]{0,0,0}\put(1081,-1861){\oval(540,900)[br]}
\put(1081,-1861){\oval(540,900)[tr]}
}%
{\color[rgb]{0,0,0}\put(1261,-1861){\oval(630,1260)[br]}
\put(1261,-1861){\oval(630,1260)[tr]}
}%
{\color[rgb]{0,0,0}\put(721,-61){\circle{202}}
}%
{\color[rgb]{0,0,0}\put(3061,-61){\circle{202}}
}%
{\color[rgb]{0,0,0}\put(721,-1861){\circle{202}}
}%
{\color[rgb]{0,0,0}\put(1081,389){\line( 1, 1){180}}
}%
{\color[rgb]{0,0,0}\put(1081,-511){\line( 1,-1){180}}
}%
{\color[rgb]{0,0,0}\put(271,-781){\framebox(1890,1440){}}
}%
{\color[rgb]{0,0,0}\put(3421,389){\line( 1, 1){180}}
}%
{\color[rgb]{0,0,0}\put(3421,-511){\line( 1,-1){180}}
}%
{\color[rgb]{0,0,0}\put(2611,-781){\framebox(1890,1440){}}
}%
{\color[rgb]{0,0,0}\put(1081,-1411){\line( 1, 1){180}}
}%
{\color[rgb]{0,0,0}\put(1081,-2311){\line( 1,-1){180}}
}%
{\color[rgb]{0,0,0}\put(271,-2581){\framebox(1890,1440){}}
}%
\put(586, 74){\makebox(0,0)[b]{\smash{{\SetFigFont{12}{24.0}{\rmdefault}{\mddefault}{\itdefault}{\color[rgb]{0,0,0}$Y^\neg Y'^\neg$}%
}}}}
\put(1531,-106){\makebox(0,0)[b]{\smash{{\SetFigFont{12}{24.0}{\rmdefault}{\mddefault}{\itdefault}{\color[rgb]{0,0,0}$F \wedge (Y \not\equiv Y')$}%
}}}}
\put(2206,434){\makebox(0,0)[lb]{\smash{{\SetFigFont{12}{24.0}{\rmdefault}{\mddefault}{\itdefault}{\color[rgb]{0,0,0}$X'$}%
}}}}
\put(2206,-1366){\makebox(0,0)[lb]{\smash{{\SetFigFont{12}{24.0}{\rmdefault}{\mddefault}{\itdefault}{\color[rgb]{0,0,0}$X'''$}%
}}}}
\put(4546,434){\makebox(0,0)[lb]{\smash{{\SetFigFont{12}{24.0}{\rmdefault}{\mddefault}{\itdefault}{\color[rgb]{0,0,0}$X''$}%
}}}}
\end{picture}%
 %
\nop
{
                                                           .
+-----------------+           +-----------------+
|            O    | X'        |            O    | X''
|  O          O   |           |  O          O   |
| Y-Y'-      O    |           | Y-Y'-      O    |
|         F(Y!=Y')|           |         F(Y!=Y')|
+-----------------+           +-----------------+
                                                           .
+-----------------+
|            O    | X'''
|  O          O   |
| Y-Y'-      O    |
|         F(Y!=Y')|
+-----------------+
                                                           .
}

The second property makes the model of $K_1$ being at distance $n$ from $K_2$,
but only if $R$ is satisfiable, and such models are in the result of merging
with $w_1>>w_2$. Formal proof follows.

\begin{theorem}

Obtainability with the weighted sum of Hamming distance from two knowledge
bases is \P{2}-complete.

\end{theorem}

\proof Membership follows from Theorem~\ref{weighted-membership}, since
checking $d(I,K) \leq x$ is in \np\  for the Hamming distance. Indeed, $d(I,K)
\leq x$ holds if there exists $J \models K$ such that $d(I,J) \leq x$, and the
distance between two models can be determined in polynomial time.

Hardness is proved by reduction from the problem $\forall \exists QBF$.

First, the problem of checking the validity of $\forall X \exists Y . F$
remains hard even if $F$ is known to be satisfiable. This is proved by
reduction from the problem without the restriction: $\forall X \exists Y . G$
is valid if and only if $\forall z \forall X \exists Y . G \vee z$ is valid,
where $z$ is a new variable: indeed, this formula is equivalent to $(\forall X
\exists Y . G \vee \top) \wedge (\forall X \exists Y . G \vee \bot)$; the first
part of this conjunction is tautological, the second is equivalent to the
original QBF.

Second, the problem of checking the validity of $\forall X \exists Y . F$ with
$F$ satisfiable is reduced to obtainability. Let $n=|X|=|Y|$ and
$Y',X_1,\ldots,X_{2n}$ be each a set of $n$ new variables.

\begin{eqnarray*}
K_1 &=& (X \equiv X_1 \equiv \cdots \equiv X_{2n}) \wedge
Y^\neg \wedge Y'^\neg
								\\
K_2 &=& (X \equiv X_1 \equiv \cdots \equiv X_{2n}) \wedge
(Y \not\equiv Y') \wedge F
								\\
R &=& K_1
\end{eqnarray*}

That the reduction works is proved in four steps:

\begin{enumerate}

\item there are models with distance vector $(0,n)$ or less;

\item the distance between models of $K_1$ or $K_2$ differing in the evaluation
of $X$ is $2n$ or more;

\item no model hsa distance vector $(0,k)$ with $k<n$;

\item a model of $K_1$ is in the result of merging if and only if its
evaluation of $X$ satisfies $F$ with some values of $Y$.

\end{enumerate}

Formula $F$ is by assumption satisfiable. Let $I$ be a model of it, and $I^X$
and $I^Y$ its parts on $X$ and $Y$, respectively. Replicating the values of
$I^X$ on $X_1,\ldots,X_{2n}$ and adding $Y^\neg$ and $Y'^\neg$ results in a
model of $K_1$. The same values of $X,X_1,\ldots,X_{2n}$ with $I^Y$ and its
negations in $Y'$ form a model of $K_2$. If $I^Y$ has $k$ positive literals
then its negated interpretation on $Y'$ has $n-k$. That makes $n$ positive
literals, while the model of $K_1$ has $Y^\neg$ and $Y'^\neg$. Since these
models coincide on $X,X_1,\ldots,X_{2n}$, the distance from the model of $K_1$
to $K_2$ is at most $n$. Since the first is a model of $K_1$, its distance
vector is $(0,n)$.

Since both $K_1$ and $K_2$ contain $X \equiv X_1 \equiv \cdots \equiv X_{2n}$,
if two of their models differ even on a single variable in $X$ they also differ
on all its $2n$ copies. Therefore, models of $K_1$ and $K_2$ with different
evaluations of $X$ are at least $2n$ apart.

To prove that no model is at distance less that $(0,n)$ suffices to consider
the models of $K_1$, since these are the only ones with $0$ in the first
position of the distance vector. Let $I$ be a model of $K_1$. By the previous
property, models of $K_2$ with a different evaluation of $X$ are at distance
$2n$ or more. The models with the same evaluation of $X$ differ only on the
values of $Y$. However, since $K_2$ contains $Y \not\equiv Y'$, all models of
$K_2$ have exactly $n$ positive literals in $Y \cup Y'$. Since $K_1$ contains
$Y^\neg$ and $Y'^\neg$, its models have all negative $Y \cup Y'$. As a result,
the distance between these models is $n$, leading to a distance vector $(0,n)$.

Since there are models with distance vector $(0,n)$, and none at distance
$(0,k)$ with $k<n$, a model of $K_1$ can be in the merge result only if it is
at distance $n$ from a model of $K_2$. Every evaluation of $X$ satisfies $K_1$
with the same values copied to $X_1,\ldots,X_{2n}$ and $Y \cup Y'$ all set to
false, and vice versa. Such model $I$ is at distance $2n$ or more from models
of $K_2$ with a different evaluation of $X$, which are therefore irrelevant to
the presence of $I$ in the result of merging: only the models of $K_2$ with the
same evaluation over $X$ matter. Such a model exists if and only if $F$ is
satisfiable for that evaluation of $X$. Moreover, $Y \not\equiv Y'$ forces
every such model at distance $n$ from $I$, making the model in the result of
merging with weights $(n+1,1)$.

This was the fourth step of the proof. Since a model of $K_1$ corresponds to an
evaluation over $X$ (and vice versa), and such a model can be in the result of
merging if and only if $F$ is consistent with that evaluation of $X$, the
whole $R$ is the result of merging if and only if $\forall X \exists Y . F$ is
valid.~\qed

 %

\subsection{Local search algorithm}
\label{search}

An algorithm using local search is shown. It employs two elements of the proof
of Lemma~\ref{weighted-obtainable} to obtain $\frac{w_1}{w_2}$ or some bounds
on its value. No assumption is made over $d(I,K)$ other than the availability
of a procedure to determine it; in the case of drastic distance this is
straightforward, as it amounts to check whether $I \models K$; for the Hamming
distance, since the problem is \np-complete, an approximate method can be used
instead. Once $\frac{w_1}{w_2}$ is determined, the knowledge bases are merged
and the result checked for equivalence to $R$. This final check is necessary
because the procedure to find models that constraint $\frac{w_1}{w_2}$ is
incomplete: not all models of $R$ and of $\neg R$ are checked.

Property~\ref{equal-distance} ensures that if two models of $R$ are such that
the denominator of $p(I,J)$ is not null, then $\frac{w_1}{w_2}=p(I,J)$. Two
such models can be looked upon using local search. During the run of the
procedure, models that do not satisfy $R$ are used to establish or refine
bounds on the value of $\frac{w_1}{w_2}$. This is useful because, as
Property~\ref{greater-distance} shows, even if for all pairs of models of $R$
the denominator of $p(I,J)$ is zero, the models that do not satisfy $R$ still
constrain $\frac{w_1}{w_2}$.

Summing up, local search search does two things at the same time:

\begin{enumerate}

\item looks for two models $I$ and $J$ of $R$ such that $p(I,J)$ has a non-zero
denominator;

\item if a model $I$ of $R$ has been found, for every model $M$ of $\neg R$
found during the search $p(I,M)$ is calculated and used to refine two bounds.

\end{enumerate}

In the following algorithm, conditions involving $I$ are to be considered false
if $I$ is unassigned, for example when the algorithm starts. The result is
$\frac{w_1}{w_2}$ or the special value ``unobtainable''; the first is assumed
to be returned as a pair of integers, rather than a (possibly truncated)
rational value. The maximal distance between two models is denoted by $n$; this
is $1$ for the drastic distance and the number of variables for the Hamming
distance. This is also the maximal value of $p(I,J)$ and the reason why $a$ is
initialized to $n+1$.

\begin{enumerate}

\item $a=n+1$; $b=-n-1$

\item $iter=0$

\item if $iter\%restart=0$ set $O$=random model
\label{iteration}

\item change $O$ by local search for a model of $R$ (see below)
\label{localmove}

\item if $O \models R$ and $I$ is unassigned set $I=O$

\item if $O \models R$ and $p(I,O)$ has a non-zero denominator, then:
\label{certain}

\begin{itemize}

\item if $p(I,O)$ is positive and between $a$ and $b$ then return $p(I,O)$

\item otherwise return unobtainable

\end{itemize}

\item if $O \not\models R$ and $d(I,K_1)-d(O,K_1)>0$ then $a=min(a,p(I,O))$

\item if $O \not\models R$ and $d(I,K_1)-d(O,K_1)<0$ then $b=max(b,p(I,O))$

\item if $a<0$ or $a \leq b$ return unobtainable

\item $iter++$

\item if $iter<maxiter$ go to Step~\ref{iteration}

\item return $\frac{a+b}{2}$
\label{uncertain}

\end{enumerate}

Point~\ref{localmove} is a step of a local search for a model of $R$: for
example, it may change the value of the variables increasing the most the
number of clauses of $R$, when this formula is in CNF. More refined methods can
be employed, such as making random moves with a certain probability, which may
remain constant or decrease with the number of iterations.

This algorithm returns $\frac{w_1}{w_2}$ as a pair of integer numbers, which
can be used as the weights $w_1$ and $w_2$. If merging with these weights
produces $R$, then they are searched weights. Otherwise, if the value is
returned from Step~\ref{certain} then $R$ is not obtainable. If it is returned
from Step~\ref{uncertain}, then one may attempt some other value between $a$
and $b$, or run the algorithm some more.

Several variants may be considered.

\begin{enumerate}

\item Step~\ref{localmove} looks for a model of $R$, but after a number of
iterations without finding one that makes the denominator of $p(I,O)$ different
than zero, it makes sense to aim at minimizing $a$ and maximizing $b$ instead;

\item models with a distance vector strictly greater than another cannot be in
the result of merge; therefore, if they satisfy $R$ then $R$ is not obtainable;
if they are not in $R$ they can be neglected;

\item instead of returning immediately after determining $p(I,J)$ in
Step~\ref{certain}, one may proceed with local search and check whether some
other models of $R$ and of $\neg R$ satisfy the conditions of
Lemma~\ref{weighted-obtainable}.

\end{enumerate}

The algorithm is based on local search which, while not guaranteed to work in
every possible case, is known to perform well in practice~\cite{aart-lens-97}.
If weights $w_1$ and $w_2$ are found and merging with them produces $R$, then
they are the correct weights. Furthermore, with Step~\ref{localmove} changing a
single variable at time, the next models is likely to have different distance
vector from the knowledge bases, which would make the algorithm terminate.

 %

\subsection{Tractable case}

\hidden{

nascosto perche' appare troppo semplice

\noindent
DRASTIC

in generale, un modello ha vettore di distanze drastiche che puo' essere solo
(0,0) (0,1) (1,0) e (1,1); la presenza del primo esclude gli altri, il secondo
e il terzo escludono il quarto; dato che K1 e K2 si assumono consistenti, gli
unici due casi possibili sono:

\begin{enumerate}

\item K1K2 consistente, e allora il risultato e' K1K2

\item K1K2 inconsistente, e allora gli unici risultati possibili sono K1, K2 e
K1vK2

\end{enumerate}

si tratta quindi di fare un test di consistenza (K1K2) e poi verificare delle
equivalenze; la prima cosa e' polinomiale per Horn, la seconda anche, dato che
si tratta di fare horn implica clausola per ogni clausola, e horn implica
clausola e' uguale alla inconsistenza di horn+letterali

[forse e' meglio evitare, risulta troppo chiaro quanto sia facile]

\

\hrule

\

non fare lemma, nascondere il fatto che il risultato e' cosi' semplice nella
dimostrazione di polinomialita'

}

This section shows a tractable case of obtainability: the measure is the
Hamming distance, the knowledge bases are conjunctions of literals and the
expected result of merging is an Horn or Krom formula.

\hidden{

The proof essentially shows that, if $K_1$ and $K_2$ are conjunctions of
literals, their merge using the weighted sum of the Hamming distances is
either:

\begin{enumerate}

\item $K_1$; or

\item $K_2$; or

\item the conjunction of the literals that occur in a base and not negated in
the other.

\end{enumerate}

See HIDDEN for reason why.

}

\begin{theorem}

If $K_1$ and $K_2$ are conjunctions of literals, determining whether a Horn or
Krom formula $R$ is obtainable by the weighted sum of the Hamming distances is
in \p.

\end{theorem}

\proof For a model $I$ and a variable $x$, let $I \cdot x$ denote a model that
is identical to $I$ except that $x$ is assigned the value true. $I \cdot \neg
x$ is the same with value false. The first step of the proof is a property of
$d(I,K)$ when $K$ entails a literal or does not mention a variable.

\begin{description}

\item[if $K$ entails $x$ then $d(J \cdot x,K)<d(J \cdot \neg x,K)$;] since
$K_1$ entails $x$, all its models set $x$ to true; this hold in particular for
every model $J$ that is one of the closest to $I$; since $I \cdot \neg x$ and
$I \cdot x$ have the same differing literals from $J$ except for $x$, which is
positive in $J$, then $d(I \cdot x,K_1)<d(I \cdot \neg x,K_1)$; the same
property holds when $K$ entails $\neg x$;

\item[if $K$ does not contain $x$ then $d(I \cdot x,K)=d(I \cdot \neg x,K)$;]
since $K$ does not mention $x$, it it is satisfied by $J \cdot x$ if and only
if it is satisfied by $J \cdot \neg x$ for every interpretation $J$; therefore,
if $J$ is a model at a minimal distance from $I$ then $J \cdot x$ is at minimal
distance from $I \cdot x$; the same holds for $\neg x$; therefore, $d(I \cdot
x,K)=d(I \cdot \neg x, K)$.

\end{description}

The second step of the proof relates merge result to the weighted distance of
$I \cdot x$ and $I \cdot \neg x$. Both are based on merge being defined from
the set of models of minimal weighted distance.

\begin{enumerate}

\item if every model $I \cdot \neg x$ has greater weighted distance from
$\{K_1,K_2\}$ than $I \cdot x$ then the merge result implies $x$, and the same
for $\neg x$; indeed, since every model where $x$ is false is further than the
same one where $x$ is true, minimal models all have $x$ true;

\item if every model $I$ is at the same weighted distance from $\{K_1,K_2\}$
than $I \cdot x$ and $I \cdot \neg x$ then the merge result does not mention
$x$; indeed, it this is true then minimal models are symmetric with respect to
$x$ and $\neg x$; the value of $x$ is therefore irrelevant to the satisfaction
of the merge result.

\end{enumerate}

The claim can now be proved. Variables are divided in the three groups: those
mentioned neither in $K_1$ nor in $K_2$; those occurring in a base but not with
the opposite sign in the other; those occurring with opposite signs.

If neither $K_1$ nor $K_2$ mention $x$ then for every $I$ it holds $d(I \cdot
x,K_1)=d(I \cdot \neg x,K_1)$ and $d(I \cdot x,K_2)=d(I \cdot \neg x,K_2)$,
which imply that $I \cdot x$ and $I \cdot \neg x$ have the same weighted
distance regardless of the weights. This implies that the merge result does not
mention $x$.

If $x$ is in $K_1$ and is not mentioned in $K_2$, then $d(I \cdot x,K_1)<d(I
\cdot \neg x,K_1)$ and $d(I \cdot x,K_2)=d(I \cdot \neg x,K_2)$, which implies
that $I \cdot x$ has lower weighted distance that $I \cdot \neg x$. If $x$ is
also in $K_2$ then $d(I \cdot x,K_2)<d(I \cdot \neg x,K_2)$, and the result is
the same. In both cases, the result of the merge entails $x$.

If $K_1 \models x$ and $K_2 \models \neg x$, then $d(I \cdot x,K_1)<d(I \cdot
\neg x,K_1)$ and $d(I \cdot x,K_2)>d(I \cdot \neg x,K_2)$. The result of merge
depends on the weights. If $w_1>w_2$ then $I \cdot x$ has lower weighted
distance than $I \cdot \neg x$, proving that the merge result entails $x$. The
same holds for all other literals that are in $K_1$. In other words, if
$w_1>w_2$ then the result of merge contains all literals in $K_1$ that occur
with the opposite sign in $K_2$. The same holds in reverse if $w_1<w_2$: the
result of merge contains all literals of $K_2$. If $w_1=w_2$ then $I \cdot x$
and $I \cdot \neg x$ have the same weighted distance, proving that the result
of merge does not mention $x$.

As a result, if $w_1>w_2$ then the result of merge contains not only the
literals that are in $K_1$ and do not occur negated in $K_2$, but also the ones
that occur negated in $K_2$. The contrary happens if $w_1<w_2$. If $w_1=w_2$
then the result of merge does not contain the variables with opposite sign in
$K_1$ and $K_2$. Each of these three possible results can be checked for
equivalence with $R$ in polynomial time because of the Horn or Krom
restriction.~\qed

 %



\section{Priority base merging}

Priority base merging~\cite{nebe-92,nebe-98,rott-93,delg-dubo-lang-06} is a
semantics that selects groups on formulae based on a {\rm priority ordering}
over them. Such an ordering over the knowledge bases $K_1,\ldots,K_m$ can be
defined as a partition $P$ of them (this representation is similar to the one
used by Rott~\cite{rott-93} for orderings over formulae); the classes of the
partition are denoted $P(1), P(2), P(3), \ldots$ and are not empty. The lower
the class $K_i$ belongs to, the higher its reliability is. Such a partition
allows comparing two sets of formulae: $L \equiv N$ if and only if $L$ and $N$
are equal; $L < N$ if and only if $P(1) \cap L = P(1) \cap N$, \ldots $P(i-1)
\cap L = P(i-1) \cap N$ and $P(i) \cap L \supset P(i) \cap N$ for some number
$i$, possibly $1$.

The {\rm maxsets} of a set of formulae $K_1,\ldots,K_m$ are its maximally
consistent subsets. Formally, $M$ is a maxset of $K_1,\ldots,K_m$ if $M$ is
consistent, $M \subseteq \{K_1,\ldots,K_m\}$ and $M \cup \{K_i\}$ is
inconsistent for every $K_i \in \{K_1,\ldots,K_m\} \backslash M$. Maxsets can
be recast in terms of base remainder
sets~\cite{alch-gard-maki-85,boot-etal-11}.

Merging $K_1,\ldots,K_m$ according to a priority ordering is disjoining the
maxsets that are minimal according to the
ordering~\cite{nebe-92,nebe-98,rott-93,delg-dubo-lang-06}. This is equivalent
to disjoining the minimal consistent subsets, including the non-maximal ones.

By definition, the result of merging is always an or-of-maxsets. However, not
all possible or-of-maxsets are produced by merging: some are not generated by
any priority partition. Given an or-of-maxsets of $K_1,\ldots,K_m$, the maxsets
it contains are called {\em selected}, the others {\em excluded}. The aim is to
find an ordering, if any, that makes the selected maxsets minimal and the other
ones non-minimal.

A formula $R$ is {\em obtainable} from $K_1,\ldots,K_m$ if it can be obtained
by merging these formulae. For the merging based on priority orderings, this
amount to checking the existence of an ordering that makes the result of
merging $K_1,\ldots,K_m$ equal to $R$. This condition is equivalent to the
existence of an ordering such that the minimal maxsets are exactly the selected
ones. The difference between ``selected'' and ``minimal'' is that the first one
is a requirement (the maxset is in the expected result $R$) while the second is
a condition over a specific ordering (it makes the maxset minimal). Not all
formulae are obtainable, and this will be formally proved.

For technical reasons, obtainability is extended to pairs $(S,E)$ where both
$S$ and $E$ are sets of sets of formulae. Such a pair is obtainable if there
exists a priority ordering such that the sets in $S$ are exactly the minimal
ones among $S \cup E$. Obtainability can be defined from this concept: $R$ is
obtainable if $R \equiv \bigvee S$, $(S,E)$ is obtainable and $S \cup E$ is the
set of all maxsets of $K_1,\ldots,K_m$.

Given formulae $R$ and $K_1,\ldots,K_m$, the problem of obtainability is that
of finding (search problem) or deciding the existence of (decision problem) a
priority ordering such that $R$ is the result of merging $K_1,\ldots,K_m$ with
that ordering.

As usual, the complexity analysis is carried over the decision version of the
problem, but the algorithm in Section~\ref{algorithm} is aimed at finding the
actual priority ordering, if one exists. Otherwise, Section~\ref{unobtainable}
describes some possible courses of actions when the expected result is
unobtainable.

A number of properties related to obtainability are shown. The first ones are
about maxsets in general, the other about the specific problem of obtaining a
formula as the result of merging with an appropriate priority ordering.

\subsection{Properties of maxsets}

A general property of maxsets is that they are pairwise inconsistent. This is
quite a folklore result, and is proved here only for the sake of completeness.

\begin{lemma}
\label{inconsistent}

Two different maxsets of the same set of formulae are mutually inconsistent.

\end{lemma}

\proof To the contrary, assume that $M$ and $N$ are two differing maxsets such
that $M \cup N$ is consistent. Since $M$ and $N$ differ, either $M \backslash
N$ or $N \backslash M$ is not empty. In the first case, since $M \cup N = N
\cup (M \backslash N)$, then $N$ is consistent with other formulae not in $N$.
This contradicts the assumption that $N$ is a maxset: no formula can be be
consistently added to $N$. A similar line proves the impossibility of the other
case.~\qed

\begin{lemma}
\label{model}

If $M$ is a maxset of $K_1,\ldots,K_m$ and $I$ one of its models,
then $M=\{K_i ~|~ I \models K_i\}$.

\end{lemma}

\proof $I$ is a model of $M$ if it is a model of all formulae of $M$, that is,
the formulae of $M$ are a subsets of those satisfied by $I$. This proves that
$M \subseteq \{K_i ~|~ I\models K_i\}$. If such a containment were strict, the
formulae $K_i$ that are not in $M$ would be consistent with $M$ because they
are satisfied by $I$, contradicting the assumption that $M$ is a maxset.~\qed

When checking minimality using a priority ordering, considering all consistent
subsets or only the maxsets does not make any difference, as the following
lemma shows.

\begin{lemma}
\label{subset}

If $N \subset M$ then $M$ is less than $N$ according to every priority
ordering.

\end{lemma}

\proof If $N \subset M$ then $N \cap P(i) \subseteq M \cap P(i)$ for every $i$.
Since the containment is strict, $M \backslash N$ is not empty. Let $K_i$ be an
element of it, and $j$ its class. Containment $N \cap P(i) \subseteq M \cap
P(i)$ holds for all $i$'s, including $i=j$. For this index, however, $K_i
\not\in N \cap P(j)$ while $K_i \in M \cap P(j)$, proving that $M$ is strictly
less than $N$ according to the ordering.~\qed

As a result, minimal consistent subsets and minimal maxsets are the same. Also,
a maxset is minimal if and only if is not less than another consistent subset.

Usually, formulae to be merged are assumed to be consistent, when taken one at
time. In such cases, the following lemma helps in identifying the minimal
maxsets.

\begin{lemma}
\label{first}

For every maxset $M$ that is minimal according to priority $P$ it holds $M \cap
P(1) \not= \emptyset$.

\end{lemma}

\proof To the contrary, assume that $M \cap P(1) = \emptyset$. By definition of
priorities, $P(1)$ is not empty. Let $K$ be a formula of it. By the assumption
that all formulae are consistent, $\{K\}$ is consistent. Moreover, $P(1) \cap M
\subset P(1) \cap \{K\}$, which by definition implies $\{K\} < M$,
contradicting the assumption that $M$ is minimal.~\qed

In words, minimal maxsets have at least a formula in the first class of the
priority partition. This result depends on all formulae being consistent and no
priority class being empty, both of which are assumed in this article.

The next lemma is useful for producing maxsets with some given property. It
tells how to build formulae in such a way the maxsets are related in some way.
In particular, it involves letters $A, B, C, D, \ldots$. These are just
arbitrary symbols. Given some sets of them, such that $\{A,B\}$, $\{B, C, D\}$,
etc., one can build a formula for $A$, a formula for $B$, etc., in such a way
the maxsets of these formulae are exactly the given sets $\{A,B\}$, $\{B, C,
D\}$, etc. The only requirements is that none of these sets is contained in
another: for example, if $\{A,B\}$ is given then $\{A,B,C\}$ cannot.

\begin{lemma}
\label{synthesis}

Given some sets of letters, none of these sets contained in another, there
exists a formula for each letter so that the maxsets of these formulae
correspond to the given sets of letters.

\end{lemma}

\proof For $n$ sets, $\lceil\log n\rceil$ propositional variables are required.
Each set of letters is associated an unique propositional interpretation; this
is possible because by construction there are at least $n$ propositional
interpretations over these variables.

For each such interpretation, one can build a formula that is satisfied only by
it. For example, if the interpretation makes $x$ and $y$ false and $z$ true,
the formula is $\neg x \wedge \neg y \wedge z$. Since each set of letters is
associated a propositional interpretation, is also associated to the
corresponding formula.

If letter $L$ is in the sets $S_1,S_2,\ldots$, and these sets corresponds to
formulae $F_1,F_2,\ldots$, the formula corresponding to $L$ is their
disjunction $F_1 \vee F_2 \vee \cdots$. As a result, the formula corresponding
to the letter $L$ is satisfied exactly by the interpretations of the sets
$S_1,S_2,\ldots$.

By construction, if a set of letters is associated to the interpretation $I$,
then the formulae corresponding to the letters in the set are satisfied by $I$.
This proves that each set of letters corresponds to a consistent set of
formulae. This set is also maximally consistent because: a. no other formula is
satisfied by that interpretation; and b. if all formulae of the set plus some
others are satisfied by another interpretation, then the set corresponding to
that interpretation strictly contains the considered one, contradicting the
assumption that none of the sets strictly contains another.

To conclude the proof, the formulae do not have other maxsets. This is because
the formulae are only satisfied by some of the interpretations corresponding to
the sets of letters, and each of them is the only model of a maxset.~\qed

Intuitively, this lemma proves that letters can be used in place of formulae,
and sets of letters for their maxsets. Provided that no set is contained in
another, it is always possible to build a set of formulae to use in place of
the letters, and the sets of letters will be their maxsets. This method can be
used for example to show that maxsets may form a sort of ``cycles''. The first
step is to define the sets of letters:

\begin{enumerate}

\item $\{A,B\}$

\item $\{A,C\}$

\item $\{B,C\}$

\end{enumerate}

Binary sets can be drawn as edges of a graph, a graphical representation that
will be used also in the rest of this article:

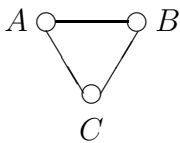
\setlength{\unitlength}{5000sp}%
\begingroup\makeatletter\ifx\SetFigFont\undefined%
\gdef\SetFigFont#1#2#3#4#5{%
  \reset@font\fontsize{#1}{#2pt}%
  \fontfamily{#3}\fontseries{#4}\fontshape{#5}%
  \selectfont}%
\fi\endgroup%
\begin{picture}(660,705)(4756,-3631)
{\color[rgb]{0,0,0}\thinlines
\put(4861,-3031){\circle{90}}
}%
{\color[rgb]{0,0,0}\put(5311,-3031){\circle{90}}
}%
{\color[rgb]{0,0,0}\put(5086,-3391){\circle{90}}
}%
{\color[rgb]{0,0,0}\put(4861,-3076){\line( 3,-5){186.618}}
}%
{\color[rgb]{0,0,0}\put(5131,-3391){\line( 3, 5){186.618}}
}%
{\color[rgb]{0,0,0}\put(4906,-3031){\line( 1, 0){360}}
}%
\put(5086,-3616){\makebox(0,0)[b]{\smash{{\SetFigFont{12}{24.0}{\rmdefault}{\mddefault}{\updefault}{\color[rgb]{0,0,0}$C$}%
}}}}
\put(4771,-3076){\makebox(0,0)[rb]{\smash{{\SetFigFont{12}{24.0}{\rmdefault}{\mddefault}{\updefault}{\color[rgb]{0,0,0}$A$}%
}}}}
\put(5401,-3076){\makebox(0,0)[lb]{\smash{{\SetFigFont{12}{24.0}{\rmdefault}{\mddefault}{\updefault}{\color[rgb]{0,0,0}$B$}%
}}}}
\end{picture}%
 %
\nop
{
A     B
O-----O
 \   /
  \ /
   O C
}

Instead of showing formulae with maxsets having the given property, the maxsets
are expressed as sets of letters, each representing a formula.
Lemma~\ref{synthesis} tells that such formulae exist, its proof how to build
them. In this case, three sets require two variables, like $x$ and $y$. The
interpretations associated to the sets can be chosen arbitrarily, for example:

\begin{itemize}

\item $\{A,B\} \Rightarrow \{x,y\}$

\item $\{A,C\} \Rightarrow \{x,\neg y\}$

\item $\{B,C\} \Rightarrow \{\neg x,y\}$

\end{itemize}

Since $A$ is in $\{A,B\}$ and in $\{A,C\}$, its formula is one satisfied by the
models of these two sets: $\{x,y\}$ and $\{x,\neg y\}$. For example, $A$ is $(x
\wedge y) \vee (x \wedge \neg y)$, which simplifies to $x$. In the same way, $B
= y$ and $C = (x \not\equiv y)$.

These formulae $x, y, x \not\equiv y$ have the required maxsets, each composed
of exactly two formulae over three. From now on, this explicit construction of
formulae from sets of letters representing their maxsets is generally not done,
with Lemma~\ref{synthesis} referenced as evidence that it is possible. This is
first done in the proof of Lemma~\ref{unobtainable}, showing that a formula
that is an or of some maxsets may not be obtainable with any ordering.

The next two lemmas show that some results are easy to obtain: selecting all
maxsets or just a single one.

\begin{lemma}
\label{all}

The priority ordering that gives maximal priority to all formulae makes all
maxsets minimal.

\end{lemma}

\proof A maxset $M$ could be non-minimal only if there exist another maxset $N$
such that $N<M$. If all formulae are in $P(1)$, the definition of ordering of
maxsets simplifies to: $N<M$ if $M \subset N$. This contradicts the assumption
that $M$ is maximally consistent.~\qed

\begin{lemma}
\label{one}

The priority ordering that gives maximal priority to exactly the formulae of a
maxset makes it the only minimal one.

\end{lemma}

\proof By contradiction, if $M$ is not minimal then $N<M$ for some other maxset
$N$. This implies either $P(1) \cap M \subseteq P(1) \cap N$ or $P(1) \cap M
\subset P(1) \cap N$. The latter contradicts $P(1)=M$. The former implies $M
\subseteq P(1) \cap N$, which is only possible if $M=N$ or $M \subset N$, and a
maxset is never contained in another.~\qed

\subsection{Properties of obtainability}

The following lemma expresses equivalent conditions for a maxset to be a
disjunct of the result of merging.

\begin{lemma}
\label{selected}

If $R$ is obtainable by priority base merging from some formulae and $M$ is a
maxset of them, the following conditions are equivalent:

\begin{itemize}

\item $M$ is consistent with $R$;

\item $M \models R$;

\item $M$ is selected in all orderings that generate $R$.

\end{itemize}

\end{lemma}

\proof Since the maxsets are mutually inconsistent by Lemma~\ref{inconsistent},
each model of $R$ is contained in exactly a maxset $M$. Therefore, $M$ is one
of the disjuncts that form $R$ if and only if it is consistent with $R$, and
this holds in every ordering that generate $R$.~\qed

By definition, merging produces a disjunction of some of the maxsets, the
minimal ones according to the priority ordering. A first question is whether
all disjunctions of maxsets are obtainable with an appropriate ordering. The
following lemma shows that the answer is no.

The counterexample uses four maxsets, of which two are selected and two
excluded. ``Selected'' and ``excluded'' indicates whether a maxset is in the
disjunction that is the expected result of merging. In other words, the
required ordering has the selected maxsets as the minimal ones. If maxsets are
binary, they can be depicted as a graph, where a crossed edge represents an
excluded maxset:

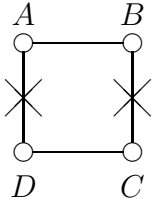
\setlength{\unitlength}{5000sp}%
\begingroup\makeatletter\ifx\SetFigFont\undefined%
\gdef\SetFigFont#1#2#3#4#5{%
  \reset@font\fontsize{#1}{#2pt}%
  \fontfamily{#3}\fontseries{#4}\fontshape{#5}%
  \selectfont}%
\fi\endgroup%
\begin{picture}(744,1020)(5029,-3991)
{\color[rgb]{0,0,0}\thinlines
\put(5131,-3211){\circle{90}}
}%
{\color[rgb]{0,0,0}\put(5671,-3211){\circle{90}}
}%
{\color[rgb]{0,0,0}\put(5131,-3751){\circle{90}}
}%
{\color[rgb]{0,0,0}\put(5671,-3751){\circle{90}}
}%
{\color[rgb]{0,0,0}\put(5176,-3211){\line( 1, 0){450}}
}%
{\color[rgb]{0,0,0}\put(5131,-3256){\line( 0,-1){450}}
}%
{\color[rgb]{0,0,0}\put(5176,-3751){\line( 1, 0){450}}
}%
{\color[rgb]{0,0,0}\put(5671,-3256){\line( 0,-1){450}}
}%
{\color[rgb]{0,0,0}\put(5581,-3391){\line( 1,-1){180}}
}%
{\color[rgb]{0,0,0}\put(5581,-3571){\line( 1, 1){180}}
}%
{\color[rgb]{0,0,0}\put(5041,-3571){\line( 1, 1){180}}
}%
{\color[rgb]{0,0,0}\put(5041,-3391){\line( 1,-1){180}}
}%
\put(5131,-3121){\makebox(0,0)[b]{\smash{{\SetFigFont{12}{24.0}{\rmdefault}{\mddefault}{\updefault}{\color[rgb]{0,0,0}$A$}%
}}}}
\put(5671,-3121){\makebox(0,0)[b]{\smash{{\SetFigFont{12}{24.0}{\rmdefault}{\mddefault}{\updefault}{\color[rgb]{0,0,0}$B$}%
}}}}
\put(5671,-3976){\makebox(0,0)[b]{\smash{{\SetFigFont{12}{24.0}{\rmdefault}{\mddefault}{\updefault}{\color[rgb]{0,0,0}$C$}%
}}}}
\put(5131,-3976){\makebox(0,0)[b]{\smash{{\SetFigFont{12}{24.0}{\rmdefault}{\mddefault}{\updefault}{\color[rgb]{0,0,0}$D$}%
}}}}
\end{picture}%
 %
\nop
{
A O-----O B
  |     |
  X     X
  |     |
D O-----O C
}

\begin{lemma}
\label{unobtainable}

No priority ordering selects $\{A,B\}$ and $\{C,D\}$ while excluding $\{B,C\}$
and $\{D,A\}$.

\end{lemma}

\proof By Lemma~\ref{synthesis}, letters and sets of letters can be used in
place of formulae and their maxsets, respectively. The following maxsets are
proved not be obtained by any ordering:

\begin{enumerate}

\item $\{A,B\}$ selected
\item $\{B,C\}$ excluded
\item $\{C,D\}$ selected
\item $\{D,A\}$ excluded

\end{enumerate}

In words, no priority ordering makes the first and third maxsets minimal out of
these four.

To the contrary, assume that such an ordering exists. By Lemma~\ref{first},
since $\{A,B\}$ is selected, either $A$ or $B$ is in the first class of the
priority partition. For the same reason, either $C$ or $D$ is.

The first class cannot include both $A$ and $D$, as otherwise $\{A,D\}$ would
be minimal. For the same reason, it cannot include both $B$ and $C$, since
$\{B,C\}$ is excluded. The only remaining cases are $A$ and $C$ in the first
class, or $B$ and $D$. The second case is omitted by symmetry: it is the same
as the first swapping $A$ with $B$ and $C$ with $D$.

In the first case, $B$ and $D$ are not in the first class of the priority
partition. Since both $\{A,B\}$ and $\{C,D\}$ are selected, if one of them is
not in the second class either, so is the other. Since classes cannot be empty,
$B$ and $D$ are in the second class:

\begin{tabular}{ccc}
A & C \\
\hline
B & D
\end{tabular}

This ordering selects $\{A,B\}$ and $\{C,D\}$ as required, but also $\{B,C\}$.
This contradicts the assumption that $\{B,C\}$ is excluded.~\qed

By Lemma~\ref{synthesis}, letters $A, B, C, D$ can be replaced by formulae in
such a way the four sets in the lemma represent their maxsets. The
impossibility of selecting the first and third while excluding the second and
fourth proves that the or of the first and third maxsets is not obtainable.

\begin{corollary}

There exists $R$ and $K_1,\ldots,K_m$ such that $R$ is the disjunction of some
of the maxsets of $K_1,\ldots,K_m$ but is not obtainable by priority base
merging.

\end{corollary}

An application of Lemma~\ref{synthesis} allows finding the actual formulae to
use in place of $A, B, C, D$. The unobtainable result is then $(A \wedge B)
\vee (C \wedge D)$. Formulae like these are later used as the basis of an
hardness result.

The maxsets of this lemma form a cycle in which selected and excluded maxsets
alternates. This condition is shown to be necessary and sufficient in the case
of maxsets comprising two formulae or less.

The counterexample involves four formulae and four maxsets. This is the minimal
condition for unobtainability: a result that is an or-of-maxsets is always
obtainable if the formulae to be merged are three or less.

\begin{theorem}
\label{three}

Every consistent or-of-maxsets is obtainable by priority base merging if the
maxsets are less than four.

\end{theorem}

\proof If a set of formulae has a single maxset, the only possible result of
merge is the maxset itself, which is therefore always obtainable. With two
maxsets, only two cases are possible: select one of them, or both.
Lemma~\ref{one} and Lemma~\ref{all} cover both cases.

With three maxsets, these lemmas proves that selecting one or all of them is
always possible. The only remaining case is that of two selected maxsets out of
three. Let them be $M$, $N$, and $L$, where the first two are selected. Being
maxsets, $M$ has a formula not in $L$, and the same for $N$:

\begin{itemize}

\item $M \backslash L \not= \emptyset$

\item $N \backslash L \not= \emptyset$

\end{itemize}

If $M \backslash L$ and $N \backslash L$ intersect, place this intersection in
$P(1)$ and all other formulae in $P(2)$. This way, $M$ and $N$ have the same
formulae in $P(1)$ while $L$ has none, proving that $M$ and $N$ are selected
while $L$ is not.
 
If $M \backslash L$ and $N \backslash L$ do not intersect, place their union in
$P(1)$ and all other formulae in $P(2)$. This ordering guarantees that both $M$
and $N$ have formulae in $P(1)$ while $L$ has none, and that $P(1) \cap M$ and
$P(1) \cap N$ are not contained one in the other.~\qed

Since three formulae have at most three maxsets, this theorems proves that
every consistent or-of-maxsets of three formulae is obtainable with an
appropriate priority ordering.

Lemma~\ref{unobtainable} uses four formulae, indeed: $\{A,B\}$, $\{B,C\}$,
$\{C,D\}$, $\{D,A\}$. The disjunction of the first three of these maxsets is
also unobtainable: this can be proved in the same line as
Lemma~\ref{unobtainable}, and shows a case where all maxsets but one are
unobtainable. In contrast, Lemma~\ref{all} and Lemma~\ref{one} state that a
single maxset and all maxsets are always obtainable.

The four maxsets form a cycle, when seen as a graph: $\{A,B\}, \{B,C\},
\{C,D\}, \{D,A\}$. When considering maxsets comprising more than two elements,
the notion of Berge--acyclicity~\cite{fagi-83} for hypergraphs ensure
obtainability, as the next theorem shows.

\begin{theorem}
\label{acyclic}

Every disjunction of a nonempty subset of a set of maxsets that is
Berge--acyclic is obtainable by priority base merging.

\end{theorem}

\proof A set of sets that is Berge-acyclic can be seen as a tree of sets, where
each set shares a single node with its parent and one with each of its
children. A priority ordering can be build starting from a maxset, labeling its
formulae and then moving to its children.

At each step, a set having a single labeled node is considered, and the
labeling is extended to its other nodes. A label is either a single number $n$
greater than one or a pair $1,n$ with $n$ greater than one. The meaning of
$1,n$ will be clarified later, but it roughly means that the node is part of a
selected maxset whose other nodes are labeled $n$.

The procedure includes some choices, such as the root and a node in each set.
It is however not nondeterministic, as it works for any of these choices; in
other words, every choice can be resolved by taking arbitrary choices.

The procedure starts from the root. If this maxset is selected, an arbitrary
one of its nodes is labeled $1,2$:

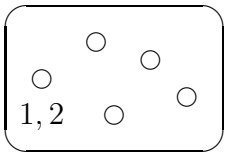
\setlength{\unitlength}{5000sp}%
\begingroup\makeatletter\ifx\SetFigFont\undefined%
\gdef\SetFigFont#1#2#3#4#5{%
  \reset@font\fontsize{#1}{#2pt}%
  \fontfamily{#3}\fontseries{#4}\fontshape{#5}%
  \selectfont}%
\fi\endgroup%
\begin{picture}(1104,744)(4669,-3853)
{\color[rgb]{0,0,0}\thinlines
\put(4861,-3481){\circle{90}}
}%
{\color[rgb]{0,0,0}\put(5131,-3301){\circle{90}}
}%
{\color[rgb]{0,0,0}\put(5221,-3661){\circle{90}}
}%
{\color[rgb]{0,0,0}\put(5401,-3391){\circle{90}}
}%
{\color[rgb]{0,0,0}\put(5581,-3571){\circle{90}}
}%
{\color[rgb]{0,0,0}\put(4786,-3736){\oval(210,210)[bl]}
\put(4786,-3226){\oval(210,210)[tl]}
\put(5656,-3736){\oval(210,210)[br]}
\put(5656,-3226){\oval(210,210)[tr]}
\put(4786,-3841){\line( 1, 0){870}}
\put(4786,-3121){\line( 1, 0){870}}
\put(4681,-3736){\line( 0, 1){510}}
\put(5761,-3736){\line( 0, 1){510}}
}%
\put(4861,-3706){\makebox(0,0)[b]{\smash{{\SetFigFont{12}{24.0}{\rmdefault}{\mddefault}{\updefault}{\color[rgb]{0,0,0}$1,2$}%
}}}}
\end{picture}%
 %
\nop
{
}

If it is excluded, an arbitrary one of its nodes is labeled $2$:

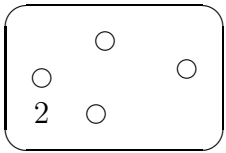
\setlength{\unitlength}{5000sp}%
\begingroup\makeatletter\ifx\SetFigFont\undefined%
\gdef\SetFigFont#1#2#3#4#5{%
  \reset@font\fontsize{#1}{#2pt}%
  \fontfamily{#3}\fontseries{#4}\fontshape{#5}%
  \selectfont}%
\fi\endgroup%
\begin{picture}(1104,744)(4669,-3853)
{\color[rgb]{0,0,0}\thinlines
\put(4861,-3481){\circle{90}}
}%
{\color[rgb]{0,0,0}\put(5176,-3301){\circle{90}}
}%
{\color[rgb]{0,0,0}\put(5581,-3436){\circle{90}}
}%
{\color[rgb]{0,0,0}\put(5131,-3661){\circle{90}}
}%
{\color[rgb]{0,0,0}\put(4786,-3736){\oval(210,210)[bl]}
\put(4786,-3226){\oval(210,210)[tl]}
\put(5656,-3736){\oval(210,210)[br]}
\put(5656,-3226){\oval(210,210)[tr]}
\put(4786,-3841){\line( 1, 0){870}}
\put(4786,-3121){\line( 1, 0){870}}
\put(4681,-3736){\line( 0, 1){510}}
\put(5761,-3736){\line( 0, 1){510}}
}%
\put(4861,-3706){\makebox(0,0)[b]{\smash{{\SetFigFont{12}{24.0}{\rmdefault}{\mddefault}{\updefault}{\color[rgb]{0,0,0}$2$}%
}}}}
\end{picture}%
 %
\nop
{
}

The algorithm descends the tree. When moving from the parent to a child, the
former is all labeled and the latter shares a single labeled node with it and
its other nodes are unlabeled. Labels are added to them, and the procedure
moves to the children.

Labels are added to selected edges are follows:

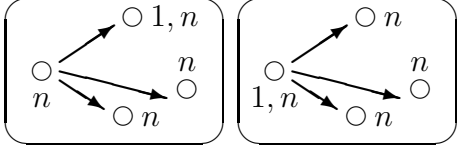
\setlength{\unitlength}{5000sp}%
\begingroup\makeatletter\ifx\SetFigFont\undefined%
\gdef\SetFigFont#1#2#3#4#5{%
  \reset@font\fontsize{#1}{#2pt}%
  \fontfamily{#3}\fontseries{#4}\fontshape{#5}%
  \selectfont}%
\fi\endgroup%
\begin{picture}(1104,744)(4669,-3853)
{\color[rgb]{0,0,0}\thinlines
\put(4861,-3481){\circle{90}}
}%
{\color[rgb]{0,0,0}\put(5581,-3571){\circle{90}}
}%
{\color[rgb]{0,0,0}\put(5266,-3706){\circle{90}}
}%
{\color[rgb]{0,0,0}\put(5311,-3211){\circle{90}}
}%
{\color[rgb]{0,0,0}\put(4786,-3736){\oval(210,210)[bl]}
\put(4786,-3226){\oval(210,210)[tl]}
\put(5656,-3736){\oval(210,210)[br]}
\put(5656,-3226){\oval(210,210)[tr]}
\put(4786,-3841){\line( 1, 0){870}}
\put(4786,-3121){\line( 1, 0){870}}
\put(4681,-3736){\line( 0, 1){510}}
\put(5761,-3736){\line( 0, 1){510}}
}%
\thicklines
{\color[rgb]{0,0,0}\put(4951,-3436){\vector( 3, 2){270}}
}%
{\color[rgb]{0,0,0}\put(4951,-3481){\vector( 4,-1){529.412}}
}%
{\color[rgb]{0,0,0}\put(4951,-3526){\vector( 3,-2){218.077}}
}%
\put(5581,-3481){\makebox(0,0)[b]{\smash{{\SetFigFont{12}{24.0}{\rmdefault}{\mddefault}{\updefault}{\color[rgb]{0,0,0}$n$}%
}}}}
\put(5356,-3751){\makebox(0,0)[lb]{\smash{{\SetFigFont{12}{24.0}{\rmdefault}{\mddefault}{\updefault}{\color[rgb]{0,0,0}$n$}%
}}}}
\put(4861,-3661){\makebox(0,0)[b]{\smash{{\SetFigFont{12}{24.0}{\rmdefault}{\mddefault}{\updefault}{\color[rgb]{0,0,0}$n$}%
}}}}
\put(5401,-3256){\makebox(0,0)[lb]{\smash{{\SetFigFont{12}{24.0}{\rmdefault}{\mddefault}{\updefault}{\color[rgb]{0,0,0}$1,n$}%
}}}}
\end{picture}%
 %
\nop

{}
\setlength{\unitlength}{5000sp}%
\begingroup\makeatletter\ifx\SetFigFont\undefined%
\gdef\SetFigFont#1#2#3#4#5{%
  \reset@font\fontsize{#1}{#2pt}%
  \fontfamily{#3}\fontseries{#4}\fontshape{#5}%
  \selectfont}%
\fi\endgroup%
\begin{picture}(1104,744)(4669,-3853)
{\color[rgb]{0,0,0}\thinlines
\put(4861,-3481){\circle{90}}
}%
{\color[rgb]{0,0,0}\put(5581,-3571){\circle{90}}
}%
{\color[rgb]{0,0,0}\put(5311,-3211){\circle{90}}
}%
{\color[rgb]{0,0,0}\put(5266,-3706){\circle{90}}
}%
{\color[rgb]{0,0,0}\put(4786,-3736){\oval(210,210)[bl]}
\put(4786,-3226){\oval(210,210)[tl]}
\put(5656,-3736){\oval(210,210)[br]}
\put(5656,-3226){\oval(210,210)[tr]}
\put(4786,-3841){\line( 1, 0){870}}
\put(4786,-3121){\line( 1, 0){870}}
\put(4681,-3736){\line( 0, 1){510}}
\put(5761,-3736){\line( 0, 1){510}}
}%
\thicklines
{\color[rgb]{0,0,0}\put(4951,-3436){\vector( 3, 2){270}}
}%
{\color[rgb]{0,0,0}\put(4951,-3481){\vector( 4,-1){529.412}}
}%
{\color[rgb]{0,0,0}\put(4951,-3526){\vector( 3,-2){218.077}}
}%
\put(5581,-3481){\makebox(0,0)[b]{\smash{{\SetFigFont{12}{24.0}{\rmdefault}{\mddefault}{\updefault}{\color[rgb]{0,0,0}$n$}%
}}}}
\put(5401,-3256){\makebox(0,0)[lb]{\smash{{\SetFigFont{12}{24.0}{\rmdefault}{\mddefault}{\updefault}{\color[rgb]{0,0,0}$n$}%
}}}}
\put(5356,-3751){\makebox(0,0)[lb]{\smash{{\SetFigFont{12}{24.0}{\rmdefault}{\mddefault}{\updefault}{\color[rgb]{0,0,0}$n$}%
}}}}
\put(4861,-3661){\makebox(0,0)[b]{\smash{{\SetFigFont{12}{24.0}{\rmdefault}{\mddefault}{\updefault}{\color[rgb]{0,0,0}$1,n$}%
}}}}
\end{picture}%
 %
\nop

{}

In words, if the only label is $n$, an arbitrary one of the others is labeled
$1,n$ and the remaining (if any) are labeled $n$. If the only label is $1,n$,
the others are labeled $n$.

If the considered set is excluded, labels are extended as follows:

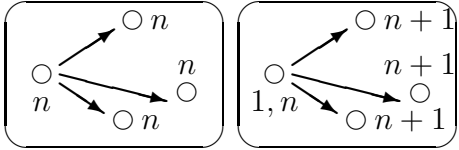
\setlength{\unitlength}{5000sp}%
\begingroup\makeatletter\ifx\SetFigFont\undefined%
\gdef\SetFigFont#1#2#3#4#5{%
  \reset@font\fontsize{#1}{#2pt}%
  \fontfamily{#3}\fontseries{#4}\fontshape{#5}%
  \selectfont}%
\fi\endgroup%
\begin{picture}(1104,744)(4669,-3853)
{\color[rgb]{0,0,0}\thinlines
\put(4861,-3481){\circle{90}}
}%
{\color[rgb]{0,0,0}\put(5581,-3571){\circle{90}}
}%
{\color[rgb]{0,0,0}\put(5311,-3211){\circle{90}}
}%
{\color[rgb]{0,0,0}\put(5266,-3706){\circle{90}}
}%
{\color[rgb]{0,0,0}\put(4786,-3736){\oval(210,210)[bl]}
\put(4786,-3226){\oval(210,210)[tl]}
\put(5656,-3736){\oval(210,210)[br]}
\put(5656,-3226){\oval(210,210)[tr]}
\put(4786,-3841){\line( 1, 0){870}}
\put(4786,-3121){\line( 1, 0){870}}
\put(4681,-3736){\line( 0, 1){510}}
\put(5761,-3736){\line( 0, 1){510}}
}%
\thicklines
{\color[rgb]{0,0,0}\put(4951,-3436){\vector( 3, 2){270}}
}%
{\color[rgb]{0,0,0}\put(4951,-3481){\vector( 4,-1){529.412}}
}%
{\color[rgb]{0,0,0}\put(4951,-3526){\vector( 3,-2){218.077}}
}%
\put(4861,-3661){\makebox(0,0)[b]{\smash{{\SetFigFont{12}{24.0}{\rmdefault}{\mddefault}{\updefault}{\color[rgb]{0,0,0}$n$}%
}}}}
\put(5401,-3256){\makebox(0,0)[lb]{\smash{{\SetFigFont{12}{24.0}{\rmdefault}{\mddefault}{\updefault}{\color[rgb]{0,0,0}$n$}%
}}}}
\put(5581,-3481){\makebox(0,0)[b]{\smash{{\SetFigFont{12}{24.0}{\rmdefault}{\mddefault}{\updefault}{\color[rgb]{0,0,0}$n$}%
}}}}
\put(5356,-3751){\makebox(0,0)[lb]{\smash{{\SetFigFont{12}{24.0}{\rmdefault}{\mddefault}{\updefault}{\color[rgb]{0,0,0}$n$}%
}}}}
\end{picture}%
 %
\nop
{
}
\setlength{\unitlength}{5000sp}%
\begingroup\makeatletter\ifx\SetFigFont\undefined%
\gdef\SetFigFont#1#2#3#4#5{%
  \reset@font\fontsize{#1}{#2pt}%
  \fontfamily{#3}\fontseries{#4}\fontshape{#5}%
  \selectfont}%
\fi\endgroup%
\begin{picture}(1104,747)(4669,-3853)
{\color[rgb]{0,0,0}\thinlines
\put(4861,-3481){\circle{90}}
}%
{\color[rgb]{0,0,0}\put(5581,-3571){\circle{90}}
}%
{\color[rgb]{0,0,0}\put(5311,-3211){\circle{90}}
}%
{\color[rgb]{0,0,0}\put(5266,-3706){\circle{90}}
}%
{\color[rgb]{0,0,0}\put(4786,-3736){\oval(210,210)[bl]}
\put(4786,-3226){\oval(210,210)[tl]}
\put(5656,-3736){\oval(210,210)[br]}
\put(5656,-3226){\oval(210,210)[tr]}
\put(4786,-3841){\line( 1, 0){870}}
\put(4786,-3121){\line( 1, 0){870}}
\put(4681,-3736){\line( 0, 1){510}}
\put(5761,-3736){\line( 0, 1){510}}
}%
\thicklines
{\color[rgb]{0,0,0}\put(4951,-3436){\vector( 3, 2){270}}
}%
{\color[rgb]{0,0,0}\put(4951,-3481){\vector( 4,-1){529.412}}
}%
{\color[rgb]{0,0,0}\put(4951,-3526){\vector( 3,-2){218.077}}
}%
\put(4861,-3661){\makebox(0,0)[b]{\smash{{\SetFigFont{12}{24.0}{\rmdefault}{\mddefault}{\updefault}{\color[rgb]{0,0,0}$1,n$}%
}}}}
\put(5401,-3256){\makebox(0,0)[lb]{\smash{{\SetFigFont{12}{24.0}{\rmdefault}{\mddefault}{\updefault}{\color[rgb]{0,0,0}$n+1$}%
}}}}
\put(5581,-3481){\makebox(0,0)[b]{\smash{{\SetFigFont{12}{24.0}{\rmdefault}{\mddefault}{\updefault}{\color[rgb]{0,0,0}$n+1$}%
}}}}
\put(5356,-3751){\makebox(0,0)[lb]{\smash{{\SetFigFont{12}{24.0}{\rmdefault}{\mddefault}{\updefault}{\color[rgb]{0,0,0}$n+1$}%
}}}}
\end{picture}%
 %
\nop
{
}

In words, if the only label is $n$, the others are $n$. If it is $1,n$, the
others are $n+1$.

This labeling is iterated until all nodes are labeled. Labels then tell the
class each formula goes into: $1,n$ means class one, $n$ means class $n$. If
the maxsets form a forest, which for example happens if there are isolated
maxsets, the procedure is iterated on all its trees.

The procedure of labelling ensures that the following conditions hold:

\begin{enumerate}

\item every maxset contains at most a label $1,n$;

\item if it does, the others are all $n$ if selected or $n+1$ if excluded;

\item otherwise, the maxset is excluded and its labels are equal to a value
greater than one;

\item every label $1,n$ is in at least a selected maxset, and every selected
maxset contains at least a label $1,n$.

\end{enumerate}

In other words, every selected maxset contains a label $1,n$ and the remaining
labels are $n$; every excluded maxset has either equal labels greater than one
or a label $1,n$ and all others $n+1$; every $1,n$ label is in at least a
selected maxset.

This way, selected maxsets are minimal because they contain a node in class
one, the rest in class $n$, and all other maxsets containing the same node in
class one have the others in class $n$. Excluded maxsets are not minimal
because they either contain no formula in class one, or otherwise they contain
a formula labeled $1,n$, the others are in class $n+1$, and the node labeled
$1,n$ is is in another maxset having formulae in class $n$.

In order to complete the proof, we show that the four conditions are ensured
when the procedure start, and that none of its step makes them false.

If the first maxset is selected, its first label is $1,2$ and the others are
$2$. If it is excluded, all its labels are $2$. The conditions therefore hold
up to this point.

At each iteration:

\begin{itemize}

\item if the maxset is selected, either has the initial node $1,n$ and is added
$n$ to the others, or it has $n$ in the first node and is added $1,n$ to one of
the others and $n$ to the remaining one; this ensures that it contains at least
a label $1,n$ and the others are all $n$;

\item if the maxset is excluded, it ends up with all labels $n>1$, or with a
single label $1,n$ and the others are $n+1$.

\end{itemize}

Either way, a set may contain a label $1,n$ only if it is the initial label,
and then no other $1,m$ is ever added, or it is added in a single node of a
selected set that has $n$ has the initial label.

Finally, a label $1,n$ is added only in a single case: on a selected set, if
the initial node is labeled $n$. As a result, every $1,n$ is in a selected set
that contains $n$ has the other labels.~\qed

While Berge-acyclic hypergraphs are obtainable, the converse is not always the
case: some Berge-cyclic hypergraphs are obtainable. Contrasting
Corollary~\ref{characterization}, which proves that alternating cycles imply
unobtainability for binary maxsets, in the general case alternating cycles may
be obtainable:

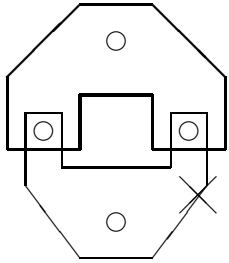
\setlength{\unitlength}{5000sp}%
\begingroup\makeatletter\ifx\SetFigFont\undefined%
\gdef\SetFigFont#1#2#3#4#5{%
  \reset@font\fontsize{#1}{#2pt}%
  \fontfamily{#3}\fontseries{#4}\fontshape{#5}%
  \selectfont}%
\fi\endgroup%
\begin{picture}(1124,1294)(4749,-4303)
{\color[rgb]{0,0,0}\thinlines
\put(4951,-3661){\circle{90}}
}%
{\color[rgb]{0,0,0}\put(5671,-3661){\circle{90}}
}%
{\color[rgb]{0,0,0}\put(5311,-4111){\circle{90}}
}%
{\color[rgb]{0,0,0}\put(5311,-3211){\circle{90}}
}%
{\color[rgb]{0,0,0}\put(4861,-3571){\line( 1, 0){180}}
\put(5041,-3571){\line( 0,-1){270}}
\put(5041,-3841){\line( 1, 0){540}}
\put(5581,-3841){\line( 0, 1){270}}
\put(5581,-3571){\line( 1, 0){180}}
\put(5761,-3571){\line( 0,-1){360}}
\put(5761,-3931){\line(-3,-4){270}}
\put(5491,-4291){\line(-1, 0){360}}
\put(5131,-4291){\line(-3, 4){270}}
\put(4861,-3931){\line( 0, 1){360}}
}%
\thicklines
{\color[rgb]{0,0,0}\put(5131,-3751){\line(-1, 0){360}}
\put(4771,-3751){\line( 0, 1){360}}
\put(4771,-3391){\line( 1, 1){360}}
\put(5131,-3031){\line( 1, 0){360}}
\put(5491,-3031){\line( 1,-1){360}}
\put(5851,-3391){\line( 0,-1){360}}
\put(5851,-3751){\line(-1, 0){360}}
\put(5491,-3751){\line( 0, 1){270}}
\put(5491,-3481){\line(-1, 0){360}}
\put(5131,-3481){\line( 0,-1){270}}
}%
\thinlines
{\color[rgb]{0,0,0}\put(5626,-4066){\line( 1, 1){180}}
}%
{\color[rgb]{0,0,0}\put(5626,-3886){\line( 1,-1){180}}
}%
\end{picture}%
 %
\nop
{
/-----------\                          .
|     O     |
|+---------+|
|| O     O ||
+-----------/
 |    O    |
 \--X------/
}

The maxset on the top is selected, the other excluded. This hypergraph is
Berge-cyclic, yet is obtained with a two-classes priority ordering:

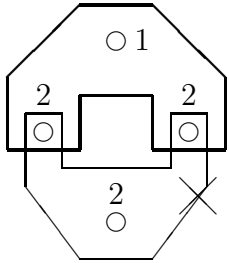
\setlength{\unitlength}{5000sp}%
\begingroup\makeatletter\ifx\SetFigFont\undefined%
\gdef\SetFigFont#1#2#3#4#5{%
  \reset@font\fontsize{#1}{#2pt}%
  \fontfamily{#3}\fontseries{#4}\fontshape{#5}%
  \selectfont}%
\fi\endgroup%
\begin{picture}(1124,1294)(4749,-4303)
{\color[rgb]{0,0,0}\thinlines
\put(4951,-3661){\circle{90}}
}%
{\color[rgb]{0,0,0}\put(5671,-3661){\circle{90}}
}%
{\color[rgb]{0,0,0}\put(5311,-4111){\circle{90}}
}%
{\color[rgb]{0,0,0}\put(5311,-3211){\circle{90}}
}%
{\color[rgb]{0,0,0}\put(4861,-3571){\line( 1, 0){180}}
\put(5041,-3571){\line( 0,-1){270}}
\put(5041,-3841){\line( 1, 0){540}}
\put(5581,-3841){\line( 0, 1){270}}
\put(5581,-3571){\line( 1, 0){180}}
\put(5761,-3571){\line( 0,-1){360}}
\put(5761,-3931){\line(-3,-4){270}}
\put(5491,-4291){\line(-1, 0){360}}
\put(5131,-4291){\line(-3, 4){270}}
\put(4861,-3931){\line( 0, 1){360}}
}%
\thicklines
{\color[rgb]{0,0,0}\put(5131,-3751){\line(-1, 0){360}}
\put(4771,-3751){\line( 0, 1){360}}
\put(4771,-3391){\line( 1, 1){360}}
\put(5131,-3031){\line( 1, 0){360}}
\put(5491,-3031){\line( 1,-1){360}}
\put(5851,-3391){\line( 0,-1){360}}
\put(5851,-3751){\line(-1, 0){360}}
\put(5491,-3751){\line( 0, 1){270}}
\put(5491,-3481){\line(-1, 0){360}}
\put(5131,-3481){\line( 0,-1){270}}
}%
\thinlines
{\color[rgb]{0,0,0}\put(5626,-3886){\line( 1,-1){180}}
}%
{\color[rgb]{0,0,0}\put(5626,-4066){\line( 1, 1){180}}
}%
\put(5401,-3256){\makebox(0,0)[lb]{\smash{{\SetFigFont{12}{24.0}{\rmdefault}{\mddefault}{\updefault}{\color[rgb]{0,0,0}$1$}%
}}}}
\put(4951,-3526){\makebox(0,0)[b]{\smash{{\SetFigFont{12}{24.0}{\rmdefault}{\mddefault}{\updefault}{\color[rgb]{0,0,0}$2$}%
}}}}
\put(5671,-3526){\makebox(0,0)[b]{\smash{{\SetFigFont{12}{24.0}{\rmdefault}{\mddefault}{\updefault}{\color[rgb]{0,0,0}$2$}%
}}}}
\put(5311,-4021){\makebox(0,0)[b]{\smash{{\SetFigFont{12}{24.0}{\rmdefault}{\mddefault}{\updefault}{\color[rgb]{0,0,0}$2$}%
}}}}
\end{picture}%
 %
\nop
{
/-----------\                         .
|     O 1   |
|+---------+|
|| O 2 2 O ||
+-----------/
 |    O 2  |
 \--X------/
}

 %

\subsection{Binary maxsets}

A particular case of the problem of obtainability by priority base merging is
when maxsets comprise at most two formulae. This may be guaranteed to hold in a
specific domains, but the main reasons for studying this case are: first, it
provides proofs of existence of some specific cases, such as one requiring $n$
classes of priority for obtainability; second, it is a subcase where a
necessary and sufficient condition for obtainability can be given, that of
alternating cycles of maxsets; third, it provides guiding principles for a
future study of the general case, where no such necessary and sufficient
condition is known.

When all maxsets comprise at most two formulae, they can be seen as a graph:

\begin{itemize}

\item nodes are formulae;

\item isolated nodes are singleton maxsets;

\item edges are maxsets of two formulae.

\end{itemize}

This section is organized as follows:

\begin{enumerate}

\item definitions and basic properties;

\item transformations on graphs;

\item properties of some specific graphs or subgraphs;

\item proof that a graph is unobtainable if and only if it contains a cycle of
alternating single excluded--odd sequence of selected edges.

\end{enumerate}

Cycles are defined as closed paths: a sequence of edges ending where it
started. They differ from simple cycles, which are not allowed to 
cross an edge more than once.

\subsubsection{Definitions}

When all maxsets contain at most two formulae, the singletons can be excluded
from consideration because of Lemma~\ref{inconsistent}: $\{A\}$ cannot be
contained in any other maxset; therefore, inclusion or exclusion do not affect
the other maxsets. What remains is a set binary maxsets, which can be seen as a
graph where nodes are formulae and edges are maxsets. Some edges correspond to
selected maxsets, the remaining ones to excluded maxsets.

\begin{definition}
\label{se-graph}

A {\em selected-excluded graph} (abbreviated: {\em se graph}) is a graph whose
edges are partitioned in two sets: selected and excluded.

\end{definition}

Since edges are maxsets, the distinction indicates which are required to be in
the result of merging and which are not. Most of the proofs regarding binary
maxsets employ assignments of some formulae to priority class.

\begin{definition}

A {\em partially assigned se graph} has some nodes assigned positive integer
values. If all nodes are assigned the graph is totally assigned.

\end{definition}

In a totally assigned se graph, all formulae are assigned a class. Therefore,
one may determine the minimal edges (maxsets) and check whether they are
exactly the selected ones.

\begin{definition}

A totally assigned graph is obtainable if the minimal edges according to the
priority ordering obtained from the numbers assigned to the nodes are exactly
the selected ones.

\end{definition}

This definition may look tautological, but is rather close to the opposite. In
a se graph, the selected edges are the maxsets that are required to be in the
result of merging: if $\{A,B\} \models R$, the edge $(A,B)$ is selected and
vice versa. The values assigned to nodes may or may not make such a maxset
minimal. If it is not, the edge is {\bf incorrectly excluded}. Similarly, an
excluded edge that is minimal according to the values is incorrectly selected.
If no edge is incorrectly selected or excluded the ordering produces the
required result.

\begin{definition}

A partially or totally assigned se graph $G$ extends another one $H$ if they
have the same nodes and edges and all nodes assigned in $H$ are also assigned
in $G$ to the same values.

\end{definition}

A se graph is therefore obtainable if and only if it can be extended to a
totally assigned se graph that is obtainable. On totally assigned se graphs
obtainability can be checked by determining the minimal maxsets according to
the ordering given by the values.

\subsubsection{Influence}

On totally assigned se graphs, one can check selection or exclusion of every
edge by determining its minimality according the values. The following lemma
shows which values affect the minimality of a particular edge.

\begin{lemma}
\label{influence}

In a totally assigned se graph, minimality of an edge $(a,b)$ depends only on:

\begin{enumerate}

\item the values of $a$ and $b$, and

\item if the value of $a$ is one and the value of $b$ is not, on the values of
the nodes linked to $a$;

\item if the value of $b$ is one and the value of $a$ is not, on the values of
the nodes linked to $b$.

\end{enumerate}

\end{lemma}

\proof If the values of $a$ and $b$ are both one, the edge is minimal no matter
of what the other values are. If $a$ and $b$ are both greater than one, the
edge is not minimal.

Of the remaining case, suffices to consider $a$ assigned to one and $b$ to a
larger value: the other is specular. If all nodes linked to $a$ are greater or
equal than $b$, then $(a,b)$ is minimal. If one of them is lesser, it is not.
In both cases, no other value of the graph affects the result.~\qed

This lemma could be also refined: of a node of value one, the only information
that counts is the minimal values of nodes linked to it.

\subsubsection{Value-depending transformations}

Se graphs can be simplified without affecting obtainability: the resulting graph
is obtainable if and only if the original one is. Correctness is proved by a
detour to the totally assigned graphs extending the original and resulting
ones. In particular:

\begin{itemize}

\item a partially assigned se graph is obtainable if and only if it can be
extended to a totally assigned one that is also obtainable;

\item obtainability on totally assigned se graphs is verified by checking that
the minimal edges are exactly the selected ones;

\item the transformations do not turn a minimal edge into a non-minimal one in
the totally assigned se graphs, and vice versa;

\item in most cases, the transformations remove or add only edges that are
correctly selected or excluded in the totally assigned se graph; otherwise,
they replace correctly/incorrectly selected or excluded edges with edges that
are equally correct or incorrect.

\end{itemize}

All this proves that the transformations are correct: they map a partially
assigned se graph into another whose extensions to totally assigned se graphs
correspond to the ones of the original graph, and this correspondence maps
obtainable graphs into obtainable graphs and vice versa. As a result, the
original graph is obtainable if and only if the resulting graph is. In most
cases, obtainability is maintained simply because edge minimality is unaffected
by the transformation.

The first simplification is disconnection, which is done in three different
ways depending on the values.

\

\noindent{\bf Disconnection, both greater than one.}

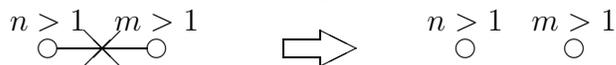
\setlength{\unitlength}{5000sp}%
\begingroup\makeatletter\ifx\SetFigFont\undefined%
\gdef\SetFigFont#1#2#3#4#5{%
  \reset@font\fontsize{#1}{#2pt}%
  \fontfamily{#3}\fontseries{#4}\fontshape{#5}%
  \selectfont}%
\fi\endgroup%
\begin{picture}(2716,342)(4808,-3493)
{\color[rgb]{0,0,0}\thinlines
\put(4861,-3391){\circle{90}}
}%
{\color[rgb]{0,0,0}\put(5401,-3391){\circle{90}}
}%
{\color[rgb]{0,0,0}\put(6931,-3391){\circle{90}}
}%
{\color[rgb]{0,0,0}\put(7471,-3391){\circle{90}}
}%
{\color[rgb]{0,0,0}\put(4906,-3391){\line( 1, 0){450}}
}%
{\color[rgb]{0,0,0}\put(5041,-3301){\line( 1,-1){180}}
}%
{\color[rgb]{0,0,0}\put(5041,-3481){\line( 1, 1){180}}
}%
{\color[rgb]{0,0,0}\put(6391,-3391){\line(-2, 1){180}}
\put(6211,-3301){\line( 0,-1){ 45}}
\put(6211,-3346){\line(-1, 0){180}}
\put(6031,-3346){\line( 0,-1){ 90}}
\put(6031,-3436){\line( 1, 0){180}}
\put(6211,-3436){\line( 0,-1){ 45}}
\put(6211,-3481){\line( 2, 1){180}}
}%
\put(4861,-3301){\makebox(0,0)[b]{\smash{{\SetFigFont{12}{24.0}{\rmdefault}{\mddefault}{\updefault}{\color[rgb]{0,0,0}$n>1$}%
}}}}
\put(5401,-3301){\makebox(0,0)[b]{\smash{{\SetFigFont{12}{24.0}{\rmdefault}{\mddefault}{\updefault}{\color[rgb]{0,0,0}$m>1$}%
}}}}
\put(6931,-3301){\makebox(0,0)[b]{\smash{{\SetFigFont{12}{24.0}{\rmdefault}{\mddefault}{\updefault}{\color[rgb]{0,0,0}$n>1$}%
}}}}
\put(7471,-3301){\makebox(0,0)[b]{\smash{{\SetFigFont{12}{24.0}{\rmdefault}{\mddefault}{\updefault}{\color[rgb]{0,0,0}$m>1$}%
}}}}
\end{picture}%
 %
\nop
{
n>1   m>1            n>1   m>1
 O--X--O      ===>    O     O
}

An edge between two nodes of value greater than one can be removed.

In every extension to a totally assigned se graph, the edge is correctly
excluded. Therefore, obtainability in both the graph before and after the
change depends only on the minimality of the other edges.

If an edge does not touch the disconnected one, by Lemma~\ref{influence} its
minimality is unaffected by the change. But the lemma implies the same for
edges touching the deleted one:

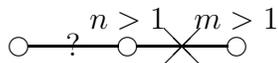
\setlength{\unitlength}{5000sp}%
\begingroup\makeatletter\ifx\SetFigFont\undefined%
\gdef\SetFigFont#1#2#3#4#5{%
  \reset@font\fontsize{#1}{#2pt}%
  \fontfamily{#3}\fontseries{#4}\fontshape{#5}%
  \selectfont}%
\fi\endgroup%
\begin{picture}(1186,342)(4268,-3493)
{\color[rgb]{0,0,0}\thinlines
\put(4861,-3391){\circle{90}}
}%
{\color[rgb]{0,0,0}\put(5401,-3391){\circle{90}}
}%
{\color[rgb]{0,0,0}\put(4321,-3391){\circle{90}}
}%
{\color[rgb]{0,0,0}\put(4906,-3391){\line( 1, 0){450}}
}%
{\color[rgb]{0,0,0}\put(5041,-3301){\line( 1,-1){180}}
}%
{\color[rgb]{0,0,0}\put(5041,-3481){\line( 1, 1){180}}
}%
{\color[rgb]{0,0,0}\put(4366,-3391){\line( 1, 0){450}}
}%
\put(4861,-3301){\makebox(0,0)[b]{\smash{{\SetFigFont{12}{24.0}{\rmdefault}{\mddefault}{\updefault}{\color[rgb]{0,0,0}$n>1$}%
}}}}
\put(4591,-3436){\makebox(0,0)[b]{\smash{{\SetFigFont{12}{24.0}{\rmdefault}{\mddefault}{\updefault}{\color[rgb]{0,0,0}$?$}%
}}}}
\put(5401,-3301){\makebox(0,0)[b]{\smash{{\SetFigFont{12}{24.0}{\rmdefault}{\mddefault}{\updefault}{\color[rgb]{0,0,0}$m>1$}%
}}}}
\end{picture}%
 %
\nop
{
     n>1   m>1
O--?--O--X--O
}

In this and the following figures, a question mark indicates that the edge may
be selected or excluded, and the following reasoning holds in both cases.

Since $n>1$, minimality of the other edge depends on $n$ only, and not on nodes
linked to the one of value $n$. The presence of the removed edge is therefore
irrelevant.

\

\noindent{\bf Disconnection, one assigned one.}

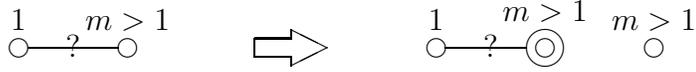
\setlength{\unitlength}{5000sp}%
\begingroup\makeatletter\ifx\SetFigFont\undefined%
\gdef\SetFigFont#1#2#3#4#5{%
  \reset@font\fontsize{#1}{#2pt}%
  \fontfamily{#3}\fontseries{#4}\fontshape{#5}%
  \selectfont}%
\fi\endgroup%
\begin{picture}(3256,387)(4808,-3493)
{\color[rgb]{0,0,0}\thinlines
\put(4861,-3391){\circle{90}}
}%
{\color[rgb]{0,0,0}\put(5401,-3391){\circle{90}}
}%
{\color[rgb]{0,0,0}\put(6931,-3391){\circle{90}}
}%
{\color[rgb]{0,0,0}\put(8011,-3391){\circle{90}}
}%
{\color[rgb]{0,0,0}\put(7471,-3391){\circle{90}}
}%
{\color[rgb]{0,0,0}\put(7471,-3391){\circle{180}}
}%
{\color[rgb]{0,0,0}\put(4906,-3391){\line( 1, 0){450}}
}%
{\color[rgb]{0,0,0}\put(6391,-3391){\line(-2, 1){180}}
\put(6211,-3301){\line( 0,-1){ 45}}
\put(6211,-3346){\line(-1, 0){180}}
\put(6031,-3346){\line( 0,-1){ 90}}
\put(6031,-3436){\line( 1, 0){180}}
\put(6211,-3436){\line( 0,-1){ 45}}
\put(6211,-3481){\line( 2, 1){180}}
}%
{\color[rgb]{0,0,0}\put(6976,-3391){\line( 1, 0){405}}
}%
\put(5401,-3301){\makebox(0,0)[b]{\smash{{\SetFigFont{12}{24.0}{\rmdefault}{\mddefault}{\updefault}{\color[rgb]{0,0,0}$m>1$}%
}}}}
\put(4861,-3301){\makebox(0,0)[b]{\smash{{\SetFigFont{12}{24.0}{\rmdefault}{\mddefault}{\updefault}{\color[rgb]{0,0,0}$1$}%
}}}}
\put(6931,-3301){\makebox(0,0)[b]{\smash{{\SetFigFont{12}{24.0}{\rmdefault}{\mddefault}{\updefault}{\color[rgb]{0,0,0}$1$}%
}}}}
\put(5131,-3436){\makebox(0,0)[b]{\smash{{\SetFigFont{12}{24.0}{\rmdefault}{\mddefault}{\updefault}{\color[rgb]{0,0,0}$?$}%
}}}}
\put(7201,-3436){\makebox(0,0)[b]{\smash{{\SetFigFont{12}{24.0}{\rmdefault}{\mddefault}{\updefault}{\color[rgb]{0,0,0}$?$}%
}}}}
\put(8011,-3301){\makebox(0,0)[b]{\smash{{\SetFigFont{12}{24.0}{\rmdefault}{\mddefault}{\updefault}{\color[rgb]{0,0,0}$m>1$}%
}}}}
\put(7471,-3256){\makebox(0,0)[b]{\smash{{\SetFigFont{12}{24.0}{\rmdefault}{\mddefault}{\updefault}{\color[rgb]{0,0,0}$m>1$}%
}}}}
\end{picture}%
 %
\nop
{
1    m>1            1     m>1    m>1
O--?--O     ===>    O--?--(O)     O
}

The double circle is a new node connected to none else. In this transformation,
an edge between a node of value one and a node of value greater than one
becomes an edge between the first and an isolated copy of the second.

In the totally assigned se graph extending the original one the edge may be
minimal or not, but either way its status is not changed by the transformation,
as its nodes maintain their value and its node of value one is connected to the
same nodes as before. As a result, selection is either correct in both graphs
or incorrect in both.

Regarding the other edges, minimality is not changed by the disconnection. If
one such edge does not touch the disconnected one, or touches the node greater
than one, Lemma~\ref{influence} tells that its minimality is not affected. But
the same also holds for edges touching the node of value one, since this is
connected to the same nodes as before, except that instead of the old node of
value $m$ is connected to a new node of value $m$.

\

\noindent {\bf Disconnection, both assigned one.}

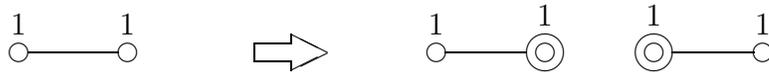
\setlength{\unitlength}{5000sp}%
\begingroup\makeatletter\ifx\SetFigFont\undefined%
\gdef\SetFigFont#1#2#3#4#5{%
  \reset@font\fontsize{#1}{#2pt}%
  \fontfamily{#3}\fontseries{#4}\fontshape{#5}%
  \selectfont}%
\fi\endgroup%
\begin{picture}(3796,387)(4808,-3493)
{\color[rgb]{0,0,0}\thinlines
\put(4861,-3391){\circle{90}}
}%
{\color[rgb]{0,0,0}\put(5401,-3391){\circle{90}}
}%
{\color[rgb]{0,0,0}\put(6931,-3391){\circle{90}}
}%
{\color[rgb]{0,0,0}\put(8011,-3391){\circle{90}}
}%
{\color[rgb]{0,0,0}\put(7471,-3391){\circle{90}}
}%
{\color[rgb]{0,0,0}\put(7471,-3391){\circle{180}}
}%
{\color[rgb]{0,0,0}\put(8011,-3391){\circle{180}}
}%
{\color[rgb]{0,0,0}\put(8551,-3391){\circle{90}}
}%
{\color[rgb]{0,0,0}\put(4906,-3391){\line( 1, 0){450}}
}%
{\color[rgb]{0,0,0}\put(6391,-3391){\line(-2, 1){180}}
\put(6211,-3301){\line( 0,-1){ 45}}
\put(6211,-3346){\line(-1, 0){180}}
\put(6031,-3346){\line( 0,-1){ 90}}
\put(6031,-3436){\line( 1, 0){180}}
\put(6211,-3436){\line( 0,-1){ 45}}
\put(6211,-3481){\line( 2, 1){180}}
}%
{\color[rgb]{0,0,0}\put(6976,-3391){\line( 1, 0){405}}
}%
{\color[rgb]{0,0,0}\put(8101,-3391){\line( 1, 0){405}}
}%
\put(4861,-3301){\makebox(0,0)[b]{\smash{{\SetFigFont{12}{24.0}{\rmdefault}{\mddefault}{\updefault}{\color[rgb]{0,0,0}$1$}%
}}}}
\put(6931,-3301){\makebox(0,0)[b]{\smash{{\SetFigFont{12}{24.0}{\rmdefault}{\mddefault}{\updefault}{\color[rgb]{0,0,0}$1$}%
}}}}
\put(5401,-3301){\makebox(0,0)[b]{\smash{{\SetFigFont{12}{24.0}{\rmdefault}{\mddefault}{\updefault}{\color[rgb]{0,0,0}$1$}%
}}}}
\put(7471,-3256){\makebox(0,0)[b]{\smash{{\SetFigFont{12}{24.0}{\rmdefault}{\mddefault}{\updefault}{\color[rgb]{0,0,0}$1$}%
}}}}
\put(8011,-3256){\makebox(0,0)[b]{\smash{{\SetFigFont{12}{24.0}{\rmdefault}{\mddefault}{\updefault}{\color[rgb]{0,0,0}$1$}%
}}}}
\put(8551,-3301){\makebox(0,0)[b]{\smash{{\SetFigFont{12}{24.0}{\rmdefault}{\mddefault}{\updefault}{\color[rgb]{0,0,0}$1$}%
}}}}
\end{picture}%
 %
\nop
{
1     1              1      1        1      1
O-----O     ===>     O-----(O)      (O)-----O
}

The double circles are new nodes, connected to none else. An edge between two
nodes of value one is split in two, each linking one of the original nodes to
an isolated copy of the second.

The original edge is correctly selected in the original graph, and the two new
ones are correctly selected in the resulting one. Therefore, obtainability
depends only on the minimality of the other edges, which will be proved to be
unchanged by the transformation.

By symmetry and Lemma~\ref{influence}, the only relevant case is about edges
touching the first node of the original edge. After the change, the node is
still connected to the same other nodes and to a node of value one, as before.
Therefore, minimality of the other edge is unaffected.

\

\noindent{\bf Merging of selected edges.}

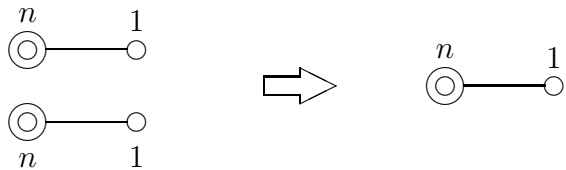
\setlength{\unitlength}{5000sp}%
\begingroup\makeatletter\ifx\SetFigFont\undefined%
\gdef\SetFigFont#1#2#3#4#5{%
  \reset@font\fontsize{#1}{#2pt}%
  \fontfamily{#3}\fontseries{#4}\fontshape{#5}%
  \selectfont}%
\fi\endgroup%
\begin{picture}(2761,840)(5033,-3631)
{\color[rgb]{0,0,0}\thinlines
\put(5131,-3031){\circle{90}}
}%
{\color[rgb]{0,0,0}\put(5671,-3031){\circle{90}}
}%
{\color[rgb]{0,0,0}\put(5131,-3391){\circle{90}}
}%
{\color[rgb]{0,0,0}\put(5671,-3391){\circle{90}}
}%
{\color[rgb]{0,0,0}\put(5131,-3031){\circle{180}}
}%
{\color[rgb]{0,0,0}\put(5131,-3391){\circle{180}}
}%
{\color[rgb]{0,0,0}\put(7201,-3211){\circle{90}}
}%
{\color[rgb]{0,0,0}\put(7201,-3211){\circle{180}}
}%
{\color[rgb]{0,0,0}\put(7741,-3211){\circle{90}}
}%
{\color[rgb]{0,0,0}\put(5221,-3031){\line( 1, 0){405}}
}%
{\color[rgb]{0,0,0}\put(5221,-3391){\line( 1, 0){405}}
}%
{\color[rgb]{0,0,0}\put(6661,-3211){\line(-2, 1){180}}
\put(6481,-3121){\line( 0,-1){ 45}}
\put(6481,-3166){\line(-1, 0){180}}
\put(6301,-3166){\line( 0,-1){ 90}}
\put(6301,-3256){\line( 1, 0){180}}
\put(6481,-3256){\line( 0,-1){ 45}}
\put(6481,-3301){\line( 2, 1){180}}
}%
{\color[rgb]{0,0,0}\put(7291,-3211){\line( 1, 0){405}}
}%
\put(5131,-2896){\makebox(0,0)[b]{\smash{{\SetFigFont{12}{24.0}{\rmdefault}{\mddefault}{\updefault}{\color[rgb]{0,0,0}$n$}%
}}}}
\put(5671,-2941){\makebox(0,0)[b]{\smash{{\SetFigFont{12}{24.0}{\rmdefault}{\mddefault}{\updefault}{\color[rgb]{0,0,0}$1$}%
}}}}
\put(5131,-3616){\makebox(0,0)[b]{\smash{{\SetFigFont{12}{24.0}{\rmdefault}{\mddefault}{\updefault}{\color[rgb]{0,0,0}$n$}%
}}}}
\put(5671,-3616){\makebox(0,0)[b]{\smash{{\SetFigFont{12}{24.0}{\rmdefault}{\mddefault}{\updefault}{\color[rgb]{0,0,0}$1$}%
}}}}
\put(7201,-3076){\makebox(0,0)[b]{\smash{{\SetFigFont{12}{24.0}{\rmdefault}{\mddefault}{\updefault}{\color[rgb]{0,0,0}$n$}%
}}}}
\put(7741,-3121){\makebox(0,0)[b]{\smash{{\SetFigFont{12}{24.0}{\rmdefault}{\mddefault}{\updefault}{\color[rgb]{0,0,0}$1$}%
}}}}
\end{picture}%
 %
\nop
{
 n     1
(O)----O                 n      1 
              ===>      (O)-----O
(O)----O
 n     1
}

The double circles indicate nodes connected to none else. The two original
nodes of value one may be touched by other edges, which are connected to the
merged node of value one after the transformation.

If any of the two nodes of value one is linked to one of value less than $n$,
the same happens in the resulting edge, and vice versa. As a result, if any of
the original edges is incorrectly selected so is the resulting edge, and vice
versa. Therefore, remains to show that obtainability is unaffected by the
change only if the two original nodes assigned one are not linked to a node of
value less than $n$.

Selection of edges not touching the nodes assigned one is not changed because
of Lemma~\ref{influence}. Regarding the edges touching one of these, let $k$ be
the value of the other node:

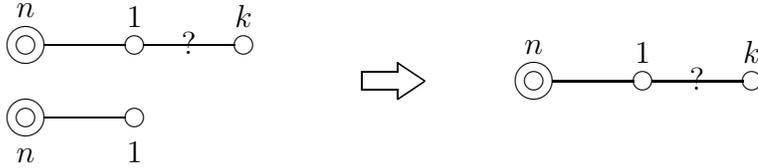
\setlength{\unitlength}{5000sp}%
\begingroup\makeatletter\ifx\SetFigFont\undefined%
\gdef\SetFigFont#1#2#3#4#5{%
  \reset@font\fontsize{#1}{#2pt}%
  \fontfamily{#3}\fontseries{#4}\fontshape{#5}%
  \selectfont}%
\fi\endgroup%
\begin{picture}(3751,840)(5033,-3631)
{\color[rgb]{0,0,0}\thinlines
\put(5131,-3031){\circle{90}}
}%
{\color[rgb]{0,0,0}\put(5671,-3031){\circle{90}}
}%
{\color[rgb]{0,0,0}\put(5131,-3391){\circle{90}}
}%
{\color[rgb]{0,0,0}\put(5671,-3391){\circle{90}}
}%
{\color[rgb]{0,0,0}\put(5131,-3031){\circle{180}}
}%
{\color[rgb]{0,0,0}\put(5131,-3391){\circle{180}}
}%
{\color[rgb]{0,0,0}\put(6211,-3031){\circle{90}}
}%
{\color[rgb]{0,0,0}\put(7651,-3211){\circle{90}}
}%
{\color[rgb]{0,0,0}\put(7651,-3211){\circle{180}}
}%
{\color[rgb]{0,0,0}\put(8191,-3211){\circle{90}}
}%
{\color[rgb]{0,0,0}\put(8731,-3211){\circle{90}}
}%
{\color[rgb]{0,0,0}\put(5221,-3031){\line( 1, 0){405}}
}%
{\color[rgb]{0,0,0}\put(5221,-3391){\line( 1, 0){405}}
}%
{\color[rgb]{0,0,0}\put(7111,-3211){\line(-3, 2){135}}
\put(6976,-3121){\line( 0,-1){ 45}}
\put(6976,-3166){\line(-1, 0){180}}
\put(6796,-3166){\line( 0,-1){ 90}}
\put(6796,-3256){\line( 1, 0){180}}
\put(6976,-3256){\line( 0,-1){ 45}}
\put(6976,-3301){\line( 3, 2){135}}
}%
{\color[rgb]{0,0,0}\put(5716,-3031){\line( 1, 0){450}}
}%
{\color[rgb]{0,0,0}\put(7741,-3211){\line( 1, 0){405}}
}%
{\color[rgb]{0,0,0}\put(8236,-3211){\line( 1, 0){450}}
}%
\put(5131,-2896){\makebox(0,0)[b]{\smash{{\SetFigFont{12}{24.0}{\rmdefault}{\mddefault}{\updefault}{\color[rgb]{0,0,0}$n$}%
}}}}
\put(5671,-2941){\makebox(0,0)[b]{\smash{{\SetFigFont{12}{24.0}{\rmdefault}{\mddefault}{\updefault}{\color[rgb]{0,0,0}$1$}%
}}}}
\put(5131,-3616){\makebox(0,0)[b]{\smash{{\SetFigFont{12}{24.0}{\rmdefault}{\mddefault}{\updefault}{\color[rgb]{0,0,0}$n$}%
}}}}
\put(5671,-3616){\makebox(0,0)[b]{\smash{{\SetFigFont{12}{24.0}{\rmdefault}{\mddefault}{\updefault}{\color[rgb]{0,0,0}$1$}%
}}}}
\put(5941,-3076){\makebox(0,0)[b]{\smash{{\SetFigFont{12}{24.0}{\rmdefault}{\mddefault}{\updefault}{\color[rgb]{0,0,0}$?$}%
}}}}
\put(6211,-2941){\makebox(0,0)[b]{\smash{{\SetFigFont{12}{24.0}{\rmdefault}{\mddefault}{\updefault}{\color[rgb]{0,0,0}$k$}%
}}}}
\put(7651,-3076){\makebox(0,0)[b]{\smash{{\SetFigFont{12}{24.0}{\rmdefault}{\mddefault}{\updefault}{\color[rgb]{0,0,0}$n$}%
}}}}
\put(8191,-3121){\makebox(0,0)[b]{\smash{{\SetFigFont{12}{24.0}{\rmdefault}{\mddefault}{\updefault}{\color[rgb]{0,0,0}$1$}%
}}}}
\put(8461,-3256){\makebox(0,0)[b]{\smash{{\SetFigFont{12}{24.0}{\rmdefault}{\mddefault}{\updefault}{\color[rgb]{0,0,0}$?$}%
}}}}
\put(8731,-3121){\makebox(0,0)[b]{\smash{{\SetFigFont{12}{24.0}{\rmdefault}{\mddefault}{\updefault}{\color[rgb]{0,0,0}$k$}%
}}}}
\end{picture}%
 %
\nop
{
 n     1     k
(O)----O--?--O                  n     1     k
                     ===>      (O)----O--?--O
(O)----O
 n     1
}

In the original totally assigned se graph, all other nodes linked to the ones
assigned one have values greater than $n$. As a result, the minimality of this
edge depends only on whether $k$ is equal to $n$ or greater. The same happens
in the resulting graph.

\

\noindent{\bf Merging of excluded edges.}

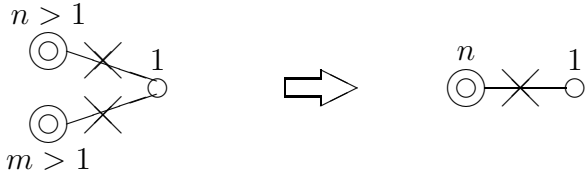
\setlength{\unitlength}{5000sp}%
\begingroup\makeatletter\ifx\SetFigFont\undefined%
\gdef\SetFigFont#1#2#3#4#5{%
  \reset@font\fontsize{#1}{#2pt}%
  \fontfamily{#3}\fontseries{#4}\fontshape{#5}%
  \selectfont}%
\fi\endgroup%
\begin{picture}(2761,885)(5033,-3631)
{\color[rgb]{0,0,0}\thinlines
\put(5131,-3031){\circle{90}}
}%
{\color[rgb]{0,0,0}\put(5131,-3391){\circle{90}}
}%
{\color[rgb]{0,0,0}\put(5131,-3031){\circle{180}}
}%
{\color[rgb]{0,0,0}\put(5131,-3391){\circle{180}}
}%
{\color[rgb]{0,0,0}\put(7201,-3211){\circle{90}}
}%
{\color[rgb]{0,0,0}\put(7201,-3211){\circle{180}}
}%
{\color[rgb]{0,0,0}\put(7741,-3211){\circle{90}}
}%
{\color[rgb]{0,0,0}\put(5671,-3211){\circle{90}}
}%
{\color[rgb]{0,0,0}\put(6661,-3211){\line(-2, 1){180}}
\put(6481,-3121){\line( 0,-1){ 45}}
\put(6481,-3166){\line(-1, 0){180}}
\put(6301,-3166){\line( 0,-1){ 90}}
\put(6301,-3256){\line( 1, 0){180}}
\put(6481,-3256){\line( 0,-1){ 45}}
\put(6481,-3301){\line( 2, 1){180}}
}%
{\color[rgb]{0,0,0}\put(7291,-3211){\line( 1, 0){405}}
}%
{\color[rgb]{0,0,0}\put(5221,-3031){\line( 3,-1){445.500}}
}%
{\color[rgb]{0,0,0}\put(5221,-3391){\line( 3, 1){445.500}}
}%
{\color[rgb]{0,0,0}\put(5311,-2986){\line( 1,-1){180}}
}%
{\color[rgb]{0,0,0}\put(5311,-3166){\line( 1, 1){180}}
}%
{\color[rgb]{0,0,0}\put(5311,-3256){\line( 1,-1){180}}
}%
{\color[rgb]{0,0,0}\put(5311,-3436){\line( 1, 1){180}}
}%
{\color[rgb]{0,0,0}\put(7381,-3121){\line( 1,-1){180}}
}%
{\color[rgb]{0,0,0}\put(7381,-3301){\line( 1, 1){180}}
}%
\put(7201,-3076){\makebox(0,0)[b]{\smash{{\SetFigFont{12}{24.0}{\rmdefault}{\mddefault}{\updefault}{\color[rgb]{0,0,0}$n$}%
}}}}
\put(7741,-3121){\makebox(0,0)[b]{\smash{{\SetFigFont{12}{24.0}{\rmdefault}{\mddefault}{\updefault}{\color[rgb]{0,0,0}$1$}%
}}}}
\put(5671,-3121){\makebox(0,0)[b]{\smash{{\SetFigFont{12}{24.0}{\rmdefault}{\mddefault}{\updefault}{\color[rgb]{0,0,0}$1$}%
}}}}
\put(5131,-2896){\makebox(0,0)[b]{\smash{{\SetFigFont{12}{24.0}{\rmdefault}{\mddefault}{\updefault}{\color[rgb]{0,0,0}$n>1$}%
}}}}
\put(5131,-3616){\makebox(0,0)[b]{\smash{{\SetFigFont{12}{24.0}{\rmdefault}{\mddefault}{\updefault}{\color[rgb]{0,0,0}$m>1$}%
}}}}
\end{picture}%
 %
\nop
{
n>1
(O)--X--\ 1           n      1
         O     ===>  (O)--X--O
(O)--X--/
m>1
}

In this figure, $n \leq m$. Double circles indicates nodes connected to none
else.

If the node of value one is only connected to nodes of value greater or equal
than $n$, then the original totally assigned se graph is unobtainable, and so
is the graph resulting from the transformation. Therefore, the only situation
where obtainability could be altered is then the node of value $1$ is connected
to at least a node of value less than $n$.

An edge that does not touch the node of value one is unaffected by the change
by Lemma~\ref{influence}. Let $k$ be the value of the other node of an edge
touching it, and $h$ the minimal values of nodes connected to the same node:

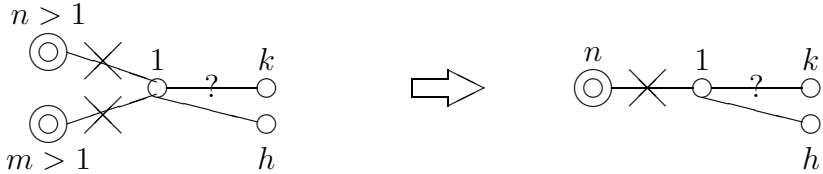
\setlength{\unitlength}{5000sp}%
\begingroup\makeatletter\ifx\SetFigFont\undefined%
\gdef\SetFigFont#1#2#3#4#5{%
  \reset@font\fontsize{#1}{#2pt}%
  \fontfamily{#3}\fontseries{#4}\fontshape{#5}%
  \selectfont}%
\fi\endgroup%
\begin{picture}(3931,885)(5033,-3631)
{\color[rgb]{0,0,0}\thinlines
\put(5131,-3031){\circle{90}}
}%
{\color[rgb]{0,0,0}\put(5131,-3391){\circle{90}}
}%
{\color[rgb]{0,0,0}\put(5131,-3031){\circle{180}}
}%
{\color[rgb]{0,0,0}\put(5131,-3391){\circle{180}}
}%
{\color[rgb]{0,0,0}\put(5671,-3211){\circle{90}}
}%
{\color[rgb]{0,0,0}\put(7831,-3211){\circle{90}}
}%
{\color[rgb]{0,0,0}\put(7831,-3211){\circle{180}}
}%
{\color[rgb]{0,0,0}\put(8371,-3211){\circle{90}}
}%
{\color[rgb]{0,0,0}\put(6211,-3211){\circle{90}}
}%
{\color[rgb]{0,0,0}\put(6211,-3391){\circle{90}}
}%
{\color[rgb]{0,0,0}\put(8911,-3211){\circle{90}}
}%
{\color[rgb]{0,0,0}\put(8911,-3391){\circle{90}}
}%
{\color[rgb]{0,0,0}\put(5221,-3031){\line( 3,-1){445.500}}
}%
{\color[rgb]{0,0,0}\put(5221,-3391){\line( 3, 1){445.500}}
}%
{\color[rgb]{0,0,0}\put(5311,-2986){\line( 1,-1){180}}
}%
{\color[rgb]{0,0,0}\put(5311,-3166){\line( 1, 1){180}}
}%
{\color[rgb]{0,0,0}\put(5311,-3256){\line( 1,-1){180}}
}%
{\color[rgb]{0,0,0}\put(5311,-3436){\line( 1, 1){180}}
}%
{\color[rgb]{0,0,0}\put(7921,-3211){\line( 1, 0){405}}
}%
{\color[rgb]{0,0,0}\put(8011,-3121){\line( 1,-1){180}}
}%
{\color[rgb]{0,0,0}\put(8011,-3301){\line( 1, 1){180}}
}%
{\color[rgb]{0,0,0}\put(5716,-3211){\line( 1, 0){450}}
}%
{\color[rgb]{0,0,0}\put(5671,-3256){\line( 4,-1){497.647}}
}%
{\color[rgb]{0,0,0}\put(7291,-3211){\line(-2, 1){180}}
\put(7111,-3121){\line( 0,-1){ 45}}
\put(7111,-3166){\line(-1, 0){180}}
\put(6931,-3166){\line( 0,-1){ 90}}
\put(6931,-3256){\line( 1, 0){180}}
\put(7111,-3256){\line( 0,-1){ 45}}
\put(7111,-3301){\line( 2, 1){180}}
}%
{\color[rgb]{0,0,0}\put(8416,-3211){\line( 1, 0){450}}
}%
{\color[rgb]{0,0,0}\put(8371,-3256){\line( 4,-1){497.647}}
}%
\put(5671,-3121){\makebox(0,0)[b]{\smash{{\SetFigFont{12}{24.0}{\rmdefault}{\mddefault}{\updefault}{\color[rgb]{0,0,0}$1$}%
}}}}
\put(7831,-3076){\makebox(0,0)[b]{\smash{{\SetFigFont{12}{24.0}{\rmdefault}{\mddefault}{\updefault}{\color[rgb]{0,0,0}$n$}%
}}}}
\put(8371,-3121){\makebox(0,0)[b]{\smash{{\SetFigFont{12}{24.0}{\rmdefault}{\mddefault}{\updefault}{\color[rgb]{0,0,0}$1$}%
}}}}
\put(6211,-3121){\makebox(0,0)[b]{\smash{{\SetFigFont{12}{24.0}{\rmdefault}{\mddefault}{\updefault}{\color[rgb]{0,0,0}$k$}%
}}}}
\put(6211,-3616){\makebox(0,0)[b]{\smash{{\SetFigFont{12}{24.0}{\rmdefault}{\mddefault}{\updefault}{\color[rgb]{0,0,0}$h$}%
}}}}
\put(5941,-3256){\makebox(0,0)[b]{\smash{{\SetFigFont{12}{24.0}{\rmdefault}{\mddefault}{\updefault}{\color[rgb]{0,0,0}$?$}%
}}}}
\put(8911,-3121){\makebox(0,0)[b]{\smash{{\SetFigFont{12}{24.0}{\rmdefault}{\mddefault}{\updefault}{\color[rgb]{0,0,0}$k$}%
}}}}
\put(8911,-3616){\makebox(0,0)[b]{\smash{{\SetFigFont{12}{24.0}{\rmdefault}{\mddefault}{\updefault}{\color[rgb]{0,0,0}$h$}%
}}}}
\put(8641,-3256){\makebox(0,0)[b]{\smash{{\SetFigFont{12}{24.0}{\rmdefault}{\mddefault}{\updefault}{\color[rgb]{0,0,0}$?$}%
}}}}
\put(5131,-2896){\makebox(0,0)[b]{\smash{{\SetFigFont{12}{24.0}{\rmdefault}{\mddefault}{\updefault}{\color[rgb]{0,0,0}$n>1$}%
}}}}
\put(5131,-3616){\makebox(0,0)[b]{\smash{{\SetFigFont{12}{24.0}{\rmdefault}{\mddefault}{\updefault}{\color[rgb]{0,0,0}$m>1$}%
}}}}
\end{picture}%
 %
\nop
{
n>1
(O)--X--\ 1    k            n      1     k
         O--?--O     ===>  (O)--X--O--?--O
(O)--X--/ \----O                    \----O
m>1            h                         h
}

Since by assumption $h$ if the minimal value of nodes connected to the node of
value $1$, minimality of the edge of values $1,k$ only depends on whether $k=h$
or not, in both the original and modified graph. This condition is not altered
by the transformation.

\

\noindent{\bf Merging of nodes of equal values, greater than one.} In this
case, no edge is added or removed. The point is therefore only to prove that
selection of an edge touching one of the two nodes is unaffected by the
transformation.

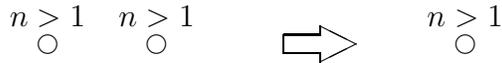
\setlength{\unitlength}{5000sp}%
\begingroup\makeatletter\ifx\SetFigFont\undefined%
\gdef\SetFigFont#1#2#3#4#5{%
  \reset@font\fontsize{#1}{#2pt}%
  \fontfamily{#3}\fontseries{#4}\fontshape{#5}%
  \selectfont}%
\fi\endgroup%
\begin{picture}(2176,342)(4718,-3403)
{\color[rgb]{0,0,0}\thinlines
\put(4771,-3301){\circle{90}}
}%
{\color[rgb]{0,0,0}\put(5311,-3301){\circle{90}}
}%
{\color[rgb]{0,0,0}\put(6841,-3301){\circle{90}}
}%
{\color[rgb]{0,0,0}\put(6301,-3301){\line(-2, 1){180}}
\put(6121,-3211){\line( 0,-1){ 45}}
\put(6121,-3256){\line(-1, 0){180}}
\put(5941,-3256){\line( 0,-1){ 90}}
\put(5941,-3346){\line( 1, 0){180}}
\put(6121,-3346){\line( 0,-1){ 45}}
\put(6121,-3391){\line( 2, 1){180}}
}%
\put(4771,-3211){\makebox(0,0)[b]{\smash{{\SetFigFont{12}{24.0}{\rmdefault}{\mddefault}{\updefault}{\color[rgb]{0,0,0}$n>1$}%
}}}}
\put(5311,-3211){\makebox(0,0)[b]{\smash{{\SetFigFont{12}{24.0}{\rmdefault}{\mddefault}{\updefault}{\color[rgb]{0,0,0}$n>1$}%
}}}}
\put(6841,-3211){\makebox(0,0)[b]{\smash{{\SetFigFont{12}{24.0}{\rmdefault}{\mddefault}{\updefault}{\color[rgb]{0,0,0}$n>1$}%
}}}}
\end{picture}%
 %
\nop
{
n>1    n>1            n>1
 O      O      ===>    O
}

Edges not touching any of the two nodes are unaffected by
Lemma~\ref{influence}. Regarding the ones that touch it, the following figure
exemplifies the situation.

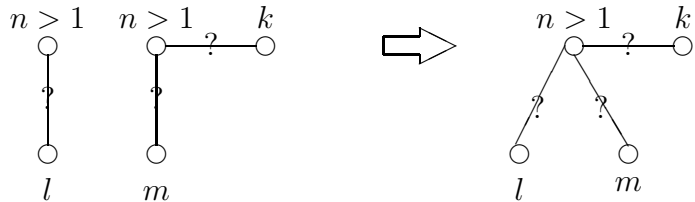
\setlength{\unitlength}{5000sp}%
\begingroup\makeatletter\ifx\SetFigFont\undefined%
\gdef\SetFigFont#1#2#3#4#5{%
  \reset@font\fontsize{#1}{#2pt}%
  \fontfamily{#3}\fontseries{#4}\fontshape{#5}%
  \selectfont}%
\fi\endgroup%
\begin{picture}(3256,1020)(4718,-4081)
{\color[rgb]{0,0,0}\thinlines
\put(4771,-3301){\circle{90}}
}%
{\color[rgb]{0,0,0}\put(5311,-3301){\circle{90}}
}%
{\color[rgb]{0,0,0}\put(5851,-3301){\circle{90}}
}%
{\color[rgb]{0,0,0}\put(4771,-3841){\circle{90}}
}%
{\color[rgb]{0,0,0}\put(5311,-3841){\circle{90}}
}%
{\color[rgb]{0,0,0}\put(7381,-3301){\circle{90}}
}%
{\color[rgb]{0,0,0}\put(7111,-3841){\circle{90}}
}%
{\color[rgb]{0,0,0}\put(7651,-3841){\circle{90}}
}%
{\color[rgb]{0,0,0}\put(7921,-3301){\circle{90}}
}%
{\color[rgb]{0,0,0}\put(5356,-3301){\line( 1, 0){450}}
}%
{\color[rgb]{0,0,0}\put(4771,-3346){\line( 0,-1){450}}
}%
{\color[rgb]{0,0,0}\put(5311,-3346){\line( 0,-1){450}}
}%
{\color[rgb]{0,0,0}\put(7336,-3301){\line(-1,-2){243}}
}%
{\color[rgb]{0,0,0}\put(7381,-3346){\line( 3,-5){270}}
}%
{\color[rgb]{0,0,0}\put(7426,-3301){\line( 1, 0){450}}
}%
{\color[rgb]{0,0,0}\put(6796,-3301){\line(-2, 1){180}}
\put(6616,-3211){\line( 0,-1){ 45}}
\put(6616,-3256){\line(-1, 0){180}}
\put(6436,-3256){\line( 0,-1){ 90}}
\put(6436,-3346){\line( 1, 0){180}}
\put(6616,-3346){\line( 0,-1){ 45}}
\put(6616,-3391){\line( 2, 1){180}}
}%
\put(4771,-3211){\makebox(0,0)[b]{\smash{{\SetFigFont{12}{24.0}{\rmdefault}{\mddefault}{\updefault}{\color[rgb]{0,0,0}$n>1$}%
}}}}
\put(5311,-3211){\makebox(0,0)[b]{\smash{{\SetFigFont{12}{24.0}{\rmdefault}{\mddefault}{\updefault}{\color[rgb]{0,0,0}$n>1$}%
}}}}
\put(5851,-3211){\makebox(0,0)[b]{\smash{{\SetFigFont{12}{24.0}{\rmdefault}{\mddefault}{\updefault}{\color[rgb]{0,0,0}$k$}%
}}}}
\put(4771,-4066){\makebox(0,0)[b]{\smash{{\SetFigFont{12}{24.0}{\rmdefault}{\mddefault}{\updefault}{\color[rgb]{0,0,0}$l$}%
}}}}
\put(5311,-4066){\makebox(0,0)[b]{\smash{{\SetFigFont{12}{24.0}{\rmdefault}{\mddefault}{\updefault}{\color[rgb]{0,0,0}$m$}%
}}}}
\put(7381,-3211){\makebox(0,0)[b]{\smash{{\SetFigFont{12}{24.0}{\rmdefault}{\mddefault}{\updefault}{\color[rgb]{0,0,0}$n>1$}%
}}}}
\put(7921,-3211){\makebox(0,0)[b]{\smash{{\SetFigFont{12}{24.0}{\rmdefault}{\mddefault}{\updefault}{\color[rgb]{0,0,0}$k$}%
}}}}
\put(7111,-4066){\makebox(0,0)[b]{\smash{{\SetFigFont{12}{24.0}{\rmdefault}{\mddefault}{\updefault}{\color[rgb]{0,0,0}$l$}%
}}}}
\put(7651,-4021){\makebox(0,0)[b]{\smash{{\SetFigFont{12}{24.0}{\rmdefault}{\mddefault}{\updefault}{\color[rgb]{0,0,0}$m$}%
}}}}
\put(5581,-3346){\makebox(0,0)[b]{\smash{{\SetFigFont{12}{24.0}{\rmdefault}{\mddefault}{\updefault}{\color[rgb]{0,0,0}$?$}%
}}}}
\put(5311,-3616){\makebox(0,0)[b]{\smash{{\SetFigFont{12}{24.0}{\rmdefault}{\mddefault}{\updefault}{\color[rgb]{0,0,0}$?$}%
}}}}
\put(4771,-3616){\makebox(0,0)[b]{\smash{{\SetFigFont{12}{24.0}{\rmdefault}{\mddefault}{\updefault}{\color[rgb]{0,0,0}$?$}%
}}}}
\put(7201,-3661){\makebox(0,0)[b]{\smash{{\SetFigFont{12}{24.0}{\rmdefault}{\mddefault}{\updefault}{\color[rgb]{0,0,0}$?$}%
}}}}
\put(7516,-3661){\makebox(0,0)[b]{\smash{{\SetFigFont{12}{24.0}{\rmdefault}{\mddefault}{\updefault}{\color[rgb]{0,0,0}$?$}%
}}}}
\put(7651,-3346){\makebox(0,0)[b]{\smash{{\SetFigFont{12}{24.0}{\rmdefault}{\mddefault}{\updefault}{\color[rgb]{0,0,0}$?$}%
}}}}
\end{picture}%
 %
\nop
{
n      n     k              n     k
O      O--?--O      ===>    O--?--O
|      |                   / \                .
?      ?                  ?   ?
|      |                  |   |
O      O                  O   O
l      m                  l   m
}

By Lemma~\ref{influence}, the edge from nodes of values $n$ and $k$ is minimal
or not depending on the value of $n$, but not on the other nodes linked to the
one of value $n$. Therefore, the new link to the node of value $l$ does not
influence to the minimality of the edge.

\

In the following, two transformations are shown that, contrary to the ones
above, do not require any condition on the value of the nodes. They can be
therefore applied to se graphs that are totally unassigned.

\subsubsection{Unassigned graphs transformations}

The simplifications in the previous section assume knowledge of the values of
nodes in the part of the graph to be changed. Some transformations that can be
applied to unassigned se graphs are now presented. Contrary to the ones in the
previous sections, these apply to nodes that are not assigned yet. They are
valid no matter which values these nodes may take: they map obtainable graphs
into obtainable graphs, and unobtainable graphs into unobtainable graphs.

\begin{definition}

The {\bf full disconnection} of a node that is only touched by excluded edges
is the replacement of the node with one for each of these edges.

\end{definition}

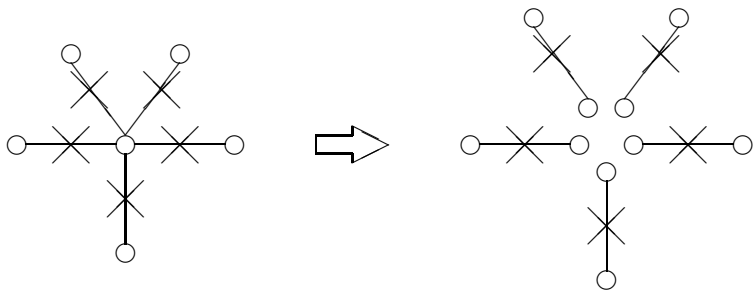
\setlength{\unitlength}{5000sp}%
\begingroup\makeatletter\ifx\SetFigFont\undefined%
\gdef\SetFigFont#1#2#3#4#5{%
  \reset@font\fontsize{#1}{#2pt}%
  \fontfamily{#3}\fontseries{#4}\fontshape{#5}%
  \selectfont}%
\fi\endgroup%
\begin{picture}(3706,1411)(4988,-4209)
\thinlines
{\color[rgb]{0,0,0}\put(5221,-3391){\line( 1,-1){180}}
}%
{\color[rgb]{0,0,0}\put(5221,-3571){\line( 1, 1){180}}
}%
{\color[rgb]{0,0,0}\put(5311,-3121){\line( 1,-1){180}}
}%
{\color[rgb]{0,0,0}\put(5311,-3301){\line( 1, 1){180}}
}%
{\color[rgb]{0,0,0}\put(5671,-3121){\line( 1,-1){180}}
}%
{\color[rgb]{0,0,0}\put(5671,-3301){\line( 1, 1){180}}
}%
{\color[rgb]{0,0,0}\put(5761,-3391){\line( 1,-1){180}}
}%
{\color[rgb]{0,0,0}\put(5761,-3571){\line( 1, 1){180}}
}%
{\color[rgb]{0,0,0}\put(5491,-3661){\line( 1,-1){180}}
}%
{\color[rgb]{0,0,0}\put(5491,-3841){\line( 1, 1){180}}
}%
{\color[rgb]{0,0,0}\put(7471,-3391){\line( 1,-1){180}}
}%
{\color[rgb]{0,0,0}\put(7471,-3571){\line( 1, 1){180}}
}%
{\color[rgb]{0,0,0}\put(7606,-2941){\line( 1,-1){180}}
}%
{\color[rgb]{0,0,0}\put(7606,-3121){\line( 1, 1){180}}
}%
{\color[rgb]{0,0,0}\put(7876,-3796){\line( 1,-1){180}}
}%
{\color[rgb]{0,0,0}\put(7876,-3976){\line( 1, 1){180}}
}%
{\color[rgb]{0,0,0}\put(8281,-3391){\line( 1,-1){180}}
}%
{\color[rgb]{0,0,0}\put(8281,-3571){\line( 1, 1){180}}
}%
{\color[rgb]{0,0,0}\put(8146,-2941){\line( 1,-1){180}}
}%
{\color[rgb]{0,0,0}\put(8146,-3121){\line( 1, 1){180}}
}%
{\color[rgb]{0,0,0}\put(5581,-3481){\circle{90}}
}%
{\color[rgb]{0,0,0}\put(5041,-3481){\circle{90}}
}%
{\color[rgb]{0,0,0}\put(6121,-3481){\circle{90}}
}%
{\color[rgb]{0,0,0}\put(5581,-4021){\circle{90}}
}%
{\color[rgb]{0,0,0}\put(7291,-3481){\circle{90}}
}%
{\color[rgb]{0,0,0}\put(7831,-3481){\circle{90}}
}%
{\color[rgb]{0,0,0}\put(8101,-3481){\circle{90}}
}%
{\color[rgb]{0,0,0}\put(7966,-3616){\circle{90}}
}%
{\color[rgb]{0,0,0}\put(7876,-3301){\circle{90}}
}%
{\color[rgb]{0,0,0}\put(8056,-3301){\circle{90}}
}%
{\color[rgb]{0,0,0}\put(8641,-3481){\circle{90}}
}%
{\color[rgb]{0,0,0}\put(7966,-4156){\circle{90}}
}%
{\color[rgb]{0,0,0}\put(5851,-3031){\circle{90}}
}%
{\color[rgb]{0,0,0}\put(5311,-3031){\circle{90}}
}%
{\color[rgb]{0,0,0}\put(7606,-2851){\circle{90}}
}%
{\color[rgb]{0,0,0}\put(8326,-2851){\circle{90}}
}%
{\color[rgb]{0,0,0}\put(5536,-3481){\line(-1, 0){450}}
}%
{\color[rgb]{0,0,0}\put(5626,-3481){\line( 1, 0){450}}
}%
{\color[rgb]{0,0,0}\put(5581,-3526){\line( 0,-1){450}}
}%
{\color[rgb]{0,0,0}\put(5311,-3076){\line( 3,-4){270}}
}%
{\color[rgb]{0,0,0}\put(5581,-3436){\line( 3, 4){270}}
}%
{\color[rgb]{0,0,0}\put(7336,-3481){\line( 1, 0){450}}
}%
{\color[rgb]{0,0,0}\put(8146,-3481){\line( 1, 0){450}}
}%
{\color[rgb]{0,0,0}\put(7966,-3661){\line( 0,-1){450}}
}%
{\color[rgb]{0,0,0}\put(7606,-2896){\line( 3,-4){270}}
}%
{\color[rgb]{0,0,0}\put(8056,-3256){\line( 3, 4){270}}
}%
{\color[rgb]{0,0,0}\put(6886,-3481){\line(-2, 1){180}}
\put(6706,-3391){\line( 0,-1){ 45}}
\put(6706,-3436){\line(-1, 0){180}}
\put(6526,-3436){\line( 0,-1){ 90}}
\put(6526,-3526){\line( 1, 0){180}}
\put(6706,-3526){\line( 0,-1){ 45}}
\put(6706,-3571){\line( 2, 1){180}}
}%
\end{picture}%
 %
\nop
{
                          O         O
                           \       /
O       O                   X     X
 \     /                     \   /
  X   X                       O O
   \ /
O-X-O-X-O      ===>       O-X-O O-X-O
    |
    X                          O
    |                          |
    O                          X
                               |
                               O
}

\begin{lemma}
\label{full}

Full disconnection maps obtainable graphs into obtainable graphs, and vice
versa.

\end{lemma}

\proof The claim is proved by showing how to map values of the original node to
values of its copies in the disconnected version of the graph. This is done as
follows: the single value is assigned to the copies; vice versa, if the copies
have different values, set the original to their maximum.

As a preliminary result, if the value of the node of a non-minimal edge is
increased, the edge remains non-minimal.

If the original graph is obtainable, there exists at least an extension of it
to a totally assigned se graph that is obtainable. If the central node has
value one, it is changed to two; the graph remains obtainable. The nodes of the
graph that results from the transformation are assigned as follows: the copies
of the node that is broken get the same value of the original node; all other
values are left unchanged. By Lemma~\ref{influence}, these edges remain
non-minimal, as they are still connected to a node of the same value greater
than one. The edges connected to them are not changed either: even if the other
node is assigned one, it is still connected to a node of the same value.

If the resulting graph is obtainable, it has at least an extension to an
obtainable totally assigned se graph. The nodes that result from the
disconnection may have the same value or not, and these values may even be all
one. In the latter case, these values are all changed to two. Otherwise, they
are all changed to the maximum of these values. This way, all these nodes are
set to the same value. The original graph is then assigned values as follows:
the node that was broken is assigned to the value of the resulting nodes; all
others are the same. By Lemma~\ref{influence}, all edges touching the broken
node remain non-minimal because they are still connected to a node of the same
value greater than one; the other edges remains minimal or not for the same
reason of the previous case.~\qed

The second transformation is about the removal of edges that do not participate
in any cycle. Such edges form chains that may be isolated to the rest of the
graph, or connected by one node only.

\begin{definition}

The {\bf removal of a tail} is the deletion of a chain of edges that do not
participate in any cycle.

\end{definition}

Removing all such edges leads to a graph where every edge is part of some
cycle.

\begin{lemma}
\label{tail}

Removing tails does not alter obtainability.

\end{lemma}

\proof The claim is proved for tails comprising a single edge. Longer tails can
be dealt with by removing edges one at time, from the end to the beginning.
That tails end is a consequence of the finiteness of the graphs and the lack of
cycles containing them.

Removing an edge release a constraint: the edge is no longer required to be
minimal if selected and non-minimal if excluded. As a result, if the original
graph is obtainable, so is the one resulting from the transformation. Remains
to prove the other direction: if the graph resulting from the removal is
obtainable, the edge can be added back without violating obtainability.

If the graph after removal is obtainable, an obtainable totally assigned se
graph extending it exists. Recovering the removed edge introduces either o node
or two. It is shown that these can be assigned values so that obtainability is
maintained.

The case of two nodes added back is only possible if the edge is connected to
none else. In this case, the values can be set to both one for a selected edge
or two for an excluded one.

In the other case, one of the nodes is also in the graph after removal, so it
has a values. This could be equal to one or greater. In the first case, the
edge could be selected or excluded. This leads to three possible cases, the
first being:

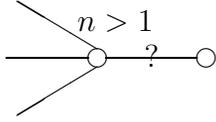
\setlength{\unitlength}{5000sp}%
\begingroup\makeatletter\ifx\SetFigFont\undefined%
\gdef\SetFigFont#1#2#3#4#5{%
  \reset@font\fontsize{#1}{#2pt}%
  \fontfamily{#3}\fontseries{#4}\fontshape{#5}%
  \selectfont}%
\fi\endgroup%
\begin{picture}(1055,567)(4669,-3673)
{\color[rgb]{0,0,0}\thinlines
\put(5131,-3391){\circle{90}}
}%
{\color[rgb]{0,0,0}\put(5671,-3391){\circle{90}}
}%
{\color[rgb]{0,0,0}\put(5176,-3391){\line( 1, 0){450}}
}%
{\color[rgb]{0,0,0}\put(5131,-3346){\line(-5, 3){397.059}}
}%
{\color[rgb]{0,0,0}\put(5086,-3391){\line(-1, 0){405}}
}%
{\color[rgb]{0,0,0}\put(5131,-3436){\line(-5,-3){397.059}}
}%
\put(5401,-3436){\makebox(0,0)[b]{\smash{{\SetFigFont{12}{24.0}{\rmdefault}{\mddefault}{\updefault}{\color[rgb]{0,0,0}$?$}%
}}}}
\put(5221,-3256){\makebox(0,0)[b]{\smash{{\SetFigFont{12}{24.0}{\rmdefault}{\mddefault}{\updefault}{\color[rgb]{0,0,0}$n>1$}%
}}}}
\end{picture}%
 %
\nop
{
  n>1
...O--?--O
}

A value is to be chosen for the reintroduced right node so that the totally
assigned se graph remains obtainable. By Lemma~\ref{influence}, minimality of
the other edges is not affected by the value of the right node, which can be
therefore set to $1$ if the edge is selected and $2$ if excluded.

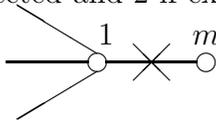
\setlength{\unitlength}{5000sp}%
\begingroup\makeatletter\ifx\SetFigFont\undefined%
\gdef\SetFigFont#1#2#3#4#5{%
  \reset@font\fontsize{#1}{#2pt}%
  \fontfamily{#3}\fontseries{#4}\fontshape{#5}%
  \selectfont}%
\fi\endgroup%
\begin{picture}(1055,564)(4669,-3673)
{\color[rgb]{0,0,0}\thinlines
\put(5131,-3391){\circle{90}}
}%
{\color[rgb]{0,0,0}\put(5671,-3391){\circle{90}}
}%
{\color[rgb]{0,0,0}\put(5176,-3391){\line( 1, 0){450}}
}%
{\color[rgb]{0,0,0}\put(5131,-3346){\line(-5, 3){397.059}}
}%
{\color[rgb]{0,0,0}\put(5086,-3391){\line(-1, 0){405}}
}%
{\color[rgb]{0,0,0}\put(5131,-3436){\line(-5,-3){397.059}}
}%
{\color[rgb]{0,0,0}\put(5311,-3301){\line( 1,-1){180}}
}%
{\color[rgb]{0,0,0}\put(5311,-3481){\line( 1, 1){180}}
}%
\put(5176,-3301){\makebox(0,0)[b]{\smash{{\SetFigFont{12}{24.0}{\rmdefault}{\mddefault}{\updefault}{\color[rgb]{0,0,0}$1$}%
}}}}
\put(5671,-3301){\makebox(0,0)[b]{\smash{{\SetFigFont{12}{24.0}{\rmdefault}{\mddefault}{\updefault}{\color[rgb]{0,0,0}$m$}%
}}}}
\end{picture}%
 %
\nop
{
   1     m
...O--X--O
}

In this second case, the left node is one and the edge is excluded: the other
node is assigned to a value that is greater than all other nodes connected to
the left one.

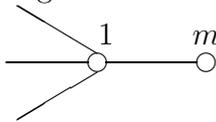
\setlength{\unitlength}{5000sp}%
\begingroup\makeatletter\ifx\SetFigFont\undefined%
\gdef\SetFigFont#1#2#3#4#5{%
  \reset@font\fontsize{#1}{#2pt}%
  \fontfamily{#3}\fontseries{#4}\fontshape{#5}%
  \selectfont}%
\fi\endgroup%
\begin{picture}(1055,564)(4669,-3673)
{\color[rgb]{0,0,0}\thinlines
\put(5131,-3391){\circle{90}}
}%
{\color[rgb]{0,0,0}\put(5671,-3391){\circle{90}}
}%
{\color[rgb]{0,0,0}\put(5176,-3391){\line( 1, 0){450}}
}%
{\color[rgb]{0,0,0}\put(5131,-3346){\line(-5, 3){397.059}}
}%
{\color[rgb]{0,0,0}\put(5086,-3391){\line(-1, 0){405}}
}%
{\color[rgb]{0,0,0}\put(5131,-3436){\line(-5,-3){397.059}}
}%
\put(5176,-3301){\makebox(0,0)[b]{\smash{{\SetFigFont{12}{24.0}{\rmdefault}{\mddefault}{\updefault}{\color[rgb]{0,0,0}$1$}%
}}}}
\put(5671,-3301){\makebox(0,0)[b]{\smash{{\SetFigFont{12}{24.0}{\rmdefault}{\mddefault}{\updefault}{\color[rgb]{0,0,0}$m$}%
}}}}
\end{picture}%
 %
\nop
{
   1     m
...O-----O
}

This is the third case. If the node of value $1$ is connected via another
selected edge to a node of value $n$, set $m=n$. If it is only touched by
excluded edges, set $m=1$.~\qed

Another transformation is the zigzag folding, where a chain of selected edges
is reduced to a single one by merging the first, third, fifth, etc. node of the
chain and the second, fourth, etc.

Correctness is proved in two steps: first, a sequence of selected edges has
alternating values ($n-m-n-m-\cdots$) in every obtainable totally assigned se
graph; second, by a sequence of transformations, this result is used to prove
that the sequence can be folded into a single selected edge.

\begin{lemma}
\label{alternating}

The nodes of a chain of selected edges in a totally assigned obtainable graph
has alternating values, that is, $n-m-n-m-n-m-\cdots$.

\end{lemma}

\proof Let $n$, $m$ and $k$ be the values of three consecutive nodes of the
chain. The claim follows from $k=n$ for every possible values of $n$ and $m$.

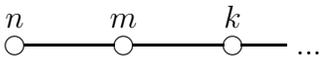
\setlength{\unitlength}{5000sp}%
\begingroup\makeatletter\ifx\SetFigFont\undefined%
\gdef\SetFigFont#1#2#3#4#5{%
  \reset@font\fontsize{#1}{#2pt}%
  \fontfamily{#3}\fontseries{#4}\fontshape{#5}%
  \selectfont}%
\fi\endgroup%
\begin{picture}(1463,300)(4448,-3541)
{\color[rgb]{0,0,0}\thinlines
\put(4501,-3481){\circle{90}}
}%
{\color[rgb]{0,0,0}\put(5041,-3481){\circle{90}}
}%
{\color[rgb]{0,0,0}\put(5581,-3481){\circle{90}}
}%
{\color[rgb]{0,0,0}\put(4546,-3481){\line( 1, 0){450}}
}%
{\color[rgb]{0,0,0}\put(5086,-3481){\line( 1, 0){450}}
}%
{\color[rgb]{0,0,0}\put(5626,-3481){\line( 1, 0){225}}
}%
\put(5896,-3526){\makebox(0,0)[lb]{\smash{{\SetFigFont{12}{24.0}{\rmdefault}{\mddefault}{\updefault}{\color[rgb]{0,0,0}$...$}%
}}}}
\put(4501,-3391){\makebox(0,0)[b]{\smash{{\SetFigFont{12}{24.0}{\rmdefault}{\mddefault}{\updefault}{\color[rgb]{0,0,0}$n$}%
}}}}
\put(5041,-3391){\makebox(0,0)[b]{\smash{{\SetFigFont{12}{24.0}{\rmdefault}{\mddefault}{\updefault}{\color[rgb]{0,0,0}$m$}%
}}}}
\put(5581,-3391){\makebox(0,0)[b]{\smash{{\SetFigFont{12}{24.0}{\rmdefault}{\mddefault}{\updefault}{\color[rgb]{0,0,0}$k$}%
}}}}
\end{picture}%
 %
\nop
{
 n     m     k
 O-----O-----O-- ...
}

Various cases are possible:

\begin{itemize}

\item $n>1$: by Lemma~\ref{first}, in every selected edge at least one node has
value $1$; therefore, $m=1$; if $k<n$, then $(1,k)$ is preferred over $(1,n)$;
if $l>n$, the converse happens; since both edges are selected, $k=n$;

\item $n=1$, $m=1$: if $k$ greater than $1$, then $(1,1)$ is preferred over
$(1,k)$; therefore, $k=1$;

\item $n=1$, $m>1$: the edge values $(m,k)$ is selected; by Lemma~\ref{first},
one between $m$ and $k$ is $1$; since $m>1$, if follows that $k=1$, which is
the same as $n$.

\end{itemize}

Since the alternation holds for every triple of consecutive nodes, it holds for
the whole chain.~\qed

This property implies that, regardless of the values of the other nodes of the
graph, the only way to produce a correct assignment is by setting the nodes of
the chain to values that alternate between two values.

\begin{definition}

Given a se graph, a {\em zigzag folding} of a chain of selected edges is the
merging of all nodes of odd position and nodes of even position.

\end{definition}

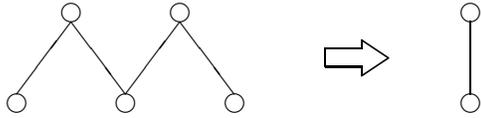
\setlength{\unitlength}{5000sp}%
\begingroup\makeatletter\ifx\SetFigFont\undefined%
\gdef\SetFigFont#1#2#3#4#5{%
  \reset@font\fontsize{#1}{#2pt}%
  \fontfamily{#3}\fontseries{#4}\fontshape{#5}%
  \selectfont}%
\fi\endgroup%
\begin{picture}(2356,556)(4808,-3714)
{\color[rgb]{0,0,0}\thinlines
\put(4861,-3661){\circle{90}}
}%
{\color[rgb]{0,0,0}\put(5131,-3211){\circle{90}}
}%
{\color[rgb]{0,0,0}\put(5401,-3661){\circle{90}}
}%
{\color[rgb]{0,0,0}\put(5671,-3211){\circle{90}}
}%
{\color[rgb]{0,0,0}\put(5941,-3661){\circle{90}}
}%
{\color[rgb]{0,0,0}\put(7111,-3661){\circle{90}}
}%
{\color[rgb]{0,0,0}\put(7111,-3211){\circle{90}}
}%
{\color[rgb]{0,0,0}\put(4861,-3616){\line( 3, 4){270}}
}%
{\color[rgb]{0,0,0}\put(5131,-3256){\line( 3,-4){270}}
}%
{\color[rgb]{0,0,0}\put(5401,-3616){\line( 3, 4){270}}
}%
{\color[rgb]{0,0,0}\put(5671,-3256){\line( 3,-4){270}}
}%
{\color[rgb]{0,0,0}\put(6706,-3436){\line(-3, 2){135}}
\put(6571,-3346){\line( 0,-1){ 45}}
\put(6571,-3391){\line(-1, 0){180}}
\put(6391,-3391){\line( 0,-1){ 90}}
\put(6391,-3481){\line( 1, 0){180}}
\put(6571,-3481){\line( 0,-1){ 45}}
\put(6571,-3526){\line( 3, 2){135}}
}%
{\color[rgb]{0,0,0}\put(7111,-3256){\line( 0,-1){360}}
}%
\end{picture}%
 %
\nop
{
  O   O                 O
 / \ / \ /     ===>     |
O   O   O               O
}

\begin{lemma}
\label{zigzag}

The zigzag folding maps obtainable graphs into obtainable graphs and vice
versa.

\end{lemma}

\proof In every totally assigned se graph extending the given one, the nodes of
the chain have alternating values $n-m-n-m-\cdots$ by Lemma~\ref{alternating}.
By Lemma~\ref{first}, one between $n$ and $m$ is one. The other may be one or
greater.

Let $n=1$ and $m>1$. Disconnecting all edges of the chain produces:

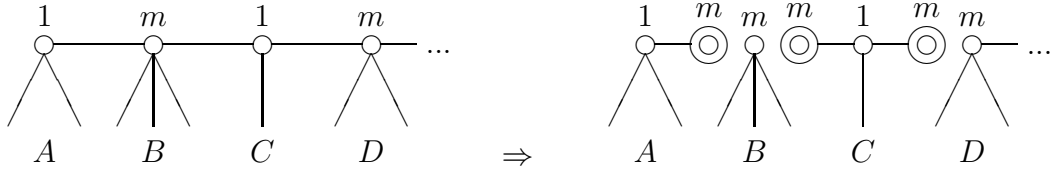
\ttytex{
\begin{tabular}{ccc}
\setlength{\unitlength}{5000sp}%
\begingroup\makeatletter\ifx\SetFigFont\undefined%
\gdef\SetFigFont#1#2#3#4#5{%
  \reset@font\fontsize{#1}{#2pt}%
  \fontfamily{#3}\fontseries{#4}\fontshape{#5}%
  \selectfont}%
\fi\endgroup%
\begin{picture}(2097,840)(4309,-4171)
{\color[rgb]{0,0,0}\thinlines
\put(4501,-3571){\circle{90}}
}%
{\color[rgb]{0,0,0}\put(5041,-3571){\circle{90}}
}%
{\color[rgb]{0,0,0}\put(5581,-3571){\circle{90}}
}%
{\color[rgb]{0,0,0}\put(6121,-3571){\circle{90}}
}%
{\color[rgb]{0,0,0}\put(4546,-3571){\line( 1, 0){450}}
}%
{\color[rgb]{0,0,0}\put(5086,-3571){\line( 1, 0){450}}
}%
{\color[rgb]{0,0,0}\put(5626,-3571){\line( 1, 0){450}}
}%
{\color[rgb]{0,0,0}\put(6166,-3571){\line( 1, 0){180}}
}%
{\color[rgb]{0,0,0}\put(4501,-3616){\line(-1,-2){180}}
}%
{\color[rgb]{0,0,0}\put(4501,-3616){\line( 1,-2){180}}
}%
{\color[rgb]{0,0,0}\put(5581,-3616){\line( 0,-1){360}}
}%
{\color[rgb]{0,0,0}\put(6121,-3616){\line(-1,-2){180}}
}%
{\color[rgb]{0,0,0}\put(6121,-3616){\line( 1,-2){180}}
}%
{\color[rgb]{0,0,0}\put(5041,-3616){\line( 0,-1){360}}
}%
{\color[rgb]{0,0,0}\put(5041,-3616){\line( 1,-2){180}}
}%
{\color[rgb]{0,0,0}\put(5041,-3616){\line(-1,-2){180}}
}%
\put(6391,-3616){\makebox(0,0)[lb]{\smash{{\SetFigFont{12}{24.0}{\rmdefault}{\mddefault}{\updefault}{\color[rgb]{0,0,0}$...$}%
}}}}
\put(4501,-3481){\makebox(0,0)[b]{\smash{{\SetFigFont{12}{24.0}{\rmdefault}{\mddefault}{\updefault}{\color[rgb]{0,0,0}$1$}%
}}}}
\put(5041,-3481){\makebox(0,0)[b]{\smash{{\SetFigFont{12}{24.0}{\rmdefault}{\mddefault}{\updefault}{\color[rgb]{0,0,0}$m$}%
}}}}
\put(5581,-3481){\makebox(0,0)[b]{\smash{{\SetFigFont{12}{24.0}{\rmdefault}{\mddefault}{\updefault}{\color[rgb]{0,0,0}$1$}%
}}}}
\put(6121,-3481){\makebox(0,0)[b]{\smash{{\SetFigFont{12}{24.0}{\rmdefault}{\mddefault}{\updefault}{\color[rgb]{0,0,0}$m$}%
}}}}
\put(4501,-4156){\makebox(0,0)[b]{\smash{{\SetFigFont{12}{24.0}{\rmdefault}{\mddefault}{\updefault}{\color[rgb]{0,0,0}$A$}%
}}}}
\put(5041,-4156){\makebox(0,0)[b]{\smash{{\SetFigFont{12}{24.0}{\rmdefault}{\mddefault}{\updefault}{\color[rgb]{0,0,0}$B$}%
}}}}
\put(5581,-4156){\makebox(0,0)[b]{\smash{{\SetFigFont{12}{24.0}{\rmdefault}{\mddefault}{\updefault}{\color[rgb]{0,0,0}$C$}%
}}}}
\put(6121,-4156){\makebox(0,0)[b]{\smash{{\SetFigFont{12}{24.0}{\rmdefault}{\mddefault}{\updefault}{\color[rgb]{0,0,0}$D$}%
}}}}
\end{picture}%
 %
 %
& ~ ~ $\Rightarrow$ ~ ~ &
\setlength{\unitlength}{5000sp}%
\begingroup\makeatletter\ifx\SetFigFont\undefined%
\gdef\SetFigFont#1#2#3#4#5{%
  \reset@font\fontsize{#1}{#2pt}%
  \fontfamily{#3}\fontseries{#4}\fontshape{#5}%
  \selectfont}%
\fi\endgroup%
\begin{picture}(2097,840)(4309,-4171)
{\color[rgb]{0,0,0}\thinlines
\put(4501,-3571){\circle{90}}
}%
{\color[rgb]{0,0,0}\put(5041,-3571){\circle{90}}
}%
{\color[rgb]{0,0,0}\put(5581,-3571){\circle{90}}
}%
{\color[rgb]{0,0,0}\put(6121,-3571){\circle{90}}
}%
{\color[rgb]{0,0,0}\put(4816,-3571){\circle{90}}
}%
{\color[rgb]{0,0,0}\put(4816,-3571){\circle{180}}
}%
{\color[rgb]{0,0,0}\put(5266,-3571){\circle{90}}
}%
{\color[rgb]{0,0,0}\put(5266,-3571){\circle{180}}
}%
{\color[rgb]{0,0,0}\put(5896,-3571){\circle{90}}
}%
{\color[rgb]{0,0,0}\put(5896,-3571){\circle{180}}
}%
{\color[rgb]{0,0,0}\put(4501,-3616){\line(-1,-2){180}}
}%
{\color[rgb]{0,0,0}\put(4501,-3616){\line( 1,-2){180}}
}%
{\color[rgb]{0,0,0}\put(5041,-3616){\line(-1,-2){180}}
}%
{\color[rgb]{0,0,0}\put(5041,-3616){\line( 0,-1){360}}
}%
{\color[rgb]{0,0,0}\put(5041,-3616){\line( 1,-2){180}}
}%
{\color[rgb]{0,0,0}\put(5581,-3616){\line( 0,-1){360}}
}%
{\color[rgb]{0,0,0}\put(6121,-3616){\line(-1,-2){180}}
}%
{\color[rgb]{0,0,0}\put(6121,-3616){\line( 1,-2){180}}
}%
{\color[rgb]{0,0,0}\put(4546,-3571){\line( 1, 0){180}}
}%
{\color[rgb]{0,0,0}\put(5356,-3571){\line( 1, 0){180}}
}%
{\color[rgb]{0,0,0}\put(5626,-3571){\line( 1, 0){180}}
}%
{\color[rgb]{0,0,0}\put(6166,-3571){\line( 1, 0){180}}
}%
\put(6391,-3616){\makebox(0,0)[lb]{\smash{{\SetFigFont{12}{24.0}{\rmdefault}{\mddefault}{\updefault}{\color[rgb]{0,0,0}$...$}%
}}}}
\put(4501,-3481){\makebox(0,0)[b]{\smash{{\SetFigFont{12}{24.0}{\rmdefault}{\mddefault}{\updefault}{\color[rgb]{0,0,0}$1$}%
}}}}
\put(5041,-3481){\makebox(0,0)[b]{\smash{{\SetFigFont{12}{24.0}{\rmdefault}{\mddefault}{\updefault}{\color[rgb]{0,0,0}$m$}%
}}}}
\put(5581,-3481){\makebox(0,0)[b]{\smash{{\SetFigFont{12}{24.0}{\rmdefault}{\mddefault}{\updefault}{\color[rgb]{0,0,0}$1$}%
}}}}
\put(6121,-3481){\makebox(0,0)[b]{\smash{{\SetFigFont{12}{24.0}{\rmdefault}{\mddefault}{\updefault}{\color[rgb]{0,0,0}$m$}%
}}}}
\put(4816,-3436){\makebox(0,0)[b]{\smash{{\SetFigFont{12}{24.0}{\rmdefault}{\mddefault}{\updefault}{\color[rgb]{0,0,0}$m$}%
}}}}
\put(5266,-3436){\makebox(0,0)[b]{\smash{{\SetFigFont{12}{24.0}{\rmdefault}{\mddefault}{\updefault}{\color[rgb]{0,0,0}$m$}%
}}}}
\put(5896,-3436){\makebox(0,0)[b]{\smash{{\SetFigFont{12}{24.0}{\rmdefault}{\mddefault}{\updefault}{\color[rgb]{0,0,0}$m$}%
}}}}
\put(4501,-4156){\makebox(0,0)[b]{\smash{{\SetFigFont{12}{24.0}{\rmdefault}{\mddefault}{\updefault}{\color[rgb]{0,0,0}$A$}%
}}}}
\put(5041,-4156){\makebox(0,0)[b]{\smash{{\SetFigFont{12}{24.0}{\rmdefault}{\mddefault}{\updefault}{\color[rgb]{0,0,0}$B$}%
}}}}
\put(5581,-4156){\makebox(0,0)[b]{\smash{{\SetFigFont{12}{24.0}{\rmdefault}{\mddefault}{\updefault}{\color[rgb]{0,0,0}$C$}%
}}}}
\put(6121,-4156){\makebox(0,0)[b]{\smash{{\SetFigFont{12}{24.0}{\rmdefault}{\mddefault}{\updefault}{\color[rgb]{0,0,0}$D$}%
}}}}
\end{picture}%
 %
 %
\end{tabular}
}{
(A)   (B)   (C)   (D)              (A)       (B)       (C)       (D)
 O-----O-----O-----O--...   ===>    O----(O)  O  (O)----O----(O)  O ...
 1     m     1     m                1     m   m   m     1     m   m 
}

In this figure, $A$ indicates the connections of the first node of the chain,
$B$ to the second, etc. Merging of selected edges and nodes of value greater
than one collapse the nodes into two ones:

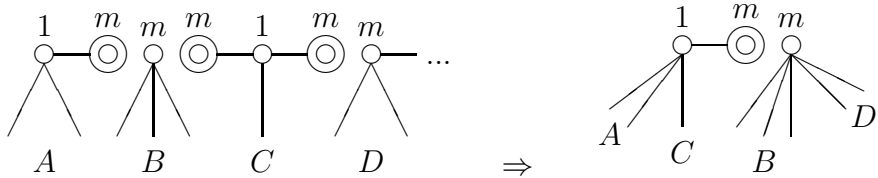
\ttytex{
\begin{tabular}{ccc}
\setlength{\unitlength}{5000sp}%
\begingroup\makeatletter\ifx\SetFigFont\undefined%
\gdef\SetFigFont#1#2#3#4#5{%
  \reset@font\fontsize{#1}{#2pt}%
  \fontfamily{#3}\fontseries{#4}\fontshape{#5}%
  \selectfont}%
\fi\endgroup%
\begin{picture}(2097,840)(4309,-4171)
{\color[rgb]{0,0,0}\thinlines
\put(4501,-3571){\circle{90}}
}%
{\color[rgb]{0,0,0}\put(5041,-3571){\circle{90}}
}%
{\color[rgb]{0,0,0}\put(5581,-3571){\circle{90}}
}%
{\color[rgb]{0,0,0}\put(6121,-3571){\circle{90}}
}%
{\color[rgb]{0,0,0}\put(4816,-3571){\circle{90}}
}%
{\color[rgb]{0,0,0}\put(4816,-3571){\circle{180}}
}%
{\color[rgb]{0,0,0}\put(5266,-3571){\circle{90}}
}%
{\color[rgb]{0,0,0}\put(5266,-3571){\circle{180}}
}%
{\color[rgb]{0,0,0}\put(5896,-3571){\circle{90}}
}%
{\color[rgb]{0,0,0}\put(5896,-3571){\circle{180}}
}%
{\color[rgb]{0,0,0}\put(4501,-3616){\line(-1,-2){180}}
}%
{\color[rgb]{0,0,0}\put(4501,-3616){\line( 1,-2){180}}
}%
{\color[rgb]{0,0,0}\put(5041,-3616){\line(-1,-2){180}}
}%
{\color[rgb]{0,0,0}\put(5041,-3616){\line( 0,-1){360}}
}%
{\color[rgb]{0,0,0}\put(5041,-3616){\line( 1,-2){180}}
}%
{\color[rgb]{0,0,0}\put(5581,-3616){\line( 0,-1){360}}
}%
{\color[rgb]{0,0,0}\put(6121,-3616){\line(-1,-2){180}}
}%
{\color[rgb]{0,0,0}\put(6121,-3616){\line( 1,-2){180}}
}%
{\color[rgb]{0,0,0}\put(4546,-3571){\line( 1, 0){180}}
}%
{\color[rgb]{0,0,0}\put(5356,-3571){\line( 1, 0){180}}
}%
{\color[rgb]{0,0,0}\put(5626,-3571){\line( 1, 0){180}}
}%
{\color[rgb]{0,0,0}\put(6166,-3571){\line( 1, 0){180}}
}%
\put(6391,-3616){\makebox(0,0)[lb]{\smash{{\SetFigFont{12}{24.0}{\rmdefault}{\mddefault}{\updefault}{\color[rgb]{0,0,0}$...$}%
}}}}
\put(4501,-3481){\makebox(0,0)[b]{\smash{{\SetFigFont{12}{24.0}{\rmdefault}{\mddefault}{\updefault}{\color[rgb]{0,0,0}$1$}%
}}}}
\put(5041,-3481){\makebox(0,0)[b]{\smash{{\SetFigFont{12}{24.0}{\rmdefault}{\mddefault}{\updefault}{\color[rgb]{0,0,0}$m$}%
}}}}
\put(5581,-3481){\makebox(0,0)[b]{\smash{{\SetFigFont{12}{24.0}{\rmdefault}{\mddefault}{\updefault}{\color[rgb]{0,0,0}$1$}%
}}}}
\put(6121,-3481){\makebox(0,0)[b]{\smash{{\SetFigFont{12}{24.0}{\rmdefault}{\mddefault}{\updefault}{\color[rgb]{0,0,0}$m$}%
}}}}
\put(4816,-3436){\makebox(0,0)[b]{\smash{{\SetFigFont{12}{24.0}{\rmdefault}{\mddefault}{\updefault}{\color[rgb]{0,0,0}$m$}%
}}}}
\put(5266,-3436){\makebox(0,0)[b]{\smash{{\SetFigFont{12}{24.0}{\rmdefault}{\mddefault}{\updefault}{\color[rgb]{0,0,0}$m$}%
}}}}
\put(5896,-3436){\makebox(0,0)[b]{\smash{{\SetFigFont{12}{24.0}{\rmdefault}{\mddefault}{\updefault}{\color[rgb]{0,0,0}$m$}%
}}}}
\put(4501,-4156){\makebox(0,0)[b]{\smash{{\SetFigFont{12}{24.0}{\rmdefault}{\mddefault}{\updefault}{\color[rgb]{0,0,0}$A$}%
}}}}
\put(5041,-4156){\makebox(0,0)[b]{\smash{{\SetFigFont{12}{24.0}{\rmdefault}{\mddefault}{\updefault}{\color[rgb]{0,0,0}$B$}%
}}}}
\put(5581,-4156){\makebox(0,0)[b]{\smash{{\SetFigFont{12}{24.0}{\rmdefault}{\mddefault}{\updefault}{\color[rgb]{0,0,0}$C$}%
}}}}
\put(6121,-4156){\makebox(0,0)[b]{\smash{{\SetFigFont{12}{24.0}{\rmdefault}{\mddefault}{\updefault}{\color[rgb]{0,0,0}$D$}%
}}}}
\end{picture}%
 %
 %
& ~ ~ $\Rightarrow$ ~ ~ &
\setlength{\unitlength}{5000sp}%
\begingroup\makeatletter\ifx\SetFigFont\undefined%
\gdef\SetFigFont#1#2#3#4#5{%
  \reset@font\fontsize{#1}{#2pt}%
  \fontfamily{#3}\fontseries{#4}\fontshape{#5}%
  \selectfont}%
\fi\endgroup%
\begin{picture}(1290,885)(4126,-4216)
{\color[rgb]{0,0,0}\thinlines
\put(4501,-3571){\circle{90}}
}%
{\color[rgb]{0,0,0}\put(5041,-3571){\circle{90}}
}%
{\color[rgb]{0,0,0}\put(4816,-3571){\circle{90}}
}%
{\color[rgb]{0,0,0}\put(4816,-3571){\circle{180}}
}%
{\color[rgb]{0,0,0}\put(4501,-3616){\line(-4,-3){360}}
}%
{\color[rgb]{0,0,0}\put(4501,-3616){\line(-3,-4){270}}
}%
{\color[rgb]{0,0,0}\put(5041,-3616){\line(-3,-4){270}}
}%
{\color[rgb]{0,0,0}\put(5041,-3616){\line(-1,-3){135}}
}%
{\color[rgb]{0,0,0}\put(5041,-3616){\line( 0,-1){405}}
}%
{\color[rgb]{0,0,0}\put(4546,-3571){\line( 1, 0){180}}
}%
{\color[rgb]{0,0,0}\put(4501,-3616){\line( 0,-1){360}}
}%
{\color[rgb]{0,0,0}\put(5041,-3616){\line( 2,-1){360}}
}%
{\color[rgb]{0,0,0}\put(5041,-3616){\line( 1,-1){270}}
}%
\put(4501,-3481){\makebox(0,0)[b]{\smash{{\SetFigFont{12}{24.0}{\rmdefault}{\mddefault}{\updefault}{\color[rgb]{0,0,0}$1$}%
}}}}
\put(5041,-3481){\makebox(0,0)[b]{\smash{{\SetFigFont{12}{24.0}{\rmdefault}{\mddefault}{\updefault}{\color[rgb]{0,0,0}$m$}%
}}}}
\put(4816,-3436){\makebox(0,0)[b]{\smash{{\SetFigFont{12}{24.0}{\rmdefault}{\mddefault}{\updefault}{\color[rgb]{0,0,0}$m$}%
}}}}
\put(4501,-4156){\makebox(0,0)[b]{\smash{{\SetFigFont{12}{24.0}{\rmdefault}{\mddefault}{\updefault}{\color[rgb]{0,0,0}$C$}%
}}}}
\put(5401,-3976){\makebox(0,0)[b]{\smash{{\SetFigFont{12}{24.0}{\rmdefault}{\mddefault}{\updefault}{\color[rgb]{0,0,0}$D$}%
}}}}
\put(4906,-4201){\makebox(0,0)[b]{\smash{{\SetFigFont{12}{24.0}{\rmdefault}{\mddefault}{\updefault}{\color[rgb]{0,0,0}$B$}%
}}}}
\put(4141,-4066){\makebox(0,0)[b]{\smash{{\SetFigFont{12}{24.0}{\rmdefault}{\mddefault}{\updefault}{\color[rgb]{0,0,0}$A$}%
}}}}
\end{picture}%
 %
 %
\end{tabular}
}{
(1)       (2)       (3)       (4)              (1)(3)       (2)(4)
 O----(O)  O  (O)----O----(O)  O ...   ===>       O----(O)    O
 1     m   m   m     1     m   m                  1     m     m
}

The two nodes of value $m$ can be then merged back by applying disconnection in
reverse:

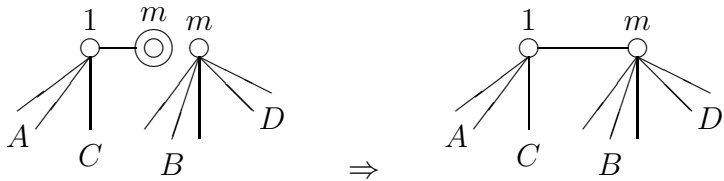
\ttytex{
\begin{tabular}{ccc}
\setlength{\unitlength}{5000sp}%
\begingroup\makeatletter\ifx\SetFigFont\undefined%
\gdef\SetFigFont#1#2#3#4#5{%
  \reset@font\fontsize{#1}{#2pt}%
  \fontfamily{#3}\fontseries{#4}\fontshape{#5}%
  \selectfont}%
\fi\endgroup%
\begin{picture}(1290,885)(4126,-4216)
{\color[rgb]{0,0,0}\thinlines
\put(4501,-3571){\circle{90}}
}%
{\color[rgb]{0,0,0}\put(5041,-3571){\circle{90}}
}%
{\color[rgb]{0,0,0}\put(4816,-3571){\circle{90}}
}%
{\color[rgb]{0,0,0}\put(4816,-3571){\circle{180}}
}%
{\color[rgb]{0,0,0}\put(4501,-3616){\line(-4,-3){360}}
}%
{\color[rgb]{0,0,0}\put(4501,-3616){\line(-3,-4){270}}
}%
{\color[rgb]{0,0,0}\put(5041,-3616){\line(-3,-4){270}}
}%
{\color[rgb]{0,0,0}\put(5041,-3616){\line(-1,-3){135}}
}%
{\color[rgb]{0,0,0}\put(5041,-3616){\line( 0,-1){405}}
}%
{\color[rgb]{0,0,0}\put(4546,-3571){\line( 1, 0){180}}
}%
{\color[rgb]{0,0,0}\put(4501,-3616){\line( 0,-1){360}}
}%
{\color[rgb]{0,0,0}\put(5041,-3616){\line( 2,-1){360}}
}%
{\color[rgb]{0,0,0}\put(5041,-3616){\line( 1,-1){270}}
}%
\put(4501,-3481){\makebox(0,0)[b]{\smash{{\SetFigFont{12}{24.0}{\rmdefault}{\mddefault}{\updefault}{\color[rgb]{0,0,0}$1$}%
}}}}
\put(5041,-3481){\makebox(0,0)[b]{\smash{{\SetFigFont{12}{24.0}{\rmdefault}{\mddefault}{\updefault}{\color[rgb]{0,0,0}$m$}%
}}}}
\put(4816,-3436){\makebox(0,0)[b]{\smash{{\SetFigFont{12}{24.0}{\rmdefault}{\mddefault}{\updefault}{\color[rgb]{0,0,0}$m$}%
}}}}
\put(4501,-4156){\makebox(0,0)[b]{\smash{{\SetFigFont{12}{24.0}{\rmdefault}{\mddefault}{\updefault}{\color[rgb]{0,0,0}$C$}%
}}}}
\put(5401,-3976){\makebox(0,0)[b]{\smash{{\SetFigFont{12}{24.0}{\rmdefault}{\mddefault}{\updefault}{\color[rgb]{0,0,0}$D$}%
}}}}
\put(4906,-4201){\makebox(0,0)[b]{\smash{{\SetFigFont{12}{24.0}{\rmdefault}{\mddefault}{\updefault}{\color[rgb]{0,0,0}$B$}%
}}}}
\put(4141,-4066){\makebox(0,0)[b]{\smash{{\SetFigFont{12}{24.0}{\rmdefault}{\mddefault}{\updefault}{\color[rgb]{0,0,0}$A$}%
}}}}
\end{picture}%
 %
 %
& ~ ~ $\Rightarrow$ ~ ~ &
\setlength{\unitlength}{5000sp}%
\begingroup\makeatletter\ifx\SetFigFont\undefined%
\gdef\SetFigFont#1#2#3#4#5{%
  \reset@font\fontsize{#1}{#2pt}%
  \fontfamily{#3}\fontseries{#4}\fontshape{#5}%
  \selectfont}%
\fi\endgroup%
\begin{picture}(1290,885)(4126,-4216)
{\color[rgb]{0,0,0}\thinlines
\put(4501,-3571){\circle{90}}
}%
{\color[rgb]{0,0,0}\put(5041,-3571){\circle{90}}
}%
{\color[rgb]{0,0,0}\put(4501,-3616){\line(-4,-3){360}}
}%
{\color[rgb]{0,0,0}\put(4501,-3616){\line(-3,-4){270}}
}%
{\color[rgb]{0,0,0}\put(5041,-3616){\line(-3,-4){270}}
}%
{\color[rgb]{0,0,0}\put(5041,-3616){\line(-1,-3){135}}
}%
{\color[rgb]{0,0,0}\put(5041,-3616){\line( 0,-1){405}}
}%
{\color[rgb]{0,0,0}\put(4546,-3571){\line( 1, 0){450}}
}%
{\color[rgb]{0,0,0}\put(4501,-3616){\line( 0,-1){360}}
}%
{\color[rgb]{0,0,0}\put(5041,-3616){\line( 2,-1){360}}
}%
{\color[rgb]{0,0,0}\put(5041,-3616){\line( 1,-1){270}}
}%
\put(4501,-3481){\makebox(0,0)[b]{\smash{{\SetFigFont{12}{24.0}{\rmdefault}{\mddefault}{\updefault}{\color[rgb]{0,0,0}$1$}%
}}}}
\put(5041,-3481){\makebox(0,0)[b]{\smash{{\SetFigFont{12}{24.0}{\rmdefault}{\mddefault}{\updefault}{\color[rgb]{0,0,0}$m$}%
}}}}
\put(4501,-4156){\makebox(0,0)[b]{\smash{{\SetFigFont{12}{24.0}{\rmdefault}{\mddefault}{\updefault}{\color[rgb]{0,0,0}$C$}%
}}}}
\put(5401,-3976){\makebox(0,0)[b]{\smash{{\SetFigFont{12}{24.0}{\rmdefault}{\mddefault}{\updefault}{\color[rgb]{0,0,0}$D$}%
}}}}
\put(4906,-4201){\makebox(0,0)[b]{\smash{{\SetFigFont{12}{24.0}{\rmdefault}{\mddefault}{\updefault}{\color[rgb]{0,0,0}$B$}%
}}}}
\put(4141,-4066){\makebox(0,0)[b]{\smash{{\SetFigFont{12}{24.0}{\rmdefault}{\mddefault}{\updefault}{\color[rgb]{0,0,0}$A$}%
}}}}
\end{picture}%
 %
 %
\end{tabular}
}{
(1)(3)       (2)(4)             (1)(3)  (2)(4)
   O----(O)    O       ===>        O------O
   1     m     m                   1      m
}

The same can be done if $n=m=1$, or $m=1$ and $n>1$. This proves that,
regardless of the two values of the nodes of the chain, obtainability is the
same if the chain is folded in a zigzag manner. In other words:

\begin{enumerate}

\item for every se graph, every obtainable totally assigned se graph extending
it has alternating values for the nodes of the chain;

\item no matter what these values are, obtainability is not altered by folding
the chain.

\end{enumerate}

Therefore, folding turns an obtainable graph into an obtainable graph. If the
original graph is instead unobtainable, still has extensions to totally
assigned se graph with alternating values for the chain; however, these
extensions incorrectly select or exclude some edge. This condition is not
changed by the folding, either.~\qed

This lemma proves that every chain of selected edges can be turned into a
single edge. The same can be done iteratively until the graph is left with no
such a chain, so that no selected edge touches another one. Excluded edges may
still form chains of arbitrary length, though.

\subsubsection{Forced values}

Some graphs requires values to obey some simple conditions for obtaining the
expected result.

\begin{lemma}
\label{triangle}

In any obtainable total assigned se graph containing a triangle of selected
edges, the nodes of the triangle have value one.

\end{lemma}

\proof A triangle of selected edges is also a chain:

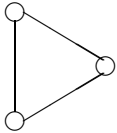
\setlength{\unitlength}{5000sp}%
\begingroup\makeatletter\ifx\SetFigFont\undefined%
\gdef\SetFigFont#1#2#3#4#5{%
  \reset@font\fontsize{#1}{#2pt}%
  \fontfamily{#3}\fontseries{#4}\fontshape{#5}%
  \selectfont}%
\fi\endgroup%
\begin{picture}(556,646)(5078,-3714)
{\color[rgb]{0,0,0}\thinlines
\put(5131,-3121){\circle{90}}
}%
{\color[rgb]{0,0,0}\put(5131,-3661){\circle{90}}
}%
{\color[rgb]{0,0,0}\put(5581,-3391){\circle{90}}
}%
{\color[rgb]{0,0,0}\put(5131,-3166){\line( 0,-1){450}}
}%
{\color[rgb]{0,0,0}\put(5176,-3121){\line( 5,-3){397.059}}
}%
{\color[rgb]{0,0,0}\put(5176,-3661){\line( 5, 3){397.059}}
}%
\end{picture}%
 %
\nop
{
O---\                   .
|    O
O---/
}

Let $n$, $m$, and $k$ be the values of these nodes. By Lemma~\ref{alternating},
$n=k$. But also $m=n$, as the sequence is $n-m-k-n$. Since either $n$ or $m$ is
equal to one by Lemma~\ref{first}, it follows $n=m=1$ and also $k=n=1$.~\qed

The following lemma shows that values are forced to increase in a chain of
edges that are alternatively excluded and selected. In this configuration, if
the first node is assigned $1$ the values are $1-n-1-m-k-\cdots$ with
$1<n<m<k<\ldots$. At a minimum, these values are $2$, $3$, $4$, etc.

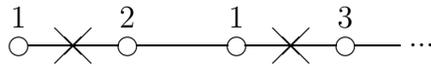
\setlength{\unitlength}{5000sp}%
\begingroup\makeatletter\ifx\SetFigFont\undefined%
\gdef\SetFigFont#1#2#3#4#5{%
  \reset@font\fontsize{#1}{#2pt}%
  \fontfamily{#3}\fontseries{#4}\fontshape{#5}%
  \selectfont}%
\fi\endgroup%
\begin{picture}(2003,342)(4448,-3403)
{\color[rgb]{0,0,0}\thinlines
\put(4501,-3301){\circle{90}}
}%
{\color[rgb]{0,0,0}\put(5041,-3301){\circle{90}}
}%
{\color[rgb]{0,0,0}\put(5581,-3301){\circle{90}}
}%
{\color[rgb]{0,0,0}\put(6121,-3301){\circle{90}}
}%
{\color[rgb]{0,0,0}\put(4546,-3301){\line( 1, 0){450}}
}%
{\color[rgb]{0,0,0}\put(5086,-3301){\line( 1, 0){450}}
}%
{\color[rgb]{0,0,0}\put(5626,-3301){\line( 1, 0){450}}
}%
{\color[rgb]{0,0,0}\put(6166,-3301){\line( 1, 0){225}}
}%
{\color[rgb]{0,0,0}\put(4681,-3211){\line( 1,-1){180}}
}%
{\color[rgb]{0,0,0}\put(4681,-3391){\line( 1, 1){180}}
}%
{\color[rgb]{0,0,0}\put(5761,-3211){\line( 1,-1){180}}
}%
{\color[rgb]{0,0,0}\put(5761,-3391){\line( 1, 1){180}}
}%
\put(6436,-3301){\makebox(0,0)[lb]{\smash{{\SetFigFont{12}{24.0}{\rmdefault}{\mddefault}{\updefault}{\color[rgb]{0,0,0}$...$}%
}}}}
\put(4501,-3211){\makebox(0,0)[b]{\smash{{\SetFigFont{12}{24.0}{\rmdefault}{\mddefault}{\updefault}{\color[rgb]{0,0,0}$1$}%
}}}}
\put(5041,-3211){\makebox(0,0)[b]{\smash{{\SetFigFont{12}{24.0}{\rmdefault}{\mddefault}{\updefault}{\color[rgb]{0,0,0}$2$}%
}}}}
\put(5581,-3211){\makebox(0,0)[b]{\smash{{\SetFigFont{12}{24.0}{\rmdefault}{\mddefault}{\updefault}{\color[rgb]{0,0,0}$1$}%
}}}}
\put(6121,-3211){\makebox(0,0)[b]{\smash{{\SetFigFont{12}{24.0}{\rmdefault}{\mddefault}{\updefault}{\color[rgb]{0,0,0}$3$}%
}}}}
\end{picture}%
 %
\nop
{
1     2     1     3    ...
O--X--O-----O--X--O----
}

\begin{lemma}
\label{increasing}

In any obtainable total assigned se graph containing a chain of alternating
excluded-selected edges with the first node assigned one, the values of the
other even nodes are one and of the even nodes are strictly increasing.

\end{lemma}

\proof The chain begins with value $1$ and an excluded edge:

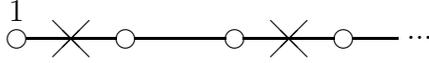
\setlength{\unitlength}{5000sp}%
\begingroup\makeatletter\ifx\SetFigFont\undefined%
\gdef\SetFigFont#1#2#3#4#5{%
  \reset@font\fontsize{#1}{#2pt}%
  \fontfamily{#3}\fontseries{#4}\fontshape{#5}%
  \selectfont}%
\fi\endgroup%
\begin{picture}(2003,342)(4448,-3403)
{\color[rgb]{0,0,0}\thinlines
\put(4501,-3301){\circle{90}}
}%
{\color[rgb]{0,0,0}\put(5041,-3301){\circle{90}}
}%
{\color[rgb]{0,0,0}\put(5581,-3301){\circle{90}}
}%
{\color[rgb]{0,0,0}\put(6121,-3301){\circle{90}}
}%
{\color[rgb]{0,0,0}\put(4546,-3301){\line( 1, 0){450}}
}%
{\color[rgb]{0,0,0}\put(5086,-3301){\line( 1, 0){450}}
}%
{\color[rgb]{0,0,0}\put(5626,-3301){\line( 1, 0){450}}
}%
{\color[rgb]{0,0,0}\put(6166,-3301){\line( 1, 0){225}}
}%
{\color[rgb]{0,0,0}\put(4681,-3211){\line( 1,-1){180}}
}%
{\color[rgb]{0,0,0}\put(4681,-3391){\line( 1, 1){180}}
}%
{\color[rgb]{0,0,0}\put(5761,-3211){\line( 1,-1){180}}
}%
{\color[rgb]{0,0,0}\put(5761,-3391){\line( 1, 1){180}}
}%
\put(6436,-3301){\makebox(0,0)[lb]{\smash{{\SetFigFont{12}{24.0}{\rmdefault}{\mddefault}{\updefault}{\color[rgb]{0,0,0}$...$}%
}}}}
\put(4501,-3211){\makebox(0,0)[b]{\smash{{\SetFigFont{12}{24.0}{\rmdefault}{\mddefault}{\updefault}{\color[rgb]{0,0,0}$1$}%
}}}}
\end{picture}%
 %
\nop
{
1
O--X--O-----O--X--O----
}

The next node cannot be one, as otherwise the edge would have values $1$ and
$1$, so it would be minimal. Let $n>1$ be the value of this node:

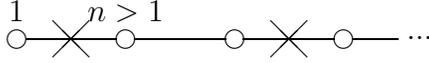
\setlength{\unitlength}{5000sp}%
\begingroup\makeatletter\ifx\SetFigFont\undefined%
\gdef\SetFigFont#1#2#3#4#5{%
  \reset@font\fontsize{#1}{#2pt}%
  \fontfamily{#3}\fontseries{#4}\fontshape{#5}%
  \selectfont}%
\fi\endgroup%
\begin{picture}(2003,342)(4448,-3403)
{\color[rgb]{0,0,0}\thinlines
\put(4501,-3301){\circle{90}}
}%
{\color[rgb]{0,0,0}\put(5041,-3301){\circle{90}}
}%
{\color[rgb]{0,0,0}\put(5581,-3301){\circle{90}}
}%
{\color[rgb]{0,0,0}\put(6121,-3301){\circle{90}}
}%
{\color[rgb]{0,0,0}\put(4546,-3301){\line( 1, 0){450}}
}%
{\color[rgb]{0,0,0}\put(5086,-3301){\line( 1, 0){450}}
}%
{\color[rgb]{0,0,0}\put(5626,-3301){\line( 1, 0){450}}
}%
{\color[rgb]{0,0,0}\put(6166,-3301){\line( 1, 0){225}}
}%
{\color[rgb]{0,0,0}\put(4681,-3211){\line( 1,-1){180}}
}%
{\color[rgb]{0,0,0}\put(4681,-3391){\line( 1, 1){180}}
}%
{\color[rgb]{0,0,0}\put(5761,-3211){\line( 1,-1){180}}
}%
{\color[rgb]{0,0,0}\put(5761,-3391){\line( 1, 1){180}}
}%
\put(6436,-3301){\makebox(0,0)[lb]{\smash{{\SetFigFont{12}{24.0}{\rmdefault}{\mddefault}{\updefault}{\color[rgb]{0,0,0}$...$}%
}}}}
\put(4501,-3211){\makebox(0,0)[b]{\smash{{\SetFigFont{12}{24.0}{\rmdefault}{\mddefault}{\updefault}{\color[rgb]{0,0,0}$1$}%
}}}}
\put(5041,-3211){\makebox(0,0)[b]{\smash{{\SetFigFont{12}{24.0}{\rmdefault}{\mddefault}{\updefault}{\color[rgb]{0,0,0}$n>1$}%
}}}}
\end{picture}%
 %
\nop
{
1    n>1
O--X--O-----O--X--O----
}

The second edge is selected: by Lemma~\ref{first}, it has at least a node
assigned one. Since $n>1$, this cannot be other than the third node:

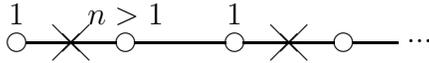
\setlength{\unitlength}{5000sp}%
\begingroup\makeatletter\ifx\SetFigFont\undefined%
\gdef\SetFigFont#1#2#3#4#5{%
  \reset@font\fontsize{#1}{#2pt}%
  \fontfamily{#3}\fontseries{#4}\fontshape{#5}%
  \selectfont}%
\fi\endgroup%
\begin{picture}(2003,342)(4448,-3403)
{\color[rgb]{0,0,0}\thinlines
\put(4501,-3301){\circle{90}}
}%
{\color[rgb]{0,0,0}\put(5041,-3301){\circle{90}}
}%
{\color[rgb]{0,0,0}\put(5581,-3301){\circle{90}}
}%
{\color[rgb]{0,0,0}\put(6121,-3301){\circle{90}}
}%
{\color[rgb]{0,0,0}\put(4546,-3301){\line( 1, 0){450}}
}%
{\color[rgb]{0,0,0}\put(5086,-3301){\line( 1, 0){450}}
}%
{\color[rgb]{0,0,0}\put(5626,-3301){\line( 1, 0){450}}
}%
{\color[rgb]{0,0,0}\put(6166,-3301){\line( 1, 0){225}}
}%
{\color[rgb]{0,0,0}\put(4681,-3211){\line( 1,-1){180}}
}%
{\color[rgb]{0,0,0}\put(4681,-3391){\line( 1, 1){180}}
}%
{\color[rgb]{0,0,0}\put(5761,-3211){\line( 1,-1){180}}
}%
{\color[rgb]{0,0,0}\put(5761,-3391){\line( 1, 1){180}}
}%
\put(6436,-3301){\makebox(0,0)[lb]{\smash{{\SetFigFont{12}{24.0}{\rmdefault}{\mddefault}{\updefault}{\color[rgb]{0,0,0}$...$}%
}}}}
\put(4501,-3211){\makebox(0,0)[b]{\smash{{\SetFigFont{12}{24.0}{\rmdefault}{\mddefault}{\updefault}{\color[rgb]{0,0,0}$1$}%
}}}}
\put(5581,-3211){\makebox(0,0)[b]{\smash{{\SetFigFont{12}{24.0}{\rmdefault}{\mddefault}{\updefault}{\color[rgb]{0,0,0}$1$}%
}}}}
\put(5041,-3211){\makebox(0,0)[b]{\smash{{\SetFigFont{12}{24.0}{\rmdefault}{\mddefault}{\updefault}{\color[rgb]{0,0,0}$n>1$}%
}}}}
\end{picture}%
 %
\nop
{
1    n>1    1
O--X--O-----O--X--O----
}

The values of the second edge are $n>1$ and $1$. The third edge also has the
node assigned $1$. In order to be non-minimal, the other value has to be
greater than $n$:

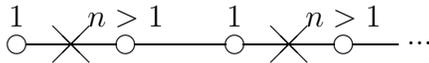
\setlength{\unitlength}{5000sp}%
\begingroup\makeatletter\ifx\SetFigFont\undefined%
\gdef\SetFigFont#1#2#3#4#5{%
  \reset@font\fontsize{#1}{#2pt}%
  \fontfamily{#3}\fontseries{#4}\fontshape{#5}%
  \selectfont}%
\fi\endgroup%
\begin{picture}(2003,342)(4448,-3403)
{\color[rgb]{0,0,0}\thinlines
\put(4501,-3301){\circle{90}}
}%
{\color[rgb]{0,0,0}\put(5041,-3301){\circle{90}}
}%
{\color[rgb]{0,0,0}\put(5581,-3301){\circle{90}}
}%
{\color[rgb]{0,0,0}\put(6121,-3301){\circle{90}}
}%
{\color[rgb]{0,0,0}\put(4546,-3301){\line( 1, 0){450}}
}%
{\color[rgb]{0,0,0}\put(5086,-3301){\line( 1, 0){450}}
}%
{\color[rgb]{0,0,0}\put(5626,-3301){\line( 1, 0){450}}
}%
{\color[rgb]{0,0,0}\put(6166,-3301){\line( 1, 0){225}}
}%
{\color[rgb]{0,0,0}\put(4681,-3211){\line( 1,-1){180}}
}%
{\color[rgb]{0,0,0}\put(4681,-3391){\line( 1, 1){180}}
}%
{\color[rgb]{0,0,0}\put(5761,-3211){\line( 1,-1){180}}
}%
{\color[rgb]{0,0,0}\put(5761,-3391){\line( 1, 1){180}}
}%
\put(6436,-3301){\makebox(0,0)[lb]{\smash{{\SetFigFont{12}{24.0}{\rmdefault}{\mddefault}{\updefault}{\color[rgb]{0,0,0}$...$}%
}}}}
\put(4501,-3211){\makebox(0,0)[b]{\smash{{\SetFigFont{12}{24.0}{\rmdefault}{\mddefault}{\updefault}{\color[rgb]{0,0,0}$1$}%
}}}}
\put(5581,-3211){\makebox(0,0)[b]{\smash{{\SetFigFont{12}{24.0}{\rmdefault}{\mddefault}{\updefault}{\color[rgb]{0,0,0}$1$}%
}}}}
\put(5041,-3211){\makebox(0,0)[b]{\smash{{\SetFigFont{12}{24.0}{\rmdefault}{\mddefault}{\updefault}{\color[rgb]{0,0,0}$n>1$}%
}}}}
\put(6121,-3211){\makebox(0,0)[b]{\smash{{\SetFigFont{12}{24.0}{\rmdefault}{\mddefault}{\updefault}{\color[rgb]{0,0,0}$n>1$}%
}}}}
\end{picture}%
 %
\nop
{
1    n>1    1    m>n
O--X--O-----O--X--O----
}

The proof can be iterated indefinitely, showing that each node of odd position
has value one, and each node of even position has a value that is greater than
the node two positions on the left of it.~\qed

\subsubsection{Graphs requiring $n$ values to be obtainable}

Several results are affected by whether values are equal to one or greater.
This may suggest that what really matters about a value is whether it is one or
not. In some cases, for example, a priority ordering the produces the expected
result can be obtained by placing a formula for each maxset in class one, and
all remaining ones in class two. This is however not always the case, as the
next lemma shows: some graphs can be obtained only with $n$ priority classes.

\begin{lemma}
\label{levels}

For every $n$ there exists a graph that is only obtained by assignments with at
least $n$ different values.

\end{lemma}

\proof The graph is as follows, where the chain is $2n$ long:

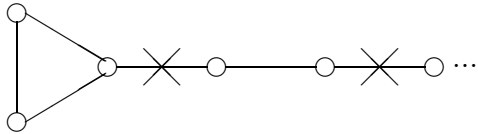
\setlength{\unitlength}{5000sp}%
\begingroup\makeatletter\ifx\SetFigFont\undefined%
\gdef\SetFigFont#1#2#3#4#5{%
  \reset@font\fontsize{#1}{#2pt}%
  \fontfamily{#3}\fontseries{#4}\fontshape{#5}%
  \selectfont}%
\fi\endgroup%
\begin{picture}(2228,646)(5078,-3714)
{\color[rgb]{0,0,0}\thinlines
\put(5131,-3121){\circle{90}}
}%
{\color[rgb]{0,0,0}\put(5131,-3661){\circle{90}}
}%
{\color[rgb]{0,0,0}\put(5581,-3391){\circle{90}}
}%
{\color[rgb]{0,0,0}\put(6121,-3391){\circle{90}}
}%
{\color[rgb]{0,0,0}\put(6661,-3391){\circle{90}}
}%
{\color[rgb]{0,0,0}\put(7201,-3391){\circle{90}}
}%
{\color[rgb]{0,0,0}\put(5131,-3166){\line( 0,-1){450}}
}%
{\color[rgb]{0,0,0}\put(5176,-3121){\line( 5,-3){397.059}}
}%
{\color[rgb]{0,0,0}\put(5176,-3661){\line( 5, 3){397.059}}
}%
{\color[rgb]{0,0,0}\put(6166,-3391){\line( 1, 0){450}}
}%
{\color[rgb]{0,0,0}\put(5626,-3391){\line( 1, 0){450}}
}%
{\color[rgb]{0,0,0}\put(6706,-3391){\line( 1, 0){450}}
}%
{\color[rgb]{0,0,0}\put(5761,-3301){\line( 1,-1){180}}
}%
{\color[rgb]{0,0,0}\put(5761,-3481){\line( 1, 1){180}}
}%
{\color[rgb]{0,0,0}\put(6841,-3301){\line( 1,-1){180}}
}%
{\color[rgb]{0,0,0}\put(6841,-3481){\line( 1, 1){180}}
}%
\put(7291,-3391){\makebox(0,0)[lb]{\smash{{\SetFigFont{12}{24.0}{\rmdefault}{\mddefault}{\updefault}{\color[rgb]{0,0,0}$...$}%
}}}}
\end{picture}%
 %
\nop
{
O---\                                                .
|    O--X--O-----O--X---O-----O ...
O---/
}

By Lemma~\ref{triangle}, the nodes of the triangle have value one in all
totally assigned se graph extending this one. This also holds for the starting
node of the chain, making Lemma~\ref{alternating} applicable. The values of the
chain are therefore $1$, $n>1$, $1$, $m>n$, $1$, $k>m$, \ldots\  Since the
chain is $2n$ long, it contains $n$ strictly increasing values.~\qed

When this lemma on graphs is recast in terms of formulae, it shows a sort of
counterexample to the converse of Property~\ref{first}: a priority ordering
cannot always obtained by choosing one formula for each maxset to place in
class one. To the contrary, some results can be obtained only with a large
number of classes.

\begin{corollary}

For any $n$, there exists $R$ and $K_1,\ldots,K_m$ such that $R$ is obtainable
by priority base merging from $K_1,\ldots,K_m$ only with priority partitions
having $n$ classes or more.

\end{corollary}

\subsubsection{Unobtainable graphs}

\begin{lemma}
\label{cycle}

A graph containing a cycle of alternating (single excluded edge)--(chain of odd
selected edges) is unobtainable.

\end{lemma}

\proof By Lemma~\ref{zigzag}, chains of odd selected edges can be folded into a
single edge where the first and last nodes are the same. After this
transformation, the cycle becomes a sequence of alternating excluded and
selected edges. An arbitrary selected edge can be taken as the starting point:

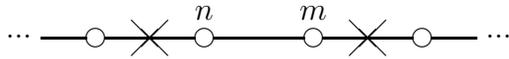
\setlength{\unitlength}{5000sp}%
\begingroup\makeatletter\ifx\SetFigFont\undefined%
\gdef\SetFigFont#1#2#3#4#5{%
  \reset@font\fontsize{#1}{#2pt}%
  \fontfamily{#3}\fontseries{#4}\fontshape{#5}%
  \selectfont}%
\fi\endgroup%
\begin{picture}(2280,297)(4261,-3493)
{\color[rgb]{0,0,0}\thinlines
\put(5131,-3391){\circle{90}}
}%
{\color[rgb]{0,0,0}\put(5671,-3391){\circle{90}}
}%
{\color[rgb]{0,0,0}\put(6211,-3391){\circle{90}}
}%
{\color[rgb]{0,0,0}\put(4591,-3391){\circle{90}}
}%
{\color[rgb]{0,0,0}\put(5176,-3391){\line( 1, 0){450}}
}%
{\color[rgb]{0,0,0}\put(5086,-3391){\line(-1, 0){450}}
}%
{\color[rgb]{0,0,0}\put(5716,-3391){\line( 1, 0){450}}
}%
{\color[rgb]{0,0,0}\put(6256,-3391){\line( 1, 0){225}}
}%
{\color[rgb]{0,0,0}\put(4546,-3391){\line(-1, 0){225}}
}%
{\color[rgb]{0,0,0}\put(4771,-3301){\line( 1,-1){180}}
}%
{\color[rgb]{0,0,0}\put(4771,-3481){\line( 1, 1){180}}
}%
{\color[rgb]{0,0,0}\put(5851,-3301){\line( 1,-1){180}}
}%
{\color[rgb]{0,0,0}\put(5851,-3481){\line( 1, 1){180}}
}%
\put(6526,-3391){\makebox(0,0)[lb]{\smash{{\SetFigFont{12}{24.0}{\rmdefault}{\mddefault}{\updefault}{\color[rgb]{0,0,0}$...$}%
}}}}
\put(4276,-3391){\makebox(0,0)[rb]{\smash{{\SetFigFont{12}{24.0}{\rmdefault}{\mddefault}{\updefault}{\color[rgb]{0,0,0}$...$}%
}}}}
\put(5131,-3301){\makebox(0,0)[b]{\smash{{\SetFigFont{12}{24.0}{\rmdefault}{\mddefault}{\updefault}{\color[rgb]{0,0,0}$n$}%
}}}}
\put(5671,-3301){\makebox(0,0)[b]{\smash{{\SetFigFont{12}{24.0}{\rmdefault}{\mddefault}{\updefault}{\color[rgb]{0,0,0}$m$}%
}}}}
\end{picture}%
 %
\nop
{
... --O--X--O-----O--X--O-- ...
            n     m
}

Lemma~\ref{first} tells that one among $n$ and $m$ is equal to one for the edge
to be selected. It can be assumed $m=1$, the other case is symmetric proceeding
right-to-left.

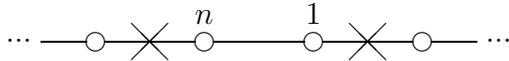
\setlength{\unitlength}{5000sp}%
\begingroup\makeatletter\ifx\SetFigFont\undefined%
\gdef\SetFigFont#1#2#3#4#5{%
  \reset@font\fontsize{#1}{#2pt}%
  \fontfamily{#3}\fontseries{#4}\fontshape{#5}%
  \selectfont}%
\fi\endgroup%
\begin{picture}(2280,342)(4261,-3493)
{\color[rgb]{0,0,0}\thinlines
\put(5131,-3391){\circle{90}}
}%
{\color[rgb]{0,0,0}\put(5671,-3391){\circle{90}}
}%
{\color[rgb]{0,0,0}\put(6211,-3391){\circle{90}}
}%
{\color[rgb]{0,0,0}\put(4591,-3391){\circle{90}}
}%
{\color[rgb]{0,0,0}\put(5176,-3391){\line( 1, 0){450}}
}%
{\color[rgb]{0,0,0}\put(5086,-3391){\line(-1, 0){450}}
}%
{\color[rgb]{0,0,0}\put(5716,-3391){\line( 1, 0){450}}
}%
{\color[rgb]{0,0,0}\put(6256,-3391){\line( 1, 0){225}}
}%
{\color[rgb]{0,0,0}\put(4546,-3391){\line(-1, 0){225}}
}%
{\color[rgb]{0,0,0}\put(4771,-3301){\line( 1,-1){180}}
}%
{\color[rgb]{0,0,0}\put(4771,-3481){\line( 1, 1){180}}
}%
{\color[rgb]{0,0,0}\put(5851,-3301){\line( 1,-1){180}}
}%
{\color[rgb]{0,0,0}\put(5851,-3481){\line( 1, 1){180}}
}%
\put(6526,-3391){\makebox(0,0)[lb]{\smash{{\SetFigFont{12}{24.0}{\rmdefault}{\mddefault}{\updefault}{\color[rgb]{0,0,0}$...$}%
}}}}
\put(4276,-3391){\makebox(0,0)[rb]{\smash{{\SetFigFont{12}{24.0}{\rmdefault}{\mddefault}{\updefault}{\color[rgb]{0,0,0}$...$}%
}}}}
\put(5131,-3301){\makebox(0,0)[b]{\smash{{\SetFigFont{12}{24.0}{\rmdefault}{\mddefault}{\updefault}{\color[rgb]{0,0,0}$n$}%
}}}}
\put(5671,-3301){\makebox(0,0)[b]{\smash{{\SetFigFont{12}{24.0}{\rmdefault}{\mddefault}{\updefault}{\color[rgb]{0,0,0}$1$}%
}}}}
\end{picture}%
 %
\nop
{
... --O--X--O-----O--X--O-- ...
            n     1
}

By Lemma~\ref{alternating}, the next values are alternating between one and an
increasing value. As an example, choosing the least possible values:

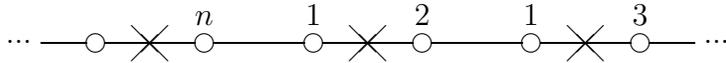
\setlength{\unitlength}{5000sp}%
\begingroup\makeatletter\ifx\SetFigFont\undefined%
\gdef\SetFigFont#1#2#3#4#5{%
  \reset@font\fontsize{#1}{#2pt}%
  \fontfamily{#3}\fontseries{#4}\fontshape{#5}%
  \selectfont}%
\fi\endgroup%
\begin{picture}(3360,342)(4261,-3493)
{\color[rgb]{0,0,0}\thinlines
\put(5131,-3391){\circle{90}}
}%
{\color[rgb]{0,0,0}\put(5671,-3391){\circle{90}}
}%
{\color[rgb]{0,0,0}\put(6211,-3391){\circle{90}}
}%
{\color[rgb]{0,0,0}\put(4591,-3391){\circle{90}}
}%
{\color[rgb]{0,0,0}\put(6751,-3391){\circle{90}}
}%
{\color[rgb]{0,0,0}\put(7291,-3391){\circle{90}}
}%
{\color[rgb]{0,0,0}\put(5176,-3391){\line( 1, 0){450}}
}%
{\color[rgb]{0,0,0}\put(5086,-3391){\line(-1, 0){450}}
}%
{\color[rgb]{0,0,0}\put(5716,-3391){\line( 1, 0){450}}
}%
{\color[rgb]{0,0,0}\put(4546,-3391){\line(-1, 0){225}}
}%
{\color[rgb]{0,0,0}\put(4771,-3301){\line( 1,-1){180}}
}%
{\color[rgb]{0,0,0}\put(4771,-3481){\line( 1, 1){180}}
}%
{\color[rgb]{0,0,0}\put(5851,-3301){\line( 1,-1){180}}
}%
{\color[rgb]{0,0,0}\put(5851,-3481){\line( 1, 1){180}}
}%
{\color[rgb]{0,0,0}\put(6256,-3391){\line( 1, 0){450}}
}%
{\color[rgb]{0,0,0}\put(6796,-3391){\line( 1, 0){450}}
}%
{\color[rgb]{0,0,0}\put(7336,-3391){\line( 1, 0){225}}
}%
{\color[rgb]{0,0,0}\put(6931,-3301){\line( 1,-1){180}}
}%
{\color[rgb]{0,0,0}\put(6931,-3481){\line( 1, 1){180}}
}%
\put(4276,-3391){\makebox(0,0)[rb]{\smash{{\SetFigFont{12}{24.0}{\rmdefault}{\mddefault}{\updefault}{\color[rgb]{0,0,0}$...$}%
}}}}
\put(5131,-3301){\makebox(0,0)[b]{\smash{{\SetFigFont{12}{24.0}{\rmdefault}{\mddefault}{\updefault}{\color[rgb]{0,0,0}$n$}%
}}}}
\put(5671,-3301){\makebox(0,0)[b]{\smash{{\SetFigFont{12}{24.0}{\rmdefault}{\mddefault}{\updefault}{\color[rgb]{0,0,0}$1$}%
}}}}
\put(7291,-3301){\makebox(0,0)[b]{\smash{{\SetFigFont{12}{24.0}{\rmdefault}{\mddefault}{\updefault}{\color[rgb]{0,0,0}$3$}%
}}}}
\put(6751,-3301){\makebox(0,0)[b]{\smash{{\SetFigFont{12}{24.0}{\rmdefault}{\mddefault}{\updefault}{\color[rgb]{0,0,0}$1$}%
}}}}
\put(6211,-3301){\makebox(0,0)[b]{\smash{{\SetFigFont{12}{24.0}{\rmdefault}{\mddefault}{\updefault}{\color[rgb]{0,0,0}$2$}%
}}}}
\put(7606,-3391){\makebox(0,0)[lb]{\smash{{\SetFigFont{12}{24.0}{\rmdefault}{\mddefault}{\updefault}{\color[rgb]{0,0,0}$...$}%
}}}}
\end{picture}%
 %
\nop
{
... --O--X--O-----O--X--O-----O--X--O--...
            n     1     2     1     3
}

The values at the end of excluded edges are increasing. Following the cycle,
$n$ gets its value, for example $10$:

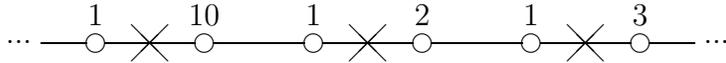
\setlength{\unitlength}{5000sp}%
\begingroup\makeatletter\ifx\SetFigFont\undefined%
\gdef\SetFigFont#1#2#3#4#5{%
  \reset@font\fontsize{#1}{#2pt}%
  \fontfamily{#3}\fontseries{#4}\fontshape{#5}%
  \selectfont}%
\fi\endgroup%
\begin{picture}(3360,342)(4261,-3493)
{\color[rgb]{0,0,0}\thinlines
\put(5131,-3391){\circle{90}}
}%
{\color[rgb]{0,0,0}\put(5671,-3391){\circle{90}}
}%
{\color[rgb]{0,0,0}\put(6211,-3391){\circle{90}}
}%
{\color[rgb]{0,0,0}\put(4591,-3391){\circle{90}}
}%
{\color[rgb]{0,0,0}\put(6751,-3391){\circle{90}}
}%
{\color[rgb]{0,0,0}\put(7291,-3391){\circle{90}}
}%
{\color[rgb]{0,0,0}\put(5176,-3391){\line( 1, 0){450}}
}%
{\color[rgb]{0,0,0}\put(5086,-3391){\line(-1, 0){450}}
}%
{\color[rgb]{0,0,0}\put(5716,-3391){\line( 1, 0){450}}
}%
{\color[rgb]{0,0,0}\put(4546,-3391){\line(-1, 0){225}}
}%
{\color[rgb]{0,0,0}\put(4771,-3301){\line( 1,-1){180}}
}%
{\color[rgb]{0,0,0}\put(4771,-3481){\line( 1, 1){180}}
}%
{\color[rgb]{0,0,0}\put(5851,-3301){\line( 1,-1){180}}
}%
{\color[rgb]{0,0,0}\put(5851,-3481){\line( 1, 1){180}}
}%
{\color[rgb]{0,0,0}\put(6256,-3391){\line( 1, 0){450}}
}%
{\color[rgb]{0,0,0}\put(6796,-3391){\line( 1, 0){450}}
}%
{\color[rgb]{0,0,0}\put(7336,-3391){\line( 1, 0){225}}
}%
{\color[rgb]{0,0,0}\put(6931,-3301){\line( 1,-1){180}}
}%
{\color[rgb]{0,0,0}\put(6931,-3481){\line( 1, 1){180}}
}%
\put(4276,-3391){\makebox(0,0)[rb]{\smash{{\SetFigFont{12}{24.0}{\rmdefault}{\mddefault}{\updefault}{\color[rgb]{0,0,0}$...$}%
}}}}
\put(5671,-3301){\makebox(0,0)[b]{\smash{{\SetFigFont{12}{24.0}{\rmdefault}{\mddefault}{\updefault}{\color[rgb]{0,0,0}$1$}%
}}}}
\put(7291,-3301){\makebox(0,0)[b]{\smash{{\SetFigFont{12}{24.0}{\rmdefault}{\mddefault}{\updefault}{\color[rgb]{0,0,0}$3$}%
}}}}
\put(6751,-3301){\makebox(0,0)[b]{\smash{{\SetFigFont{12}{24.0}{\rmdefault}{\mddefault}{\updefault}{\color[rgb]{0,0,0}$1$}%
}}}}
\put(6211,-3301){\makebox(0,0)[b]{\smash{{\SetFigFont{12}{24.0}{\rmdefault}{\mddefault}{\updefault}{\color[rgb]{0,0,0}$2$}%
}}}}
\put(7606,-3391){\makebox(0,0)[lb]{\smash{{\SetFigFont{12}{24.0}{\rmdefault}{\mddefault}{\updefault}{\color[rgb]{0,0,0}$...$}%
}}}}
\put(4591,-3301){\makebox(0,0)[b]{\smash{{\SetFigFont{12}{24.0}{\rmdefault}{\mddefault}{\updefault}{\color[rgb]{0,0,0}$1$}%
}}}}
\put(5131,-3301){\makebox(0,0)[b]{\smash{{\SetFigFont{12}{24.0}{\rmdefault}{\mddefault}{\updefault}{\color[rgb]{0,0,0}$10$}%
}}}}
\end{picture}%
 %
\nop
{
... --O--X--O-----O--X--O-----O--X--O--...
      1     10    1     2     1     3
}

The first edge has values $1$ and $10$, the next one has the same node of value
one and another of value $2$. Therefore, the second is minimal and the first is
not, opposite to the requirement.~\qed

The following lemma shows a necessary and sufficient condition to
obtainability.

\begin{lemma}
\label{empty}

A graph is obtainable if and only if the result of applying full disconnection,
removal of tails and zigzag folding as far as possible is an empty graph.

\end{lemma}

\proof These operations does not change obtainability. An empty graph is
obtainable, as it does not contain edges on which selection can be incorrect;
therefore, if the transformations lead to an empty graph, the original one is
obtainable.

In the other way around, if the resulting graph is not empty:

\begin{enumerate}

\item every node is touched by at least two edges, as otherwise the single
edge would have been deleted by removal of tails;

\item every node is touched by exactly one selected edge and one or more
excluded edges; otherwise, two selected edges would have been folded, and
excluded edges only separated by full disconnection.

\end{enumerate}

As a result of the second point, if the graph is not empty it contains at least
a selected edge. For the graph to be obtainable, either one or its two nodes
has to be assigned one by Lemma~\ref{first}. The other node may be one or a
greater value. 

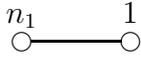
\setlength{\unitlength}{5000sp}%
\begingroup\makeatletter\ifx\SetFigFont\undefined%
\gdef\SetFigFont#1#2#3#4#5{%
  \reset@font\fontsize{#1}{#2pt}%
  \fontfamily{#3}\fontseries{#4}\fontshape{#5}%
  \selectfont}%
\fi\endgroup%
\begin{picture}(646,293)(4718,-3444)
{\color[rgb]{0,0,0}\thinlines
\put(4771,-3391){\circle{90}}
}%
{\color[rgb]{0,0,0}\put(5311,-3391){\circle{90}}
}%
{\color[rgb]{0,0,0}\put(4816,-3391){\line( 1, 0){450}}
}%
\put(4771,-3301){\makebox(0,0)[b]{\smash{{\SetFigFont{12}{24.0}{\rmdefault}{\mddefault}{\updefault}{\color[rgb]{0,0,0}$n_1$}%
}}}}
\put(5311,-3301){\makebox(0,0)[b]{\smash{{\SetFigFont{12}{24.0}{\rmdefault}{\mddefault}{\updefault}{\color[rgb]{0,0,0}$1$}%
}}}}
\end{picture}%
 %
\nop
{
    n1    1
    O-----O
}

The case in which the values are reversed is identical.

By the first property of this graph, the node of value $1$ is touched by at
last another edge, which is excluded because of the second property.

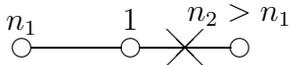
\setlength{\unitlength}{5000sp}%
\begingroup\makeatletter\ifx\SetFigFont\undefined%
\gdef\SetFigFont#1#2#3#4#5{%
  \reset@font\fontsize{#1}{#2pt}%
  \fontfamily{#3}\fontseries{#4}\fontshape{#5}%
  \selectfont}%
\fi\endgroup%
\begin{picture}(1186,387)(4718,-3493)
{\color[rgb]{0,0,0}\thinlines
\put(4771,-3391){\circle{90}}
}%
{\color[rgb]{0,0,0}\put(5311,-3391){\circle{90}}
}%
{\color[rgb]{0,0,0}\put(5851,-3391){\circle{90}}
}%
{\color[rgb]{0,0,0}\put(4816,-3391){\line( 1, 0){450}}
}%
{\color[rgb]{0,0,0}\put(5356,-3391){\line( 1, 0){450}}
}%
{\color[rgb]{0,0,0}\put(5491,-3301){\line( 1,-1){180}}
}%
{\color[rgb]{0,0,0}\put(5491,-3481){\line( 1, 1){180}}
}%
\put(4771,-3301){\makebox(0,0)[b]{\smash{{\SetFigFont{12}{24.0}{\rmdefault}{\mddefault}{\updefault}{\color[rgb]{0,0,0}$n_1$}%
}}}}
\put(5311,-3301){\makebox(0,0)[b]{\smash{{\SetFigFont{12}{24.0}{\rmdefault}{\mddefault}{\updefault}{\color[rgb]{0,0,0}$1$}%
}}}}
\put(5851,-3256){\makebox(0,0)[b]{\smash{{\SetFigFont{12}{24.0}{\rmdefault}{\mddefault}{\updefault}{\color[rgb]{0,0,0}$n_2>n_1$}%
}}}}
\end{picture}%
 %
\nop
{
    n1    1    n2>n1
    O-----O--X--O
}

By Lemma~\ref{increasing}, $n_2$ is greater than $n_1$, as otherwise the first
edge would not be selected and the second not excluded. By the two properties
of the graph, the node of value $n_2$ is connected to at least a selected edge:

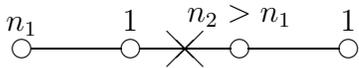
\setlength{\unitlength}{5000sp}%
\begingroup\makeatletter\ifx\SetFigFont\undefined%
\gdef\SetFigFont#1#2#3#4#5{%
  \reset@font\fontsize{#1}{#2pt}%
  \fontfamily{#3}\fontseries{#4}\fontshape{#5}%
  \selectfont}%
\fi\endgroup%
\begin{picture}(1726,387)(4718,-3493)
{\color[rgb]{0,0,0}\thinlines
\put(4771,-3391){\circle{90}}
}%
{\color[rgb]{0,0,0}\put(5311,-3391){\circle{90}}
}%
{\color[rgb]{0,0,0}\put(5851,-3391){\circle{90}}
}%
{\color[rgb]{0,0,0}\put(6391,-3391){\circle{90}}
}%
{\color[rgb]{0,0,0}\put(4816,-3391){\line( 1, 0){450}}
}%
{\color[rgb]{0,0,0}\put(5356,-3391){\line( 1, 0){450}}
}%
{\color[rgb]{0,0,0}\put(5491,-3301){\line( 1,-1){180}}
}%
{\color[rgb]{0,0,0}\put(5491,-3481){\line( 1, 1){180}}
}%
{\color[rgb]{0,0,0}\put(5896,-3391){\line( 1, 0){450}}
}%
\put(4771,-3301){\makebox(0,0)[b]{\smash{{\SetFigFont{12}{24.0}{\rmdefault}{\mddefault}{\updefault}{\color[rgb]{0,0,0}$n_1$}%
}}}}
\put(5311,-3301){\makebox(0,0)[b]{\smash{{\SetFigFont{12}{24.0}{\rmdefault}{\mddefault}{\updefault}{\color[rgb]{0,0,0}$1$}%
}}}}
\put(6391,-3301){\makebox(0,0)[b]{\smash{{\SetFigFont{12}{24.0}{\rmdefault}{\mddefault}{\updefault}{\color[rgb]{0,0,0}$1$}%
}}}}
\put(5851,-3256){\makebox(0,0)[b]{\smash{{\SetFigFont{12}{24.0}{\rmdefault}{\mddefault}{\updefault}{\color[rgb]{0,0,0}$n_2>n_1$}%
}}}}
\end{picture}%
 %
\nop
{
    n1    1     n2    1
    O-----O--X--O-----O
}

The last node is in turn connected to an excluded edge:

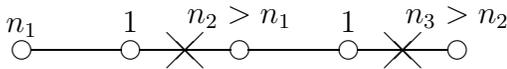
\setlength{\unitlength}{5000sp}%
\begingroup\makeatletter\ifx\SetFigFont\undefined%
\gdef\SetFigFont#1#2#3#4#5{%
  \reset@font\fontsize{#1}{#2pt}%
  \fontfamily{#3}\fontseries{#4}\fontshape{#5}%
  \selectfont}%
\fi\endgroup%
\begin{picture}(2266,387)(4718,-3493)
{\color[rgb]{0,0,0}\thinlines
\put(4771,-3391){\circle{90}}
}%
{\color[rgb]{0,0,0}\put(5311,-3391){\circle{90}}
}%
{\color[rgb]{0,0,0}\put(5851,-3391){\circle{90}}
}%
{\color[rgb]{0,0,0}\put(6391,-3391){\circle{90}}
}%
{\color[rgb]{0,0,0}\put(6931,-3391){\circle{90}}
}%
{\color[rgb]{0,0,0}\put(4816,-3391){\line( 1, 0){450}}
}%
{\color[rgb]{0,0,0}\put(5356,-3391){\line( 1, 0){450}}
}%
{\color[rgb]{0,0,0}\put(5491,-3301){\line( 1,-1){180}}
}%
{\color[rgb]{0,0,0}\put(5491,-3481){\line( 1, 1){180}}
}%
{\color[rgb]{0,0,0}\put(5896,-3391){\line( 1, 0){450}}
}%
{\color[rgb]{0,0,0}\put(6436,-3391){\line( 1, 0){450}}
}%
{\color[rgb]{0,0,0}\put(6571,-3301){\line( 1,-1){180}}
}%
{\color[rgb]{0,0,0}\put(6571,-3481){\line( 1, 1){180}}
}%
\put(4771,-3301){\makebox(0,0)[b]{\smash{{\SetFigFont{12}{24.0}{\rmdefault}{\mddefault}{\updefault}{\color[rgb]{0,0,0}$n_1$}%
}}}}
\put(5311,-3301){\makebox(0,0)[b]{\smash{{\SetFigFont{12}{24.0}{\rmdefault}{\mddefault}{\updefault}{\color[rgb]{0,0,0}$1$}%
}}}}
\put(6391,-3301){\makebox(0,0)[b]{\smash{{\SetFigFont{12}{24.0}{\rmdefault}{\mddefault}{\updefault}{\color[rgb]{0,0,0}$1$}%
}}}}
\put(5851,-3256){\makebox(0,0)[b]{\smash{{\SetFigFont{12}{24.0}{\rmdefault}{\mddefault}{\updefault}{\color[rgb]{0,0,0}$n_2>n_1$}%
}}}}
\put(6931,-3256){\makebox(0,0)[b]{\smash{{\SetFigFont{12}{24.0}{\rmdefault}{\mddefault}{\updefault}{\color[rgb]{0,0,0}$n_3>n_2$}%
}}}}
\end{picture}%
 %
\nop
{
    n1    1     n2    1     n3
    O-----O--X--O-----O--X--O
}

Again, $n_3>n_2$ by Lemma~\ref{increasing}. The sequence proceeds alternating
selected and excluded edges. By Lemma~\ref{increasing}, the nodes at the end of
a selected edge have value $1$, the others have increasing values. Since every
node is touched by at least two edges in this graph, the sequence can be
extended indefinitely, until it reaches a node that it already crossed.

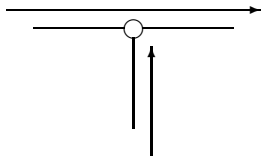
\setlength{\unitlength}{5000sp}%
\begingroup\makeatletter\ifx\SetFigFont\undefined%
\gdef\SetFigFont#1#2#3#4#5{%
  \reset@font\fontsize{#1}{#2pt}%
  \fontfamily{#3}\fontseries{#4}\fontshape{#5}%
  \selectfont}%
\fi\endgroup%
\begin{picture}(1284,774)(4849,-4033)
{\color[rgb]{0,0,0}\thinlines
\put(5491,-3391){\circle{90}}
}%
{\color[rgb]{0,0,0}\put(4996,-3391){\line( 1, 0){450}}
}%
{\color[rgb]{0,0,0}\put(5536,-3391){\line( 1, 0){450}}
}%
{\color[rgb]{0,0,0}\put(5491,-3436){\line( 0,-1){450}}
}%
{\color[rgb]{0,0,0}\put(5581,-4021){\vector( 0, 1){540}}
}%
{\color[rgb]{0,0,0}\put(4861,-3301){\vector( 1, 0){1260}}
}%
\end{picture}%
 %
\nop
{
------------------------------>
   ... -?--O--?- ...
           |
           ?  ^
           |  |
              |
}

Since the path is alternating, one of the two horizontal edges is selected and
the other is excluded, leading to two possible cases. Since no node is touched
by more than one selected edge, the one leading back to it is excluded:

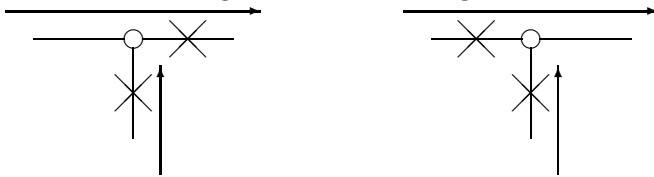
\ttytex{
\begin{tabular}{ccc}
\setlength{\unitlength}{5000sp}%
\begingroup\makeatletter\ifx\SetFigFont\undefined%
\gdef\SetFigFont#1#2#3#4#5{%
  \reset@font\fontsize{#1}{#2pt}%
  \fontfamily{#3}\fontseries{#4}\fontshape{#5}%
  \selectfont}%
\fi\endgroup%
\begin{picture}(1284,864)(4849,-4078)
{\color[rgb]{0,0,0}\thinlines
\put(5491,-3391){\circle{90}}
}%
{\color[rgb]{0,0,0}\put(4996,-3391){\line( 1, 0){450}}
}%
{\color[rgb]{0,0,0}\put(5536,-3391){\line( 1, 0){450}}
}%
{\color[rgb]{0,0,0}\put(5491,-3436){\line( 0,-1){450}}
}%
{\color[rgb]{0,0,0}\put(5671,-3301){\line( 1,-1){180}}
}%
{\color[rgb]{0,0,0}\put(5671,-3481){\line( 1, 1){180}}
}%
{\color[rgb]{0,0,0}\put(5401,-3571){\line( 1,-1){180}}
}%
{\color[rgb]{0,0,0}\put(5401,-3751){\line( 1, 1){180}}
}%
{\color[rgb]{0,0,0}\put(4861,-3256){\vector( 1, 0){1260}}
}%
{\color[rgb]{0,0,0}\put(5626,-4066){\vector( 0, 1){540}}
}%
\end{picture}%
 %
 %
& \hbox to 1cm{\hfill} &
\setlength{\unitlength}{5000sp}%
\begingroup\makeatletter\ifx\SetFigFont\undefined%
\gdef\SetFigFont#1#2#3#4#5{%
  \reset@font\fontsize{#1}{#2pt}%
  \fontfamily{#3}\fontseries{#4}\fontshape{#5}%
  \selectfont}%
\fi\endgroup%
\begin{picture}(1284,864)(4849,-4078)
{\color[rgb]{0,0,0}\thinlines
\put(5491,-3391){\circle{90}}
}%
{\color[rgb]{0,0,0}\put(4996,-3391){\line( 1, 0){450}}
}%
{\color[rgb]{0,0,0}\put(5536,-3391){\line( 1, 0){450}}
}%
{\color[rgb]{0,0,0}\put(5491,-3436){\line( 0,-1){450}}
}%
{\color[rgb]{0,0,0}\put(5131,-3301){\line( 1,-1){180}}
}%
{\color[rgb]{0,0,0}\put(5131,-3481){\line( 1, 1){180}}
}%
{\color[rgb]{0,0,0}\put(5401,-3571){\line( 1,-1){180}}
}%
{\color[rgb]{0,0,0}\put(5401,-3751){\line( 1, 1){180}}
}%
{\color[rgb]{0,0,0}\put(5626,-4066){\vector( 0, 1){540}}
}%
{\color[rgb]{0,0,0}\put(4861,-3256){\vector( 1, 0){1260}}
}%
\end{picture}%
 %
 %
\end{tabular}
}{
------------------------------>
   ... ----O--X- ...
           |
           ?  ^
           |  |
              |

------------------------------>
   ... -X--O---- ...
           |
           ?  ^
           |  |
              |
}

All values on the path obey the rules of Lemma~\ref{increasing}: one at the end
of a selected edge, increasing the others.

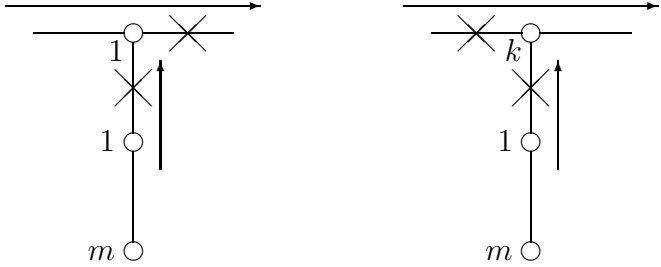
\ttytex{
\begin{tabular}{ccc}
\setlength{\unitlength}{5000sp}%
\begingroup\makeatletter\ifx\SetFigFont\undefined%
\gdef\SetFigFont#1#2#3#4#5{%
  \reset@font\fontsize{#1}{#2pt}%
  \fontfamily{#3}\fontseries{#4}\fontshape{#5}%
  \selectfont}%
\fi\endgroup%
\begin{picture}(1284,1317)(4849,-4531)
{\color[rgb]{0,0,0}\thinlines
\put(5491,-3391){\circle{90}}
}%
{\color[rgb]{0,0,0}\put(5491,-3931){\circle{90}}
}%
{\color[rgb]{0,0,0}\put(5491,-4471){\circle{90}}
}%
{\color[rgb]{0,0,0}\put(4996,-3391){\line( 1, 0){450}}
}%
{\color[rgb]{0,0,0}\put(5536,-3391){\line( 1, 0){450}}
}%
{\color[rgb]{0,0,0}\put(5491,-3436){\line( 0,-1){450}}
}%
{\color[rgb]{0,0,0}\put(5671,-3301){\line( 1,-1){180}}
}%
{\color[rgb]{0,0,0}\put(5671,-3481){\line( 1, 1){180}}
}%
{\color[rgb]{0,0,0}\put(5401,-3571){\line( 1,-1){180}}
}%
{\color[rgb]{0,0,0}\put(5401,-3751){\line( 1, 1){180}}
}%
{\color[rgb]{0,0,0}\put(5491,-3976){\line( 0,-1){450}}
}%
{\color[rgb]{0,0,0}\put(5626,-4066){\vector( 0, 1){540}}
}%
{\color[rgb]{0,0,0}\put(4861,-3256){\vector( 1, 0){1260}}
}%
\put(5401,-4516){\makebox(0,0)[rb]{\smash{{\SetFigFont{12}{24.0}{\rmdefault}{\mddefault}{\updefault}{\color[rgb]{0,0,0}$m$}%
}}}}
\put(5401,-3976){\makebox(0,0)[rb]{\smash{{\SetFigFont{12}{24.0}{\rmdefault}{\mddefault}{\updefault}{\color[rgb]{0,0,0}$1$}%
}}}}
\put(5446,-3526){\makebox(0,0)[rb]{\smash{{\SetFigFont{12}{24.0}{\rmdefault}{\mddefault}{\updefault}{\color[rgb]{0,0,0}$1$}%
}}}}
\end{picture}%
 %
 %
& \hbox to 1cm{\hfill} &
\setlength{\unitlength}{5000sp}%
\begingroup\makeatletter\ifx\SetFigFont\undefined%
\gdef\SetFigFont#1#2#3#4#5{%
  \reset@font\fontsize{#1}{#2pt}%
  \fontfamily{#3}\fontseries{#4}\fontshape{#5}%
  \selectfont}%
\fi\endgroup%
\begin{picture}(1284,1317)(4849,-4531)
{\color[rgb]{0,0,0}\thinlines
\put(5491,-3391){\circle{90}}
}%
{\color[rgb]{0,0,0}\put(5491,-3931){\circle{90}}
}%
{\color[rgb]{0,0,0}\put(5491,-4471){\circle{90}}
}%
{\color[rgb]{0,0,0}\put(4996,-3391){\line( 1, 0){450}}
}%
{\color[rgb]{0,0,0}\put(5536,-3391){\line( 1, 0){450}}
}%
{\color[rgb]{0,0,0}\put(5491,-3436){\line( 0,-1){450}}
}%
{\color[rgb]{0,0,0}\put(5131,-3301){\line( 1,-1){180}}
}%
{\color[rgb]{0,0,0}\put(5131,-3481){\line( 1, 1){180}}
}%
{\color[rgb]{0,0,0}\put(5401,-3571){\line( 1,-1){180}}
}%
{\color[rgb]{0,0,0}\put(5401,-3751){\line( 1, 1){180}}
}%
{\color[rgb]{0,0,0}\put(5491,-3976){\line( 0,-1){450}}
}%
{\color[rgb]{0,0,0}\put(4861,-3256){\vector( 1, 0){1260}}
}%
{\color[rgb]{0,0,0}\put(5626,-4066){\vector( 0, 1){540}}
}%
\put(5401,-3976){\makebox(0,0)[rb]{\smash{{\SetFigFont{12}{24.0}{\rmdefault}{\mddefault}{\updefault}{\color[rgb]{0,0,0}$1$}%
}}}}
\put(5401,-4516){\makebox(0,0)[rb]{\smash{{\SetFigFont{12}{24.0}{\rmdefault}{\mddefault}{\updefault}{\color[rgb]{0,0,0}$m$}%
}}}}
\put(5446,-3526){\makebox(0,0)[rb]{\smash{{\SetFigFont{12}{24.0}{\rmdefault}{\mddefault}{\updefault}{\color[rgb]{0,0,0}$k$}%
}}}}
\end{picture}%
 %
 %
\end{tabular}
}{
------------------------------>
           1
   ... ----O--X- ...
           |  
           X  ^
           |  |
         1 O  |
           |  |
           |  |
           |
         m O

------------------------------>
           k
   ... -X--O---- ...
           |
           X  ^
           |  |
         1 O  |
           |  |
           |  |
           |
         m O
}

In the first case, the vertical excluded edge is incorrectly selected. In the
second case, by Lemma~\ref{increasing} $m$ is greater than $k$ because it is
later in the sequence; as a result, the vertical selected edge is incorrectly
excluded.

This proves that assigning the first selected edge values $n_1$ and $1$ leads
to unobtainability. But the same happens, by symmetry, if these values are
reversed.~\qed

Lemma~\ref{cycle} shows that a graph is unobtainable if it contains an
alternating cycle. A proof similar to the one of the last lemma allows
reversing this result, if cycles are allowed to follow an edge twice in
opposite directions. An example where this is necessary is:

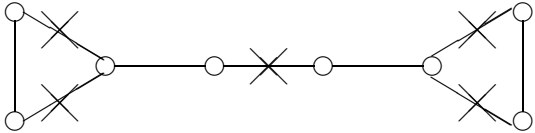
\setlength{\unitlength}{5000sp}%
\begingroup\makeatletter\ifx\SetFigFont\undefined%
\gdef\SetFigFont#1#2#3#4#5{%
  \reset@font\fontsize{#1}{#2pt}%
  \fontfamily{#3}\fontseries{#4}\fontshape{#5}%
  \selectfont}%
\fi\endgroup%
\begin{picture}(2626,646)(5078,-3714)
{\color[rgb]{0,0,0}\thinlines
\put(5131,-3121){\circle{90}}
}%
{\color[rgb]{0,0,0}\put(5131,-3661){\circle{90}}
}%
{\color[rgb]{0,0,0}\put(5581,-3391){\circle{90}}
}%
{\color[rgb]{0,0,0}\put(6121,-3391){\circle{90}}
}%
{\color[rgb]{0,0,0}\put(6661,-3391){\circle{90}}
}%
{\color[rgb]{0,0,0}\put(7201,-3391){\circle{90}}
}%
{\color[rgb]{0,0,0}\put(7651,-3121){\circle{90}}
}%
{\color[rgb]{0,0,0}\put(7651,-3661){\circle{90}}
}%
{\color[rgb]{0,0,0}\put(5131,-3166){\line( 0,-1){450}}
}%
{\color[rgb]{0,0,0}\put(5176,-3121){\line( 5,-3){397.059}}
}%
{\color[rgb]{0,0,0}\put(6166,-3391){\line( 1, 0){450}}
}%
{\color[rgb]{0,0,0}\put(5626,-3391){\line( 1, 0){450}}
}%
{\color[rgb]{0,0,0}\put(6706,-3391){\line( 1, 0){450}}
}%
{\color[rgb]{0,0,0}\put(5176,-3661){\line( 5, 3){397.059}}
}%
{\color[rgb]{0,0,0}\put(7199,-3342){\line( 5, 3){397.059}}
}%
{\color[rgb]{0,0,0}\put(7197,-3444){\line( 5,-3){397.059}}
}%
{\color[rgb]{0,0,0}\put(7651,-3166){\line( 0,-1){450}}
}%
{\color[rgb]{0,0,0}\put(5266,-3481){\line( 1,-1){180}}
}%
{\color[rgb]{0,0,0}\put(5266,-3661){\line( 1, 1){180}}
}%
{\color[rgb]{0,0,0}\put(5266,-3121){\line( 1,-1){180}}
}%
{\color[rgb]{0,0,0}\put(5266,-3301){\line( 1, 1){180}}
}%
{\color[rgb]{0,0,0}\put(6301,-3301){\line( 1,-1){180}}
}%
{\color[rgb]{0,0,0}\put(6301,-3481){\line( 1, 1){180}}
}%
{\color[rgb]{0,0,0}\put(7336,-3121){\line( 1,-1){180}}
}%
{\color[rgb]{0,0,0}\put(7336,-3301){\line( 1, 1){180}}
}%
{\color[rgb]{0,0,0}\put(7336,-3481){\line( 1,-1){180}}
}%
{\color[rgb]{0,0,0}\put(7336,-3661){\line( 1, 1){180}}
}%
\end{picture}%
 %
\nop
{
O-X                     X-O
|  \                   /  |
|   O-----O--X--O-----O   |
|  /                   \  |
O-X                     X-O
}

None of the three transformations can be applied, as the graph contains no
tails, no chain of selected edges, and no node connected to excluded edges
only. The graph is therefore unobtainable. However, the only alternating cycles
crosses the chain of three edges in the middle twice, once left-to-right and
once right-to-left.

\begin{lemma}

If a graph is unobtainable, it contains an alternating (single excluded
edge)--(chain of odd selected edges) cycle that contains the same edge at most
twice.

\end{lemma}

\proof The claim is proved in two parts: first, the transformations do not add
or remove alternating cycles; second, if the resulting graph is not empty, it
contains an alternating cycle. By Lemma~\ref{cycle}, if the graph is
unobtainable then the resulting graph is not empty; therefore, the original
graph also has an alternating chain.

\begin{itemize}

\item full disconnection does not open alternating cycles, as every node in
them is touched by a selected edge (no consecutive excluded edges); it does not
create a new one either, as it only disconnect edges;

\item tail removal only remove edges, so it never creates a new cycle; it does
not touch existing cycles, alternating or otherwise;

\item zigzag foldings do change cycles; however, it turns every path of odd
selected edges into another path of selected edges of length one, and one is
still an odd number; in the same way, paths of even edges are turned into paths
of zero length; as a result, a cycle exists after the change if and only if it
existed beforehand, and it is alternating if it was.

\end{itemize}

The second part of the proof shows that a non-empty graph resulting from
applying the three transformations contains an alternating cycle. In particular,
one alternating between single excluded edges and single selected edges. This
is shown with a proof similar to the one of the previous theorem, with a
difference. If the path reaches one of its previous nodes, this is not the end
of the cycle if this would lead to two consecutive excluded edges:

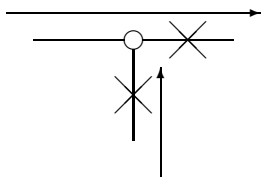
\setlength{\unitlength}{5000sp}%
\begingroup\makeatletter\ifx\SetFigFont\undefined%
\gdef\SetFigFont#1#2#3#4#5{%
  \reset@font\fontsize{#1}{#2pt}%
  \fontfamily{#3}\fontseries{#4}\fontshape{#5}%
  \selectfont}%
\fi\endgroup%
\begin{picture}(1284,864)(4849,-4078)
{\color[rgb]{0,0,0}\thinlines
\put(5491,-3391){\circle{90}}
}%
{\color[rgb]{0,0,0}\put(4996,-3391){\line( 1, 0){450}}
}%
{\color[rgb]{0,0,0}\put(5536,-3391){\line( 1, 0){450}}
}%
{\color[rgb]{0,0,0}\put(5491,-3436){\line( 0,-1){450}}
}%
{\color[rgb]{0,0,0}\put(5671,-3301){\line( 1,-1){180}}
}%
{\color[rgb]{0,0,0}\put(5671,-3481){\line( 1, 1){180}}
}%
{\color[rgb]{0,0,0}\put(5401,-3571){\line( 1,-1){180}}
}%
{\color[rgb]{0,0,0}\put(5401,-3751){\line( 1, 1){180}}
}%
{\color[rgb]{0,0,0}\put(4861,-3256){\vector( 1, 0){1260}}
}%
{\color[rgb]{0,0,0}\put(5626,-4066){\vector( 0, 1){540}}
}%
\end{picture}%
 %
\nop
{
------------------------------>
   ... ----O--X- ...
           |  
        ^  X
        |  |
        |  O
        |  |
        |  |
	|
}

If the edge after the node is selected, the cycle could be closed as an
alternating one. This being not the case, the path is continued on the left:

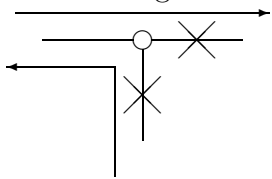
\setlength{\unitlength}{5000sp}%
\begingroup\makeatletter\ifx\SetFigFont\undefined%
\gdef\SetFigFont#1#2#3#4#5{%
  \reset@font\fontsize{#1}{#2pt}%
  \fontfamily{#3}\fontseries{#4}\fontshape{#5}%
  \selectfont}%
\fi\endgroup%
\begin{picture}(1329,864)(4804,-4078)
{\color[rgb]{0,0,0}\thinlines
\put(5491,-3391){\circle{90}}
}%
{\color[rgb]{0,0,0}\put(4996,-3391){\line( 1, 0){450}}
}%
{\color[rgb]{0,0,0}\put(5536,-3391){\line( 1, 0){450}}
}%
{\color[rgb]{0,0,0}\put(5491,-3436){\line( 0,-1){450}}
}%
{\color[rgb]{0,0,0}\put(5671,-3301){\line( 1,-1){180}}
}%
{\color[rgb]{0,0,0}\put(5671,-3481){\line( 1, 1){180}}
}%
{\color[rgb]{0,0,0}\put(5401,-3571){\line( 1,-1){180}}
}%
{\color[rgb]{0,0,0}\put(5401,-3751){\line( 1, 1){180}}
}%
{\color[rgb]{0,0,0}\put(4861,-3256){\vector( 1, 0){1260}}
}%
{\color[rgb]{0,0,0}\put(5356,-4066){\line( 0, 1){540}}
\put(5356,-3526){\vector(-1, 0){540}}
}%
\end{picture}%
 %
\nop
{
------------------------------>
   ... ----O--X- ...
   <----\  |  
        |  X
        |  |
        |  O
        |  |
        |  |
        |
}

The sequence can continue indefinitely. Since there are only a finite number of
edges and only two directions for each edge, at some point the sequence comes
back to an edge in the same direction it followed it before. The cycle is
closed at that point.~\qed

Since the existence of an alternating cycle implies unobtainability but is also
implied by it, it is a characterization of this property.

\begin{corollary}
\label{characterization}

A graph is unobtainable if and only if it contains a cycle of alternating
(single excluded edge)-(chain of odd selected edges) that crosses the same edge
at most twice.

\end{corollary}

Expressed in terms of maxsets, it leads to the following corollary.

\begin{corollary}

Formula $R$ is unobtainable from a set $K_1,\ldots,K_m$ having no maxset of
size greater than two if and only if a cycle of (single maxset not in
$R$)-(chain of odd maxsets in $R$) that crosses the same maxset at most twice
exists.

\end{corollary}

 %

\subsection{Algorithm}
\label{algorithm}


Theorem~\ref{acyclic} ensures that every or-of-maxsets is obtainable if the
maxsets form a Berge-acyclic hypergraph. The following algorithm combines the
method for iteratively labeling formulae with the search for maxsets. It is
guaranteed to work if the maxsets form a Berge--acyclic hypergraph, but may
also produce a correct result if they do not.

\begin{algorithm}
\label{labeling}

\

\begin{enumerate}

\item for each pair of formulae $K_i, K_j$, determine its consistency

\item set $L=\emptyset$

\item\label{newmaxset}
$M=\{K_i, K_j\}$, where
    $\{K_i,K_j\}$ is consistent,
    $K_i \in L$ and
    $K_j \not\in L$;
if such a pair does not exists (\eg, $L=\emptyset$) then
    $M=\{K_i\}$ with $K_i \not\in L$;
if $L$ contains all $K_i$'s, stop

\item\label{enlarge}
choose $K_j$ such that $\{K_i,K_j\}$ is consistent for every $K_i \in M$;
if no such $K_j$ exists, go to Step~\ref{addmaxset}

\item\label{consistency}
if $M \cup \{K_j\}$ is inconsistent,
go Step~\ref{enlarge} and choose another $K_j$

\item $M=M \cup \{K_j\}$ and go to Step~\ref{enlarge}

\item\label{addmaxset}
$L=L \cup M$

\item
if $M \models R$, then:

\begin{enumerate}

\item if no formula of $M$ is labeled, then label one with $1,2$ and the others
with $2$;

\item if a formula is labeled $1,n$ and the others are unlabeled, label the
others $n$

\item if a formula is labeled $n$ and the others are unlabeled, label one of
the others $1,n$ and the others $n$

\item otherwise, the set of maxsets is not acyclic: terminate with error

\end{enumerate}

\item if $M \not\models R$

\begin{enumerate}

\item if no formula of $M$ is labeled, label all of them $2$

\item if a formula is labeled $1,n$ and the others are unlabeled, label the
others $n+1$

\item if a formula is labeled $n$ and the others and unlabeled, label the
others $n$

\item otherwise, the set of maxsets is not acyclic: terminate with error

\end{enumerate}

\item go to Step~\ref{newmaxset}

\end{enumerate}

If a formula is labeled $1,n$ its priority class is one; if it is labeled $n$,
it is $n$. If the result of merging with this priority ordering is $R$, then
$R$ is obtainable.

\end{algorithm}

The final check is necessary unless $R$ is guaranteed to be an or-of-maxsets.
The algorithm includes some choices (e.g., ``choose $K_j$'', ``label one node
with 1,2'') but is not nondeterministic: arbitrary choices can be taken.

Entailment $M \models R$ can be replaced by the consistency $M \cup \{R\}$. The
algorithm can be improved by caching the inconsistent sets $M \cup \{K_j\}$
detected in Step~\ref{consistency}, especially the small ones. This information
can be useful when later checking another $M' \cup \{K_j\}$: if $M \cup \{K_j\}
\subseteq M' \cup \{K_j\}$, unsatisfiability is established at no additional
cost.

\begin{theorem}

If the maxsets of $K_1,\ldots,K_m$ are Berge-acyclic, Algorithm~\ref{labeling}
establishes the obtainability of $R$ from them and outputs a priority ordering
that generates $R$ if one exists.

\end{theorem}

\proof The algorithm works by iteratively generating a new maxset $M$ from a
labeled formula, and then labeling its other formulae according to the rules of
Theorem~\ref{acyclic}.

In particular, during the algorithm the following conditions hold:

\begin{itemize}

\item all formulae of the maxsets found so far are labeled;

\item $L$ is the union of the maxsets found so far;

\item $M$ is a subset of a maxset not (yet) in $L$.

\end{itemize}

At the beginning these conditions are vacuously true, as no maxset has been
found and no formula is labeled. No step violates them: Step~\ref{newmaxset}
guarantees that every generated $M$ is a new maxset, as it is built upon at
least a formula that is not in the previous ones; Step~\ref{addmaxset} is
reached only when $M$ is a maxset, ensuring the validity of the first of three
conditions; the two following steps label the formulae of this newly found
maxset.

Since labeling is performed as in Theorem~\ref{acyclic}, if the set of maxsets
is acyclic and $R$ is an or-of-maxsets, the result is a priority ordering
generating $R$.~\qed

If the maxsets are not Berge-acyclic, the algorithm stops when it reaches a
maxset that already contains two or more labels. In some cases, there is no way
it could continue. For example, there is no way to extend labels $1,n$ and
$1,m$ with $n \not= m$ to the rest of a selected maxset. In the other cases,
such as two labels greater than one, the algorithm may still continue and
obtain a correct ordering.

 %

\subsection{Complexity}

A necessary condition to obtainability is that the formula to obtain is the
disjunction of some maxsets of the formulae to be merged. An obvious way to
check this is to consider all possible sets of subsets of formulae, checking
that each of them is maximally consistent, and that their disjunction is
equivalent to the result to obtain. However, the problem can be reformulated in
a much simper way using some properties of maxsets.

\begin{lemma}
\label{check-or}

Formula $R$ is an or-of-maxsets of $K_1,\ldots,K_m$ if and only if, for every
$I \in \mod(R)$, it holds $M \models R$ and $M \cup \{K_i\} \models \bot$ for
every $K_i \not\in M$, where $M=\{K_i ~|~ I \models K_i\}$.

\end{lemma}

\proof By Lemma~\ref{inconsistent}, maxsets do not share models. Therefore, if
$R$ is an or-of-maxsets then each of its models is in exactly one maxset. In
particular, Lemma~\ref{model} tells that $M=\{K_i ~|~ I \models K_i\}$ is the
maxset containing $I$, if any. The additional conditions ensure that $M$ is
actually a maxset (no other formula is consistent with it) and that the
disjunction of such $M$'s do not include models not in $R$.~\qed

As a consequence of this property, checking whether $R$ is an or-of-maxsets is
not harder than propositional entailment.

\begin{theorem}
\label{complexity-orofmaxsets}

Checking whether $R$ is an or-of-maxsets of $K_1,\ldots,K_m$ is in \conp.

\end{theorem}

\proof Let $X$ be the set of variables. By Lemma~\ref{check-or}, the property
can be checked by considering each model $I$ over $X$, building $M=\{K_i ~|~ I
\models K_i\}$ and verifying a number of independent entailments: $M \models R$
and $M \cup \{K_i\} \models \bot$ for every $K_i \not\in M$. Since $M$ can be
built in polynomial time from $I$, the subproblem is equivalent to a single
validity check, and can therefore be expressed in terms of a QBF in the form
$\forall Y.F$. Since the whole problem is to check this for every model $I$
over $X$, it is equivalent to $\forall X \forall Y.F$, and is therefore in
\conp.~\qed

Hardness holds even in with only two formulae to be merged.

\begin{theorem}

Checking whether $R$ is an or-of-maxsets of a set of two formulae is
\conp-hard.

\end{theorem}

\proof The claim is proved by reduction from the problem of establishing the
unsatisfiability of a formula $F$. Reduction is as follows: formula $F$ is
inconsistent if and only if $R=\neg c$ is an or-of-maxsets of $A=\neg c$ and
$B=c \vee (d \wedge F)$, where $c$ and $d$ are two new variables, not occurring
in $F$.

Regardless, $A \wedge B$ is $\neg c \wedge d \wedge F$ by resolution. As
result, if $F$ is inconsistent so is $A \wedge B$. Therefore, the maxsets are
$\{A\}$ and $\{B\}$. Since $R$ is the same as $A$, it can be seen as the
disjunction of the single element $\{A\}$.

If $F$ is consistent, so is $A \wedge B$. Therefore, the only maxset is
$\{A,B\}$, which is equivalent to $A \wedge B = \neg c \wedge d \wedge F$.
Model $\{c=\false, d=\false\}$ falsifies this formula while satisfying $R$.
Therefore, $R$ is not an or-of-maxsets.~\qed

These results do not require $R$ to be consistent. If it is not, $R$ is still
an or-of-maxsets, as $\bigvee \emptyset = \bot$. However, this case is not
allowed as a result of merging: an inconsistent formula is never obtainable.

By Lemma~\ref{three}, if the formulae are three or less then every consistent
or-of-maxset is obtainable. By definition, obtainable formulae are
or-of-maxsets. Therefore, the last theorem also proves the complexity of
obtainability in this case.

\begin{corollary}

Checking whether a consistent formula is obtainable by priority base merging
from two formulae is \conp-hard.

\end{corollary}

Unfortunately, Theorem~\ref{complexity-orofmaxsets} does not extend to
obtainability. Indeed, while verifying whether a formula is an or-of-maxsets
can be done ``locally'', by checking each model $I$ and its maxset $M$ at time,
obtainability is a global conditions over the maxsets: one of them may be
selected or not depending on the others. This makes the problem harder than
checking whether a formula is an or-of-maxsets.

\begin{theorem}

Checking whether a formula is obtainable by priority base merging is in \S{3}.

\end{theorem}

\proof By Lemma~\ref{model}, for every model $I$ of a maxset $M$ it holds $M =
\{K_i ~|~ I \models K_i\}$. This provides a way for expressing the problem of
obtainability of $R$ from $K_1,\ldots,K_m$: there exists a priority ordering
$P$ such that every model of $R$ corresponds to a minimal maxset and every
model of $\neg R$ corresponds to a subset that is either non-minimal or not a
maxset at all.

Formally, for every model $I$ of $R$ the set $M=\{K_i ~|~ I \models K_i\}$
should be a minimal maxset. By Lemma~\ref{subset}, this is equivalent to $M$
being not greater than another consistent subset $N$. In other words, for every
$N \subseteq \{K_1,\ldots,K_m\}$ either $N$ is inconsistent or it is not less
than $M$ according to $P$. Comparing according to $P$ can be done in polynomial
time, as it amounts to checking which formulae of $M$ and $N$ are in $P(1)$,
$P(2)$, etc. The quantifiers are all universal; therefore, the subproblem can
be expressed as a $\forall QBF$.

Regarding the models $I$ not of $R$, the set $M=\{K_i ~|~ I \models K_i\}$
should not be a minimal consistent subset according to $P$. Since $M$ is
consistent (because it has the model $I$), this is equivalent to the existence
of another consistent subset $N$ that is less than it according to the
ordering. This second subproblem is therefore in the form: "for all $I
\not\models R$ there exists $N \subseteq \{K_i\}$, etc. As a result, it can be
expressed as a $\forall \exists QBF$.

Both these conditions have to hold for a priority ordering $P$: the problem is
to establish the existence of a $P$ such that both hold. As a result, the whole
problem is expressed as a $\exists \forall \exists QBF$, and is therefore in
\S{3}.~\qed

The following result shows that even with four formulae (the smallest case of
unobtainable consistent or-of-maxsets) obtainability is \conp-hard even if the
formula is assumed to be a consistent or-of-maxsets.

\begin{theorem}

Checking whether $R$ is obtainable by priority base merging from four formulae
is \conp-hard, and this result holds even assuming that $R$ is a consistent
or-of-maxsets.

\end{theorem}

\proof The claim is proved by reduction from propositional unsatisfiability. By
Lemma~\ref{unobtainable}, $R=(A \wedge B) \vee (C \wedge D)$ is not obtainable
from $A,B,C,D$ if the maxsets are $\{A,B\}$, $\{B,C\}$, $\{C,D\}$ and
$\{D,A\}$. Lemma~\ref{synthesis} gives the following formulae:

\begin{itemize}

\item $A = (x \wedge y) \vee (\neg x \wedge \neg y)$
\item $B = (x \wedge y) \vee (x \wedge \neg y)$
\item $C = (x \wedge \neg y) \vee (\neg x \wedge y)$
\item $D = (\neg x \wedge y) \vee (\neg x \wedge \neg y)$

\end{itemize}

The maxset $\{D,A\}$ is equivalent to $\neg x \wedge \neg y$. A formula $F$ can
be added to it by changing $D$ and $A$:

\begin{itemize}

\item $A' = (x \wedge y) \vee (\neg x \wedge \neg y \wedge F)$
\item $B = (x \wedge y) \vee (x \wedge \neg y)$
\item $C = (x \wedge \neg y) \vee (\neg x \wedge y)$
\item $D' = (\neg x \wedge y) \vee (\neg x \wedge \neg y \wedge F)$

\end{itemize}

This provides the required reduction from propositional unsatisfiability to
obtainability. Indeed, if $x$ and $y$ are two new variables, not occurring in
$F$, then $F$ is unsatisfiable if and only if $R=(A' \wedge B) \vee (C \wedge
D')$ is obtainable from $A',B,C,D'$.

The maxsets of the four formulae are $\{A',B\} \equiv x \wedge y$, $\{B,C\}
\equiv x \wedge \neg y$, $\{C,D'\} \equiv \neg x \wedge y$ and, if $F$ is
consistent, $\{D',A'\} \equiv \neg x \wedge \neg y \wedge F$. As a result, if
$F$ is consistent then maxsets are as in Lemma~\ref{unobtainable}, and $R$ is
therefore unobtainable. Otherwise, there are only three maxsets, and $R$ is the
disjunction of two of them. Lemma~\ref{three} ensures that every or-of-maxsets
is obtainable in this case.~\qed

Obtainability depends on the existence of orderings over the maxsets, which may
be exponentially many. This number reduces to quadratic if the maxsets comprise
at most two formulae.

\begin{theorem}

Checking whether a consistent or-of-maxsets is obtainable by priority base
merging is in \conp\  if all maxsets comprise at most two formulae.

\end{theorem}

\proof The result is unobtainable if the graph of maxsets is unobtainable,
which by Corollary~\ref{characterization} is equivalent to the presence of an
alternating cycle. Since the nodes are formulae, this condition can be
reformulated as: there exists a sequence of formulae $A_1, B_1, A_2, B_2,
\ldots$, each appearing at most twice, such that:

\begin{enumerate}

\item every pair of consecutive formulae is consistent: $A_i \wedge B_i
\not\models \bot$, $B_i \wedge A_{i+1} \not\models \bot$, \ldots; checking that
such pairs are also maximally consistent is unnecessary by the assumption that
no maxset contains more than two formulae;

\item $A_i \wedge B_i \wedge R \not\models \bot$: by Lemma~\ref{selected}, this
is equivalent to $\{A_i,B_i\}$ being selected;

\item either $B_i \wedge A_{i+1} \not\models R$ or $A_{i+1} \wedge B_{i+1}
\wedge R \not\models \bot$; still by Lemma~\ref{selected}, this condition is
equivalent to: if $\{B_i,A_{i+1}\}$ is selected, so is $\{A_{i+1},B_{i+1}\}$.

\end{enumerate}

Selection can be expressed both as $M \models R$ and $M \wedge R \not\models
\bot$. Using the first condition when the requirement is negated and the second
when it is positive allows expressing unobtainability in terms of
non-entailment only. In particular, it is reformulated as the existence of such
a cycle that satisfies a number of conditions based on non-entailment.
Therefore, unobtainability is in \np, and obtainability in \conp.~\qed

This allows for a precise characterization of complexity for the case of binary
maxsets.

\begin{corollary}

Checking whether a consistent or-of-maxsets is obtainable by priority base
merging is \conp\  complete if all maxsets comprise at most two formulae.

\end{corollary}

\subsection{Constant number of formulae}

Unobtainability is monotonic with respect to the excluded sets: adding new ones
and enlarging the existing ones does not change unobtainability. The following
lemma concerns the obtainability of a pair $(S,E)$, where $S$ and $E$ are sets
of sets of formulae, not necessarily maxsets and not necessarily all of them.
The definition is repeated here for the sake of readability: $(S,E)$ is
obtainable if there exists an ordering that makes $S$ to coincide with the set
of minimal sets among $S \cup E$. In other words, $(S,E)$ is obtainable if
there exists an ordering that makes the sets in $S$ to be the minimal ones
among $S \cup E$.

\begin{lemma}
\label{monotonic}

If $S$ and $E$ are sets of sets such that none is contained in another and
$(S,E)$ is not obtainable so is $(S,E')$, where $E'$ is the result of adding
some sets of formulae to $E$ and some formulae to some sets of $E$.

\end{lemma}

\proof Given the assumption of no mutual containment, every pair
$(S,\emptyset)$ is obtainable by placing all formulae of $S$ in class one.
Therefore, unobtainability is due to the presence of $E$: every partition that
selects $S$ also selects some $N \in E$. By definition, this means that $N$ is
minimal according to the ordering: for every $M \in S$, the two sets $N$ and
$M$ coincide up to class $n-1$ but $N \cap P(n) \not\subseteq M \cap P(n)$ for
some class $n$, possibly $n=1$. Adding formulae to $N$ or new sets to $E$ does
not change this condition.~\qed

Obtainability can be defined as follows: there exists a set $S$ such that the
result is equivalent to $\bigvee S$, $S$ is a subset of maxsets and $(S,E)$ is
obtainable, where $E$ are the maxsets not in $S$. In the case of a constant
number of formulae, their sets and therefore maxsets are in constant number as
well. Quantifying over them does not therefore increase the complexity of the
problem.

However, the remaining quantifications are not all of the same kind. For
example, the condition that $R$ is an or-of-maxsets is:

\[
\begin{array}{c}
R \mbox{ is an or-of-maxsets of } \{K_1,\ldots,K_m\} \\
\Updownarrow \\
\exists S \subseteq 2^{\{K_1,\ldots,K_m\}} \mbox{ such that } \\
\begin{array}{l}
R \equiv \bigvee S \\
\forall M \in S ~.~ M \not\models \bot \mbox{ and }
\forall K \not\in M ~.~ M \cup \{K\} \models \bot
\end{array}
\end{array}
\]

The quantifiers over $S$, $M$ and $K$ are not a problem because the choice are
on sets of constant cardinality. Instead, the formula $M \not\models \bot$ is
an existential quantification (there exists a model satisfying all formulae of
$M$) while all others are universal (e.g.\  all models satisfying $M$ also
satisfy $\bigvee S$).

Such a quantifier can be removed by relaxing the condition over $M$, accepting
some other ones. This is the technique used by Nebel~\cite{nebe-98} for the
generalized closed-world assumption (GCWA) and the WIDTIO revision: instead of
considering only the sets specified by the definition, allow others that do not
affect the final result. Omitting details, $GCWA(T)$ is $T$ with a certain set
of literals $F$ added; what made determining the exact complexity of the
problem $GCWA(T) \models A$ difficult was that checking membership of a single
literal in $F$ is already \P{2}-hard, thus requiring a polynomial calls to a
\P{2} oracle for $T \cup F \models A$. Nebel~\cite{nebe-98} overcome this
difficulty by switching from $F$ to its supersets: $T \cup F \not\models A$ if
and only if $T \cup S \not\models A$ for some $S \supseteq F$. In spite of the
seeming increase of complexity, the problem is simplified because checking
whether $S \supseteq F$ is in \S{2}. Therefore, the whole non-entailment
problem is in \S{2}, as it amounts to guess a set $S$ satisfying a condition in
\S{2} and a model that satisfies $T \cup S$ but not $A$. In a nutshell, the
core of the method is: "instead of the specific set $F$ use a group that
includes it, provided that the other sets do not affect the final result".

In the present case, the key point is that if $S$ contains an inconsistent set
$M$, then $\bigvee S = \{M\} \vee \bigvee (S \backslash \{M\}) = \bot \vee
\bigvee (S \backslash \{M\}) = \bigvee (S \backslash \{M\})$: inconsistent sets
do not contribute to the disjunction. As a result, the condition can be relaxed
by allowing such sets $M$: requiring that $M$ is a maxset is changed into just
$M \cup \{K\} \models \bot$ for every $K \not\in M$. The $M$'s satisfying this
condition are either maxsets or inconsistent sets of formulae, but the latter
do not affect $\bigvee S$.

\[
\begin{array}{c}
R \mbox{ is an or-of-maxsets of } \{K_1,\ldots,K_m\} \\
\Updownarrow \\
\exists S \subseteq 2^{\{K_1,\ldots,K_m\}} \mbox{ such that } \\
\begin{array}{l}
R \equiv \bigvee S \\
\forall M \in S ~
\forall K \not\in M ~.~ M \cup \{K\} \models \bot
\end{array}
\end{array}
\]

This condition contains only universal quantifiers: $R \equiv \bigvee S$ is
equivalent to ``every model satisfying $R$ also satisfies $\bigvee S$ and vice
versa''; $M \cup \{K\} \models \bot$ is ``every model falsifies $M \cup
\{K\}$''. The quantifiers over $S$, $M$ and $K$ are choices over sets of
constant cardinality, so they do not affect complexity. They can be replaced by
conjunctions and disjunctions.

As a result, checking whether $R$ is an or-of-maxsets is in \conp\  for a
constant number of formulae. This fact is subsumed by
Theorem~\ref{complexity-orofmaxsets}, which states the same for any number of
formulae. However, with some changes the condition extends to obtainability,
for which no similar result hold in the general case. Lemma~\ref{monotonic}
ensure the correctness of relaxing.

\begin{lemma}
\label{obtainable-constant}

$R$ is obtainable by priority base merging from $K_1,\ldots,K_m$ if and only if
there exists a nonempty $S \subseteq 2^{\{K_1,\ldots,K_m\}}$ such that:

\begin{enumerate}

\item $R \equiv \bigvee S$;

\item $\forall M \in S$, $\forall K \not\in M$, $M \cup \{K\} \models \bot$;

\item $\forall E \subseteq 2^{\{K_1,\ldots,K_M\}}$, either
  $\exists M \in E$ such that $M \models \bot$ or
  $\exists M \in E$ such that $M \subseteq M'$ for some $M' \in S$ or
  $(S,E)$ is obtainable.

\end{enumerate}

\end{lemma}

\proof The first two points are equivalent to $R$ being an or-of-maxsets. The
third resembles the definition of obtainability, but $E$ is not the set of
maxsets not in $S$. Rather, if the condition is false is an arbitrary set of
consistent subsets such that $(S,E)$ is not obtainable.

Lemma~\ref{monotonic} however ensures that such a set $E$ can be enlarged by
adding arbitrary new sets and arbitrary new formulae to existing sets, and the
pair $(S,E)$ remains unobtainable. As a result, if there exists $E$ such that
$(S,E)$ is unobtainable, $E$ can be added formulae and sets to make it the set
of maxsets not in $S$.

\begin{description}

\item[$R$ obtainable.] The three conditions above hold for $S$ equal to the set
of selected maxsets. This choice makes the first and second points true. If the
third point were false, then $(S,E)$ would be unobtainable for some set of
consistent sets $E$ such that none of its element is contained in one of $S$.
Since an $N \in E$ is not contained in a selected maxset, it can be enlarged to
make it a maxset, and that would be an excluded one. Adding the other excluded
maxsets, $E$ is turned into the set of excluded maxsets $E'$. By
Lemma~\ref{monotonic}, since $(S,E)$ is unobtainable so is $(S,E')$,
contradicting the assumption that $R$ is obtainable.

\item[$R$ unobtainable.] If $R$ is not an or-of-maxsets, then for no $S$ points
1 and 2 hold. Otherwise, $R$ is an or-of-maxsets $S$ but $(S,E)$ is not
obtainable, where $E$ is the set of the other maxsets. For such $E$ the third
point of the condition is violated.

\end{description}~\qed

The condition of this lemma only contains universal quantifier, apart the ones
on sets of constant size. The complexity of the problem is the obvious
consequence of this.

\begin{corollary}

Checking obtainability by priority base merging from a constant number of
formulae is in \conp.

\end{corollary}

Once obtainability is established, the problem is to find the ordering
generating the result. This problem can be recast as that of checking whether a
partial assignment of formulae to classes can be extended to form an ordering
generating the required result of merging.

\begin{theorem}

Checking whether a priority ordering can be extended to generate $R$ as the
result of merging a constant number of formulae $K_1,\ldots,K_m$ is \conp\
complete.

\end{theorem}

\proof The problem is hard with an empty ordering, as it is equivalent to
obtainability. It is also in \conp: it is the same as obtainability by adding
the condition that the ordering extends the given one. In the statement of
Lemma~\ref{obtainable-constant}, the only point where the ordering matters is
when $(S,E)$ is checked to be obtainable. Therefore, the problem can be
expressed by simply changing the subcondition ``$(S,E)$ is obtainable'' into
``$(S,E)$ is obtained by an ordering extending the given partial one'' in the
statement of Lemma~\ref{obtainable-constant}. Since the additional check has
cost linear in the number of the formulae, complexity remains the same.~\qed

A related question is whether a priority ordering can be uniquely extended to
generate the required result. This amounts to finding such an ordering, if any,
and then checking that no other priority ordering would do the same.

\begin{theorem}

Checking whether a priority ordering not extending a given one generates $R$ as
the result of merging a constant number of formulae $K_1,\ldots,K_m$ is
\conp-complete.

\end{theorem}

\proof Lemma~\ref{obtainable-constant} expresses this problem by changing the
condition that $(S,E)$ is obtainable to its obtainability with an ordering not
extending the given one. This proves that the problem is in \conp.

Hardness is proved using three formulae with maxsets $\{A,B\}$, $\{A,C\}$, and
$\{B,C\}$, where the latter is excluded and only exists if a formula $F$ is
satisfiable.

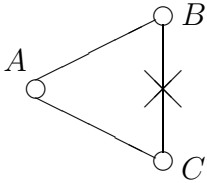
\setlength{\unitlength}{5000sp}%
\begingroup\makeatletter\ifx\SetFigFont\undefined%
\gdef\SetFigFont#1#2#3#4#5{%
  \reset@font\fontsize{#1}{#2pt}%
  \fontfamily{#3}\fontseries{#4}\fontshape{#5}%
  \selectfont}%
\fi\endgroup%
\begin{picture}(795,930)(4981,-3856)
{\color[rgb]{0,0,0}\thinlines
\put(5041,-3391){\circle{90}}
}%
{\color[rgb]{0,0,0}\put(5671,-3031){\circle{90}}
}%
{\color[rgb]{0,0,0}\put(5671,-3751){\circle{90}}
}%
{\color[rgb]{0,0,0}\put(5671,-3076){\line( 0,-1){630}}
}%
{\color[rgb]{0,0,0}\put(5041,-3436){\line( 2,-1){594}}
}%
{\color[rgb]{0,0,0}\put(5041,-3346){\line( 2, 1){594}}
}%
{\color[rgb]{0,0,0}\put(5581,-3301){\line( 1,-1){180}}
}%
{\color[rgb]{0,0,0}\put(5581,-3481){\line( 1, 1){180}}
}%
\put(5761,-3076){\makebox(0,0)[lb]{\smash{{\SetFigFont{12}{24.0}{\rmdefault}{\mddefault}{\updefault}{\color[rgb]{0,0,0}$B$}%
}}}}
\put(5761,-3841){\makebox(0,0)[lb]{\smash{{\SetFigFont{12}{24.0}{\rmdefault}{\mddefault}{\updefault}{\color[rgb]{0,0,0}$C$}%
}}}}
\put(4996,-3301){\makebox(0,0)[rb]{\smash{{\SetFigFont{12}{24.0}{\rmdefault}{\mddefault}{\updefault}{\color[rgb]{0,0,0}$A$}%
}}}}
\end{picture}%
 %
\nop
{
  /---B
 /    |
A     X
 \    |
  \---C
}

If the third maxset exists, the only ordering excluding it while selecting the
other two is the one containing $A$ in class one and $B$ and $C$ in class two.
Indeed, if both $B$ and $C$ are in class one, by Lemma~\ref{all} $\{B,C\}$
would be selected. If $A$ and $B$ are in class one and $C$ is not, $\{A,B\}$
would be excluded. Since either $A$ or $B$ is in class one by Lemma~\ref{one},
the only remaining case is $A$ in class one. The other two formulae $B$ and $C$
cannot be in different classes, as otherwise one between $\{A,B\}$ and
$\{A,C\}$ would be excluded. Therefore, the only ordering obtaining the
required result has $A$ in class one and $B$ and $C$ in class two.

The same ordering selects the same two maxsets even if the third maxset does
not exists. Since the result is the disjunction of all maxsets, Lemma~\ref{all}
applies: it is also obtained by placing all three formulae in class one.
Therefore, a second ordering selects $\{A,B\}$ and $\{A,C\}$ in this case.

The problem is therefore that of generating formulae such that $\{B,C\}$ is
consistent if and only if a formula $F$ is. Lemma~\ref{synthesis}, with $F$
added to $\{B,C\}$, gives:

\begin{eqnarray*}
A &=& (x \wedge \neg y) \vee (\neg x \wedge \neg y) \\
B &=& (x \wedge \neg y) \vee (x \wedge y \wedge F) \\
C &=& (\neg x \wedge \neg y) \vee (x \wedge y \wedge F)
\end{eqnarray*}

The set of all three formulae is inconsistent, as $A$ is only satisfied by
partial models $\{x=\true, y=\false\}$ and $\{x=\false, y=\false\}$, while $C$
is falsified by the first and $B$ by the second. Pairs of formulae are all
consistent:

\begin{eqnarray*}
\{A,B\} &=& x \wedge \neg y \\
\{A,C\} &=& \neg x \wedge \neg y \\
\{B,C\} &=& x \wedge y \wedge F
\end{eqnarray*}

The third is consistent if and only if $F$ is consistent. As a result, the
maxsets $\{A,B\}$ and $\{A,C\}$ always exist, and are selected when the
required result is $R=\neg y$ because they are consistent with it. The third
maxset $\{B,C\}$ only exists if $F$ is consistent, and if this is the case is
excluded because it is inconsistent with $R$.

As shown before, $R$ is uniquely obtainable if and only if $\{B,C\}$ is not a
maxset, which is equivalent to the inconsistency of $F$. As a result, unique
obtainability is \conp-hard.~\qed

 %

\section{What to do in case of unobtainability}

\hidden{
[titolo lungo e brutto, ma lasciare perche' e' uno degli aspetti centrali; qui
le alternative vengono esplorate solo nel caso delle priorita', ma va bene
cosi' perche' e' la sezione precedente]
}

After establishing obtainability, the next step is to determine the weights or
priority ordering. The algorithms in Section~\ref{search} and
Section~\ref{algorithm} searches for them, but of course cannot find anything
in case of unobtainability. The question that remain is therefore: what to do
in this case?

Various possibilities exist. One is to relax the condition that $R$ is exactly
the outcome of merging, still maintaining that $R$ is a formula that is known
to be true. Lifting equivalence and only requiring consistency is coherent with
this principle: $R$ does not discriminate among its models, so each could be
the actual state of the world. An integration that results in a formula
containing one of the them is still consistent with the assumptions.

\begin{lemma}

There exists a priority partition such that merging $K_1,\ldots,K_m$ is
consistent with $R$ if and only if $R$ is consistent with one of the maxsets of
$K_1,\ldots,K_m$.

\end{lemma}

\proof If one of the maxsets is consistent with $R$, the ordering of
Lemma~\ref{one} allows selecting it only. The result of merging is equal to
this maxset, which by assumption is consistent with $R$.

In the other way around, if $R$ is consistent with the result of merging
$K_1,\ldots,K_m$ with some ordering, since this result is the disjunction of
some of the maxsets, then $R$ is consistent with at least a maxset.~\qed

Even when merging is not supposed to be a process of search of a single
propositional model, a similar idea can be applied. Assuming that the situation
is characterized by a set of models, both the result of merging and $R$ result
from bounding it as close as possible. The difference is that $R$ is known to
be correct, so it contains all these models, while merging only aims at doing
the same. Under this assumption, the problem is to find an ordering such that
the set of models of $R$ is strictly contained into the result of revision.
Since what is known about this set is only that $R$ contains it, the result of
merging should be implied by $R$. Unfortunately, this condition does not
constraint the ordering at all.

\begin{lemma}

Merging $K_1,\ldots,K_m$ with some priority ordering is entailed by $R$ if and
only if $R$ entails the disjunction of all maxsets.

\end{lemma}

\proof If $R$ entails the disjunction of all maxsets, such a disjunction can be
obtained as the result of the revision by the ordering in Lemma~\ref{all}. Vice
versa, if $R$ entails the result of merging $K_1,\ldots,K_m$ with some
ordering, since this result is the disjunction of some maxsets, then $R$ also
entails the disjunction of all maxsets.~\qed

Requiring that $R$ is entailed by the result of merging or consistent with it
gives no information about the relative reliability of the sources. To obtain
such an information some additional constraint is needed, such as $R$ being as
close as possible to the result of merging, possibly also implying or being
consistent with it. In other words, the aim moves from obtaining $R$ with the
appropriate priorities to approximating it as much as possible.

If a result is unobtainable, another possible line of action is to consider
whether the given pieces of knowledge produce it using a different merging
mechanism. In other words, instead of using merging by priorities, one of the
many other
systems~\cite{koni-pere-11,pepp-08,koni-etal-04,evar-etal-10,jin-thie-07,libe-scha-98-b}
may be employed instead.

Another possible solution is to split sources based on the variables. If a
renowed computer scientist tells some property of computational classes and
that the fastest way to go a certain restaurant is to turn left at the next
turn, the first information should be assigned higher priority than the second,
as there is no a priori reason why an expert in computing should know the roads
better than anyone else. According to this principle, when a result is not
obtainable some source $K_i$ may be split into $\{K_i^1,\ldots,K_i^r\}$, for
example using a partition of the variables to decide which part of $K_i$ goes
into $K_i^1$, which in $K_i^2$, etc.

A totally different direction is to lift the assumption that $R$ is a formula
known with certainty. Instead, it could be just a formula coming from a source
of high reliability. Obtainability then generalizes to the case where no such
source may be available~\cite{libe-14}.

Even with all these alternatives, it is still possible that the known
information $R$ cannot be obtained from the knowledge bases. For example, no
semantics allows obtaining $R=x$ from $K_1=\neg x$ and $K_2=\neg x$. This is
however a rational outcome: if the knowledge bases totally agree, merging
should produce them as the result, no matter by which weights, priorities or
other relative reliability measure. If $x$ is true, then two knowledge bases
equal to $\neg x$ are just useless. Unobtainability provides significant
information even in this case: the sources are unreliable, and can therefore be
ignored from this point on.

 %

\section{Conclusions}

In this article, the problem of establishing the relative reliability of
knowledge bases given the result of their merge is studied. This is in a way a
reverse of the usual problem of merging them, in a similar way as
abduction~\cite{douv-11} reverses implication: from some information one
attempts at deriving what has generated it.

Two semantics for merging are considered for this inversion: sums of
distances~\cite{koni-pere-11,koni-lang-marq-02,koni-lang-marq-04} and priority
base merging~\cite{nebe-92,nebe-98,rott-93,delg-dubo-lang-06}. In a way, these
can be considered at the extreme opposite of the spectrum of the many possible
semantics for merging~\cite{koni-pere-11,delg-dubo-lang-06}: the first is
numeric, model-based and majority-obeying; the second is qualitative
(priority-based), syntax-dependent and not majority-obeying. The idea of
obtaining reliability information, in whichever form they are expressed, can be
however applied to other semantics for merging.

The main result proved for the semantics based on the sum of distances is an
equivalent formulation for the condition of $K_1$ and $K_2$ generating $R$ with
some weights. From this, complexity upper bounds follow, as well as the core of
a local search algorithm for determining weights. In particular, whenever the
distance measure used is in \P{i} or in \S{i}, obtainability is in \P{i+1}. Two
relevant measures are the drastic and the Hamming distances, for which the
problem is proved \conp\  and \P{2}-complete, respectively. A tractable subcase
is proved.

The complexity analysis on priority base merging shows that obtainability is
not harder than computing the result of merging with a fixed priority ordering
for the considered subcases. Given that obtainability is the existence of a
priority ordering generating a given result, at a first looks it may seem
harder. Most of the problems in belief revision are at the second level of the
polynomial
hierarchy~\cite{eite-gott-91-b,eite-gott-96,libe-97,nebe-98,libe-scha-99}, even
in some simple restrictions like two formulae to be integrated. In contrast,
obtainability proved \conp\  complete with a constant number of formulae or
with maxsets of two or less formulae. The problem of obtainability in general
is however still open, so it may prove harder. If
Corollary~\ref{characterization} extends in some form from graphs to
hypergraphs, obtainability may be still in \conp\  in the general case.

What to do if the result is not obtainable? Various alternatives are outlined:
relax the condition that $R$ is exactly the result of merging, use another
semantics of merging (for example, if $R$ is unobtainable with priority merging
one may try the weighted sum of Hamming distances), split the sources (for
example, by the variables), lift the assumption that $R$ is known with
certainty. However, in some cases a result should not be obtainable, like when
all sources agree on $x$ and the result is $\neg x$; in such cases,
unobtainability still provide the useful warning that the sources are
unreliable.

While the present article concentrates on obtainability, a sensible question is
whether a given result is uniquely obtainable or not; another question is
whether it can be obtained not with arbitrary weights or priorities, but some
obeying some constraints, such as the weight of a base being greater than that
of another.

 %

\bibliographystyle{plain}

\appendix

\end{document}